\documentclass[pdflatex,sn-mathphys]{sn-jnl}

\jyear{2022}%

\theoremstyle{thmstyleone}%
\theoremstyle{thmstyletwo}%

\theoremstyle{thmstylethree}%

\usepackage{xcolor}

\begin{document}

\title[ ]{Model interpretation using improved local regression with variable importance}


\author*[1]{\fnm{Gilson Y.} \sur{Shimizu}}\email{gilsonshimizu@yahoo.com.br}

\author[2]{\fnm{Rafael} \sur{Izbicki}}\email{rafaelizbicki@gmail.com}

\author[1]{\fnm{Andre C. P. L. F. de} \sur{Carvalho}}\email{andre@icmc.usp.br}

\affil[1]{\orgdiv{Institute of Mathematics and Computer Sciences}, \orgname{University of São Paulo}, \orgaddress{\city{São Carlos}, \state{SP}, \country{Brazil}}}

\affil[2]{\orgdiv{Department of Statistics}, \orgname{Federal University of São Carlos}, \orgaddress{\city{São Carlos}, \state{SP}, \country{Brazil}}}



\abstract{
A fundamental question on the use of ML models concerns the explanation of their predictions for increasing
transparency in decision-making. Although several interpretability methods have emerged, some gaps regarding the reliability of their explanations have been identified. For instance, most methods are unstable (meaning that they give very different explanations with small changes in the data), and do not cope well with irrelevant features (that is, features not related to the label).
This article introduces two new interpretability methods, namely VarImp and SupClus, that overcome these issues by using  local regressions fits with a weighted  distance that takes into account  variable importance. Whereas VarImp generates explanations for each instance and can be applied to datasets with more complex relationships, SupClus interprets clusters of instances with similar explanations and can be applied to simpler datasets where clusters can be found.
We compare our methods with state-of-the art approaches and show that it yields better explanations according to several metrics,
particularly in high-dimensional problems with irrelevant features, as well as when the relationship between features and target is non-linear.
}




\maketitle

\section{Introduction}\label{introduction}

The increasing popularity of machine learning (ML), together with the growing concern regarding transparency model interpretability
have demanded the
explanations for their decision-making processes, mainly model predictions.
Especially in areas that involve risk, good predictions are not sufficient, since professionals must understand the reason for a decision for ensuring trust, fairness, and transparency. Some models such as decision trees and linear models are naturally interpretable; however, the models induced by the ML algorithms with known high
predictive power are considered black boxes. The lack of interpretability prevents these models from being applied in a wider range of areas.

Model interpretability concerns how much the human being can understand the cause for a decision or predict the outcome of a model \citep{miller2019explanation, kim2016examples, molnar2022}. Linear models are highly interpretable, since the regression coefficients indicate whether the relationship between a feature and a target is positive or negative.
It is important to distinguish interpretability from explainability, another important aspect of transparency. According to \citep{BARREDO2020}, while model explainability is how well a user can understand the internal mechanisms of a model work, model interpretability is the understanding of how well a user can understand how a model makes its decisions, without need to understand the model internal work, being, therefore, more associated with model transparency.

Interpretability methods can explain all instances globally or only one instance locally, and can be either specific to a ML algorithm, or agnostic. Some ML algorithms are intrinsically interpretable (see, e.g., \citep{coscrato2019nls,agarwal2021neural} and references therein), whereas others require a post hoc method of interpretability  (see, e.g., \cite{hechtlinger2016interpretation,koh2017understanding,lundberg2017unified,ribeiro2016should,botari2020melime} for a review).
The methodologies of local and agnostic interpretability that have stood out are based on a penalized local linear regression \citep{ribeiro2016should, molnar2018iml, botari2020melime}. The key idea is that the relationship between features and target is linear in a small region around an instance. Another prominent approach is based on the Shapley values of cooperative game theory \citep{shapley1953value, vstrumbelj2014explaining, lundberg2017unified}. In this approach, features are seen as players and a prediction is the end result of a game; therefore, Shapley values represent the contribution of each feature for a given 
prediction. 
Some authors 
propose methods that interpret sets of instances with similar explanations as
clusters in a partition
created via an unsupervised algorithm or as a decision tree \citep{zafar2021deterministic, hall2017machine, hu2018locally}. In this approach, when clusters are used, interpretable linear models are further fitted for each cluster.

Although many interpretability methods have emerged in recent years, some challenges still remain. 
Despite being simple and intuitive, a criticism to
the methods with perturbed samples, such as LIME and Shapley values, is that they can generate unreliable explanations \citep{alvarez2018robustness, slack2020fooling}, since perturbed samples may not 
follow the true data distribution. 
Methods that use penalization (e.g., LASSO \citep{tibshirani1996regression}) may also be unreliable, due to their focus on prediction rather than on the inference of regression coefficients, thus requiring further inferential analyses \citep{lee2016exact}. Another 
negative
aspect is that the Euclidean distance used in local regressions may not be suitable, since relevant and irrelevant variables contribute equally to the calculation of distances.
Finally, although clustering-based methods have the advantage of generating more stable explanations, they have difficulties
to effectively group instances with similar explanations, since traditional clustering algorithms do not use the target variable. Even decision trees are often not suitable, since they do not take into account
the sign of the relationship between variables and target.

\textbf{Contributions.}
This article proposes two new interpretability methods, namely VarImp (Variable Importance) and SupClus (Supervised Cluster), to
overcome the deficiencies of reliability and quality of interpretations
of the current methods. 
The key idea of the former is to use a local variable importance as a weight in calculating distances within a local linear regression; therefore, variables more relevant to an instance will have larger weights in distance calculations. SupClus uses the coefficients obtained in VarImp to group similar instances and interpret the obtained clusters through linear models. Whereas VarImp creates different explanations for each instance, SupClus interprets the cluster to which the instance belongs. Consequently, VarImp can be applied to more complex datasets and SupClus can detect simpler datasets and be used, e.g., when
only combinations of linear models are sufficient to interpret the original model.
An R package is made available to allow readers to use the proposed methods.

The article is organized as follows: Sections \ref{sec:varimp} and \ref{sec:supclus} introduce VarImp and SupClus, respectively; 
Section \ref{sec:experiments} describes the experiments carried out on artificial and real data and presents the measures used for the evaluation of interpretability methods; finally, Section \ref{sec:results} presents and analyses the experimental
results.

\textbf{Notation.}
For simplicity of notation, we assume our i.i.d. dataset  is split into two parts with the same  size $n$: 
the set $\mathbb{D}'=\{ (\mathbf{x}'_1, y'_1), \dots, (\mathbf{x}'_n, y'_n) \}$ is used to train the original model, which we denote by  $f$, while the set $\mathbb{D}=\{ (\mathbf{x}_1, f(\mathbf{x}_1)), \dots, (\mathbf{x}_n, f(\mathbf{x}_n)) \}$ is employed to train the local model interpreter. In a regression task, $f$ represents the  regression prediction, while in a classification task, $f$ represents the estimated probabilities of the chosen label. We denote the local model interpreter as $g$. Finally, we denote by $x_{i,j}$  the value of the $j$-th feature for an instance $i$, with $j=1,\dots,d$.

\section{VarImp method}\label{sec:varimp}

As LIME \citep{ribeiro2016should}, VarImp is based on the assumption that  predictions of the original model can be locally approximated by a linear model. However,  rather than computing weights needed to implement such model using plain Euclidean (or related) distance, we define
a matrix that uses  importance weights.
Moreover, since penalization methods can cause instability in the inference of coefficients \citep{lee2016exact}, a Forward Stepwise variable selection algorithm \citep{james2013introduction} was adopted, with the possibility of choosing a maximum number of features $K \leq d$.

\begin{figure}[ht]
        \centering
        \begin{subfigure}[b]{0.49\textwidth}
            \centering
            \includegraphics[width=\textwidth]{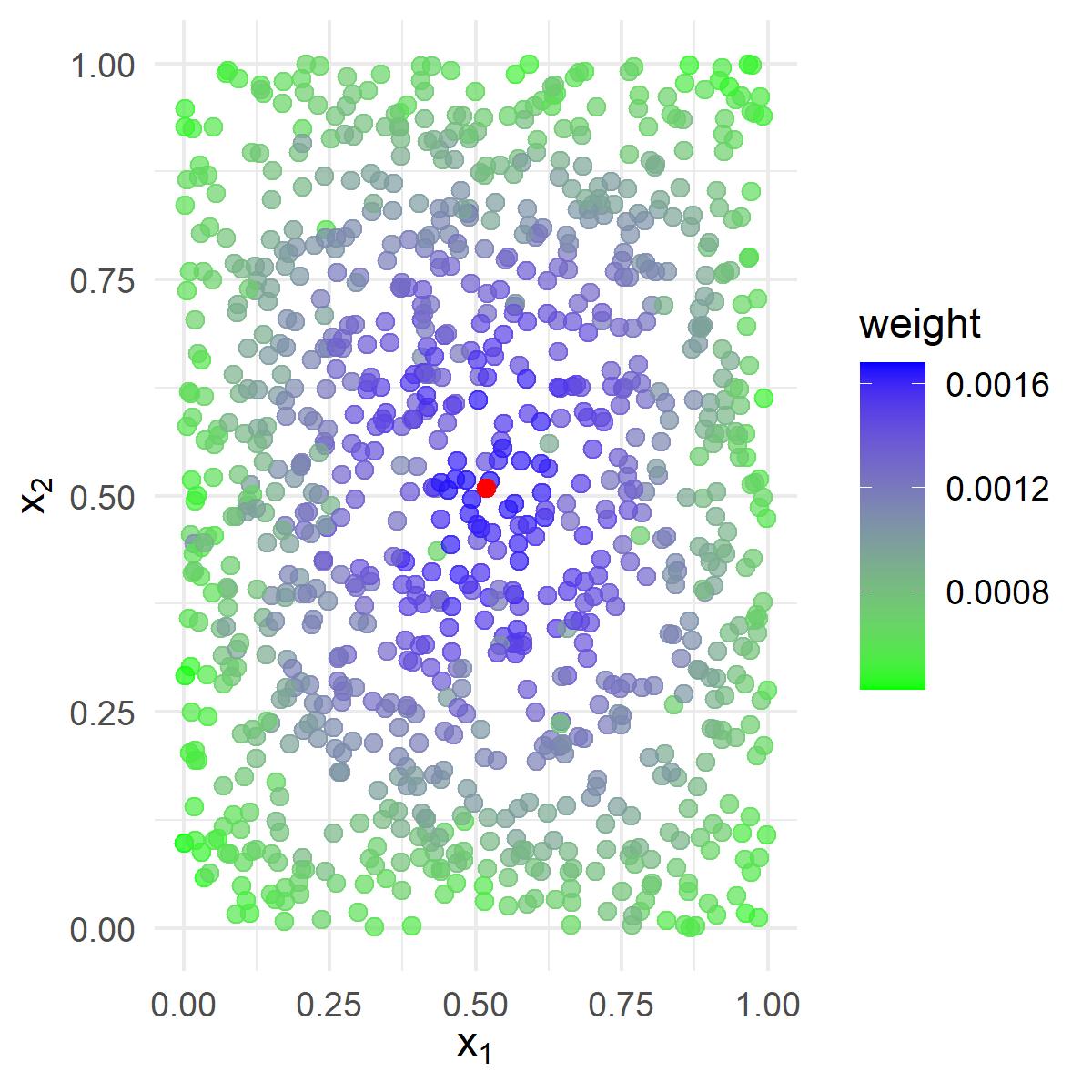}
            \caption[]%
            {{\small Weighted Euclidean distance}}    
        \end{subfigure}
        \hfill
        \begin{subfigure}[b]{0.49\textwidth}  
            \centering 
            \includegraphics[width=\textwidth]{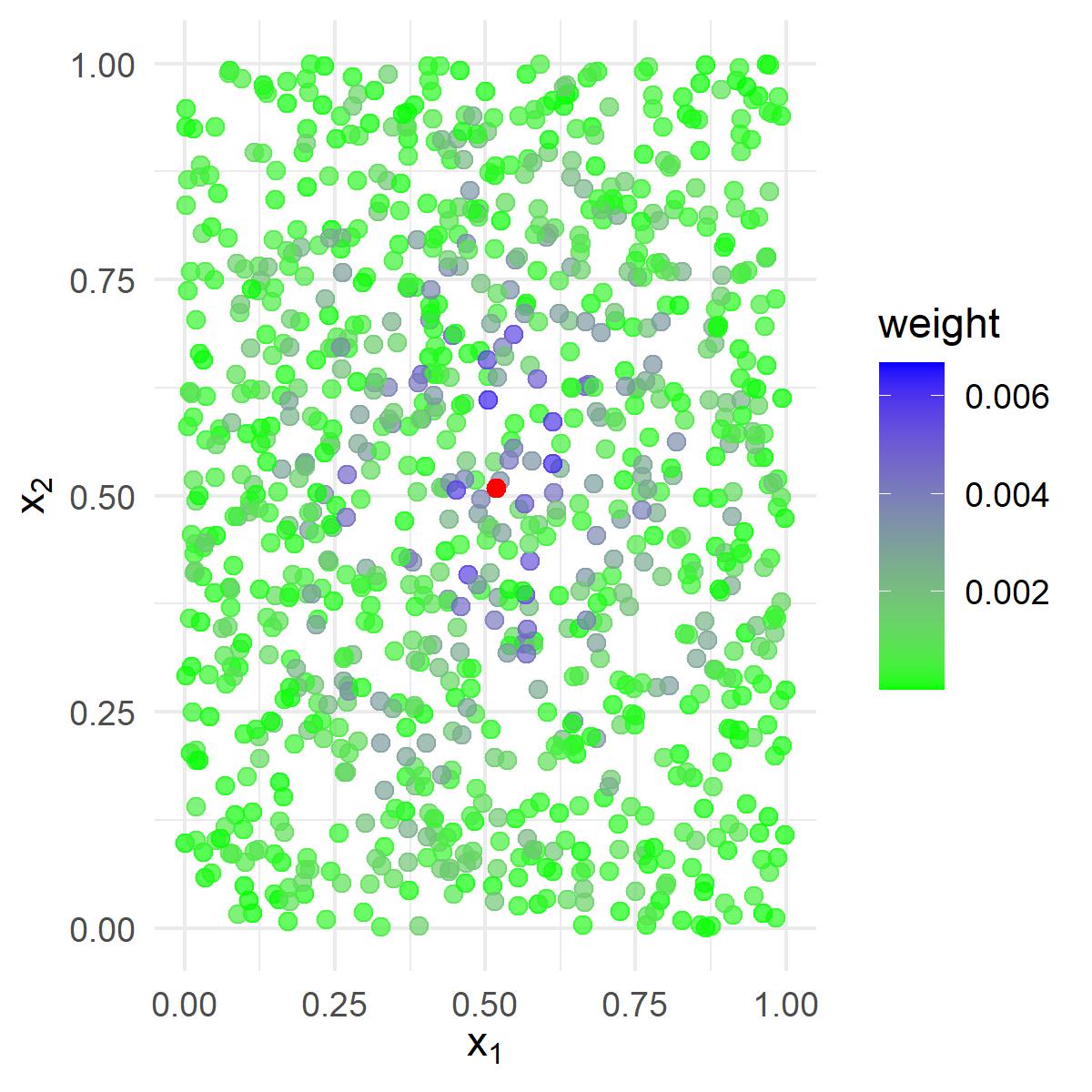}
            \caption[]%
            {{\small Standard Euclidean distance}}    
        \end{subfigure}
        \caption[]
        {\small Comparison of weights $w$ by weighted Euclidean distance and standard Euclidean distance around an instance (red) for artificial dataset: $Y=1+5X_1-5X_2+\epsilon,$ $X_j \sim \text{Uniform}(0,1)$, $\epsilon \sim \text{Normal}(0,1)$ and $d=4$. The weighted Euclidean distance deals better with irrelevant features.} 
        \label{fig:distance}
\end{figure}

The details are as follows. For a specific instance $i$ with a feature vector $\mathbf{x}_i$, we fit the following local model: $$g(\mathbf{x}_i) = \hat{\beta}_{i,0} + \sum_{j=1}^{d}{\hat{\beta}_{i,j} x_{i,j}},$$ where the coefficients $\hat{\beta}_{i,0}, \hat{\beta}_{i,1}, \dots, \hat{\beta}_{i,n}$ are estimated by weighted least squares through $$\operatorname*{arg\,min}_{\beta_{i,0}, \beta_{i,1}, \dots, \beta_{i,n}} \sum_{l=1}^{n} w_l (\mathbf{x_i}) \left( f(\mathbf{x_l} ) - \beta_{i,0} - \sum_{j=1}^{d} \beta_{i,j}x_{l,j} \right)^2.$$  As in LIME, the weights $w_l$ are defined using a Gaussian kernel, i.e., $w_l (\mathbf{x_i}) = \frac{K(\mathbf{x_i}, \mathbf{x_l})}{\sum_{l=1}^{n} K(\mathbf{x_i}, \mathbf{x_l}) }$ with $K(\mathbf{x_i}, \mathbf{x_l}) = \left( \sqrt{2\pi h^2} \right)^{-1} \exp \{ -\frac{\mathrm{d}^2(\mathbf{x_i}, \mathbf{x_l})}{2h^2} \},$ $h>0.$ However, to better deal
with irrelevant variables, and by considering that some variables are more relevant than others, a weighted Euclidean distance $\mathrm{d}^2(\mathbf{x_i}, \mathbf{x_l}) = \sum_{j=1}^{d} v_{i,j} \left( x_{i,j}-x_{l,j} \right)^2$, where $v_{i,j}$ represents the weight of variable $j$ for instance $i$, was used, resulting in more stable weights $w_l$ and more reliable local coefficients. Figure \ref{fig:distance} illustrates the advantage of using weighted Euclidean distance (versus the unweighted one) on an artificial dataset with two relevant and two irrelevant variables.

Weights $v_{i,j}$ are calculated using the \emph{local variable importance} provided by the
Random Forest algorithm \citep{breiman2001random} applied in $\mathbb{D}$. That is, for each tree, we consider the sample not used in its construction (out-of-bag data). The values of the variable $j$ are randomly permuted in each sample. Therefore, only trees that used instance $i$ were considered and the differences in mean square error with and without permutation were calculated. Thus, $v_{i,j} = \frac{I_{i,j}}{\sum_{j} I_{i,j} }$, with $I_{i,j} = \text{MSE}_\text{with permutation} - \text{MSE}$.

Instead of an analysis of coefficients $\beta_{i,j}$, an investigation on the effects of $\phi_{i,j} = \beta_{i,j} x_{i,j}$ may be more intuitive \citep{molnar2018iml}. Since $g(\mathbf{x}_i) = \sum_{j=0}^{d} \hat{\phi}_{i,j}$, each $\hat{\phi}_{i,j}$ is interpreted as the contribution of variable $j$ to prediction $g(\mathbf{x}_i)$. Thus, we use $\phi_{i,j}$ as a measure of importance.

The pseudocode of the VarImp method is described in Algorithm \ref{algo:varimp}. 

\begin{algorithm}
\caption{VarImp}\label{algo:varimp}
\begin{algorithmic}[1]
\Require Data $\mathbb{D}=\{ (\mathbf{x}_1, f(\mathbf{x}_1)), \dots, (\mathbf{x}_n, f(\mathbf{x}_n)) \}$, maximum number of features $M$, kernel bandwidth $h$ (hyperparameter), instance to be explained $i$
\Ensure Local coefficients  $ \hat{\boldsymbol{\beta}}_i$ 
\State $ \mathbf{I} \leftarrow \text{varimp}(\mathbb{D})$ // Matrix with local variable importance 
\State $\mathbf{W}_i \leftarrow \text{ker}(I, \mathbf{X}, h, i)$ // Weights for local linear regression
\State $ \hat{\boldsymbol{\beta}}_i \leftarrow \text{locstep}(\mathbb{D}, \mathbf{W}_i, M)$ // Fitting Local Linear Regression with Forward Stepwise 
\State \textbf{Return} $\hat{\boldsymbol{\beta}}_i$
\end{algorithmic}
\end{algorithm}

\section{SupClus method}\label{sec:supclus}

\begin{figure}[ht]
        \centering
        \begin{subfigure}[b]{0.49\textwidth}
            \centering
            \includegraphics[width=\textwidth]{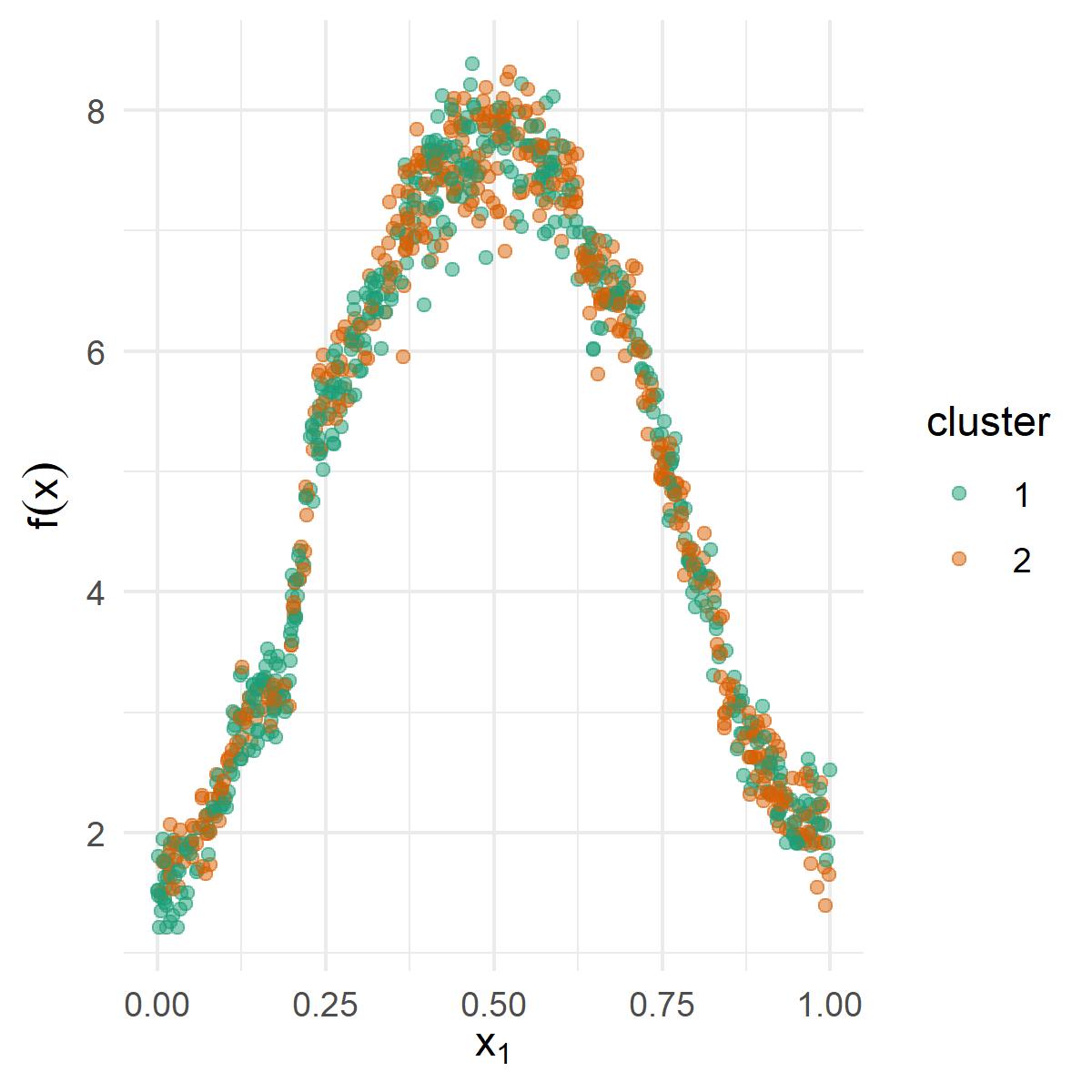}
            \caption[]%
            {{\small Unsupervised cluster}}   
            \label{fig:clustera}
        \end{subfigure}
        \hfill
        \begin{subfigure}[b]{0.49\textwidth}  
            \centering 
            \includegraphics[width=\textwidth]{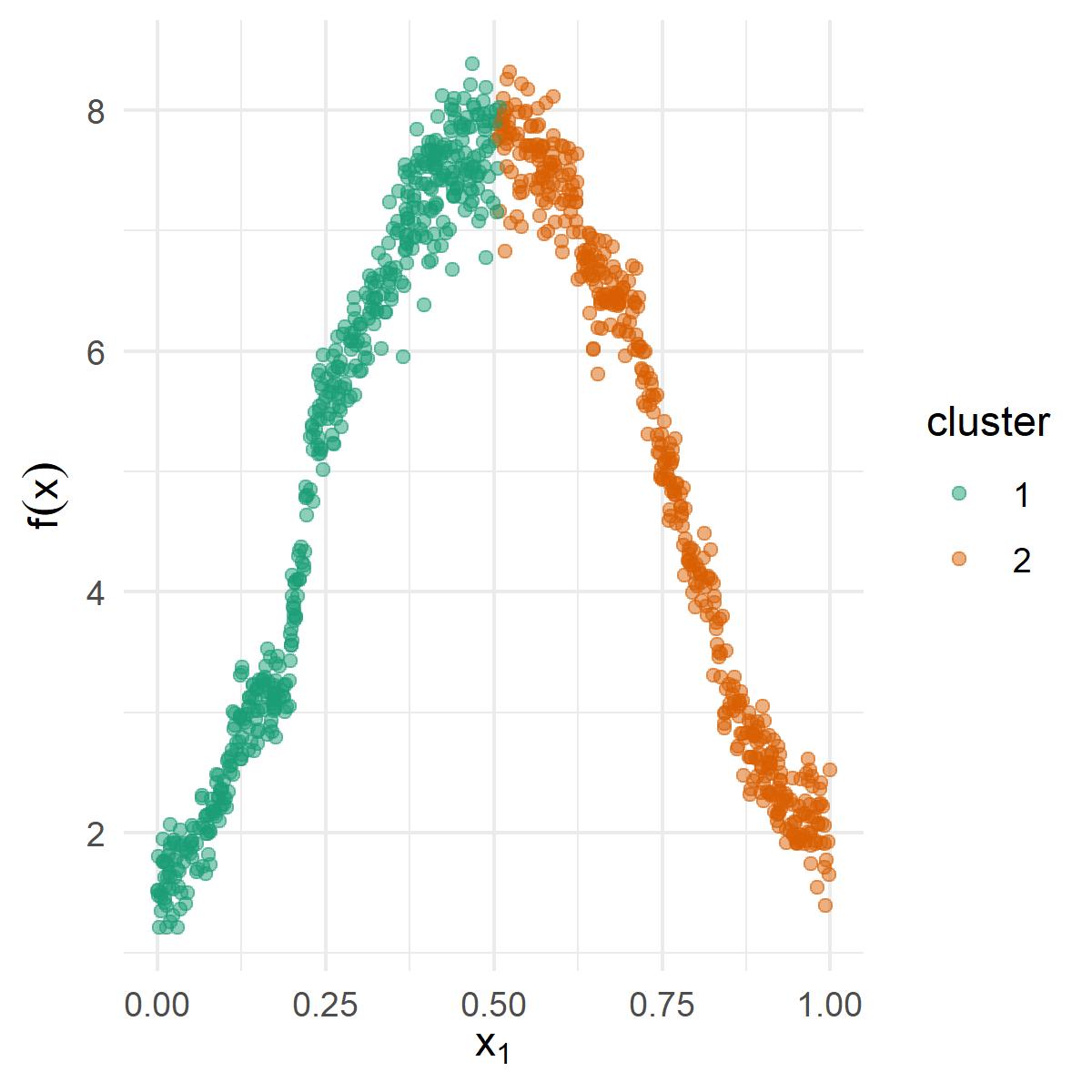}
            \caption[]%
            {{\small Supervised cluster}} 
            \label{fig:clusterb}
        \end{subfigure}
        \caption[]
        {\small Comparison of unsupervised and supervised clusters for artificial dataset: $Y = 20X_1 \mathbf{1}_{X_1 \leq 0.5}(X_1) + (20-20X_1)\mathbf{1}_{X_1 > 0.5}(X_1) + \epsilon$, $X_j \sim \text{Unif}(0,1)$, $\epsilon \sim \text{Normal}(0,1)$ and $d=20$. Supervised cluster groups instances with similar explanations.} 
        \label{fig:cluster}
\end{figure}

SupClus builds on VarImp, and  was designed based on the assumption that some datasets may not need a different interpretation for each instance. Thus, inspired by \cite{hall2017machine,zafar2021deterministic}, we  create clusters of the data, and give the same interpretation to all instances that fall into the same cluster.
To describe how SupClus  works, let us use
a synthetic dataset with $Y = 20X_1 \mathbf{1}_{X_1 \leq 0.5}(X_1) + (20-20X_1)\mathbf{1}_{X_1 > 0.5}(X_1) + \epsilon$, $X_j \sim \text{Unif}(0,1)$, $\epsilon \sim \text{Normal}(0,1)$ and $d=20$ (Figure \ref{fig:cluster}). In this example, it would be sufficient to generate interpretations for only two segments of the data. The standard approach creates clusters using only the features. However, this fails to find clusters with different interpretations (Figure \ref{fig:clustera}, left). To overcome this, SupClus is based on the training of a supervised cluster that considers relationships between variables and predictions in the creation of interpretable clusters (Figure \ref{fig:clusterb}, right).

SupClus works as follows.
 First, we implement VarImp  (Section \ref{sec:varimp}) on $\mathbb{D}$ and collect the matrix of local coefficients $\hat{\mathbf{B}}=\left[ \hat{\boldsymbol{\beta}}_1 \dots \hat{\boldsymbol{\beta}}_n \right]^t$. Then,
 we cluster the instances using the representation $\hat{\mathbf{B}}$. By construction, instances that fall into the same cluster will share a similar interpretation. In order to allow for general shapes for the clusters, we use  Spectral Cluster Analysis (the version proposed by \cite{ng2001spectral}). Next, we train a different linear model  for each cluster using all instances that fall on it. These models use the output $f( \mathbf{x} )$ as the target.  Again, a Forward Stepwise variable selection algorithm with a maximum number of variables $M \leq d$ is used in each cluster. The interpretation for each cluster is given by the coefficients of such model. In order to define the optimal number of clusters, we use an adjusted $R^2$ statistic, weighted by the size of the clusters: $R^2 = \frac{1}{n} \sum_{l=1}^{k} \lvert c_l \rvert R_l^2$, where $R_l^2 = 1 - \frac{(\lvert c_l \rvert - 1) \sum_{i \in c_l } \left( f( \mathbf{x}_i) - g(\mathbf{x}_i) \right)^2 }{(\lvert c_l \rvert - M) \sum_{i \in c_l} \left( f( \mathbf{x}_i) - \overline{f_l( \mathbf{x})} \right)^2 }$ and $\overline{f_l( \mathbf{x})} = \sum_{i \in c_l} f( \mathbf{x}_i) / \lvert c_l \rvert $.
Algorithm \ref{algo:supclus} shows the pseudocode for SupClus.

\begin{algorithm}
\caption{SupClus}\label{algo:supclus}
\begin{algorithmic}[1]
\Require Data $\mathbb{D}=\{ (\mathbf{x}_1, f(\mathbf{x}_1)), \dots, (\mathbf{x}_n, f(\mathbf{x}_n)) \}$, maximum number of features $M$, local coefficient matrix $\hat{\mathbf{B}}$ from VarImp, number of clusters $k$ (hyperparameter), instance to be explained $i$
\Ensure Local coefficients $ \hat{\boldsymbol{\gamma}}_i$ and weighted coefficient of determination $r_k$ 
\State $\mathbf{c} \leftarrow \text{specc}(\hat{\mathbf{B}}, M, k)$ // Cluster using local VarImp coefficients
\State $\hat{\mathbf{A}} = \left[ \hat{\boldsymbol{\alpha}}_1 \dots \hat{\boldsymbol{\alpha}}_M \right]^t$ e $r_k \leftarrow \text{regcluster}(\mathbf{c}, \mathbb{D})$ // Local linear regression fit with Forward Stepwise for each cluster and $R^2$ weighted by the clusters size
\State $\hat{\boldsymbol{\gamma}}_i = \hat{\boldsymbol{\alpha}}_l:i \in c_l$ // Finds coefficients of the cluster to which $i$ belongs 
\State \textbf{Return} $\hat{\boldsymbol{\gamma}}_i$
\end{algorithmic}
\end{algorithm}

\section{Experimental design}\label{sec:experiments}

We apply VarImp and SupClus both to  artificial datasets, where the true local coefficients are known, and to real datasets. To assess their performance, they were compared with LIME \citep{ribeiro2016should}, Shapley values \citep{shapley1953value, vstrumbelj2014explaining, lundberg2017unified}, and IML \citep{molnar2018iml}, a version of LIME with no perturbed samples.

First, the model to  be interpreted was trained with the use of a sample $\mathbb{D}'$ in all experiments. Due to its easy and fast
training the Random Forest algorithm with default hyperparameter values was used for the evaluation of the methods. However, since the methods used are agnostic, any other supervised algorithm could have been used. For the comparison, the interpretability methods were applied to dataset $\mathbb{D}$ using the predictions from the original model as target.

Six artificial datasets (Table \ref{tab:simdata}) were created, each trying to reflect what could be observed on  real datasets
(e.g., data with irrelevant attributes or data with non-linear relationships between features and target). In all cases, $d=20$ and an error $\epsilon \sim \text{Normal}(0,1)$ was considered. The proposed methods were also applied to five real datasets (Table \ref{tab:realdata}) from 
different application domains:
health, education, economics, mobility, and ecology. Dataset 9 involves a classification problem where methods will be used to explain the predicted probability of heart disease.

\begin{table}[ht]
\caption{Artificial datasets representing a variety of real datasets.}
\label{tab:simdata}
\resizebox{\columnwidth}{!}{%
\begin{tabular}{@{}lll@{}}
\toprule
\multicolumn{1}{c}{\textbf{Dataset}}                      & \multicolumn{1}{c}{\textbf{Features}} & \multicolumn{1}{c}{\textbf{Target}}                                                       \\ \midrule
Linear regression w/ irrelevant features (1)              & $X_j \sim \text{Unif}(0,1)$           & $Y=1+5X_1-5X_2+\epsilon$                                                                  \\
                                                          &                                       &                                                                                           \\
Linear regressions combination w/ irrelevant features (2) & $X_j \sim \text{Unif}(0,1)$           & $Y = 20X_1 \mathbf{1}_{X_1 \leq 0.5}(X_1) + (20-20X_1)\mathbf{1}_{X_1 > 0.5}(X_1) + \epsilon$ \\
                                                          &                                       &                                                                                           \\
Relevant continuous features (3)                          & $X_j \sim \text{Unif}(0,1)$           & $Y = \sum_{j=1}^{d} (-1)^{j+1} X_j + \epsilon$                                            \\
                                                          &                                       &                                                                                           \\
Relevant binary features (4)                              & $X_j \sim \text{Ber}(1/2)$            & $Y = \sum_{j=1}^{d} (-1)^{j+1} X_j + \epsilon$                                            \\
                                                          &                                       &                                                                                           \\
Step function w/ irrelevant features (5)                  & $X_j \sim \text{Unif}(0,1)$           & $Y = 20X_1 \mathbf{1}_{ 1/3 \leq X_1 \leq 2/3}(X_1) + \epsilon$                             \\
                                                          &                                       &                                                                                           \\
Sine function w/ irrelevant features (6)                  & $X_j \sim \text{Unif}(0,1)$           & $Y = 20 \sin(2 \pi X_1) + \epsilon$                                                                  \\ \bottomrule
\end{tabular}%
}
\end{table}

\begin{table}[ht]
\caption{Real datasets from different application domains: health, education, economics, mobility, and ecology.}
\label{tab:realdata}
\resizebox{\columnwidth}{!}{%
\begin{tabular}{ll}
\hline
\multicolumn{1}{c}{\textbf{Dataset}} & \multicolumn{1}{c}{\textbf{Description}}                                                                                                                                                                                                                                        \\ \hline
Bike rental (1)                      & \begin{tabular}[c]{@{}l@{}}Dataset with daily bike rental counts (target) in Washington DC. \\ Features: temperature, humidity, wind speed, etc. \\ Source: http://archive.ics.uci.edu/ml/datasets/Bike+Sharing+Dataset.\end{tabular}                                           \\
                                     &                                                                                                                                                                                                                                                                                 \\
Ecology (2)                          & \begin{tabular}[c]{@{}l@{}}Above-ground biomass (target) in Australia. \\ Features: temperature, precipitation, soil density, pH, etc. \\ Source: https://datadryad.org/stash/dataset/doi:10.5061/dryad.dr7sqv9w9.\end{tabular}                                                 \\
                                     &                                                                                                                                                                                                                                                                                 \\
Heart diseases (3)                   & \begin{tabular}[c]{@{}l@{}}Data on heart diseases (target) and risk factors in the USA. \\ Features: age, body mass index, other diseases, etc. \\ Source: https://www.kaggle.com/datasets/kamilpytlak/personal-key-indicators-of-heart-disease?resource=download.\end{tabular} \\
                                     &                                                                                                                                                                                                                                                                                 \\
HDI (4)                              & \begin{tabular}[c]{@{}l@{}}Data with Human Development Index (target) from 5565 Brazilian cities.\\ Features: information on income, education, and health. \\ Source: https://www.ipea.gov.br/ipeageo/arquivos/bases.\end{tabular}                                             \\
                                     &                                                                                                                                                                                                                                                                                 \\
Student´s performance (5)            & \begin{tabular}[c]{@{}l@{}}Data on performance (target) of secondary students in Portugal.\\ Features: number of absences, previous grades, age, hours of study, travel time, etc.\\ Source: https://archive.ics.uci.edu/ml/datasets/Student+Performance.\end{tabular}          \\ \hline
\end{tabular}%
}
\end{table}

\begin{figure}[ht]
        \centering
        \begin{subfigure}[b]{0.32\textwidth}
            \centering
            \includegraphics[width=\textwidth]{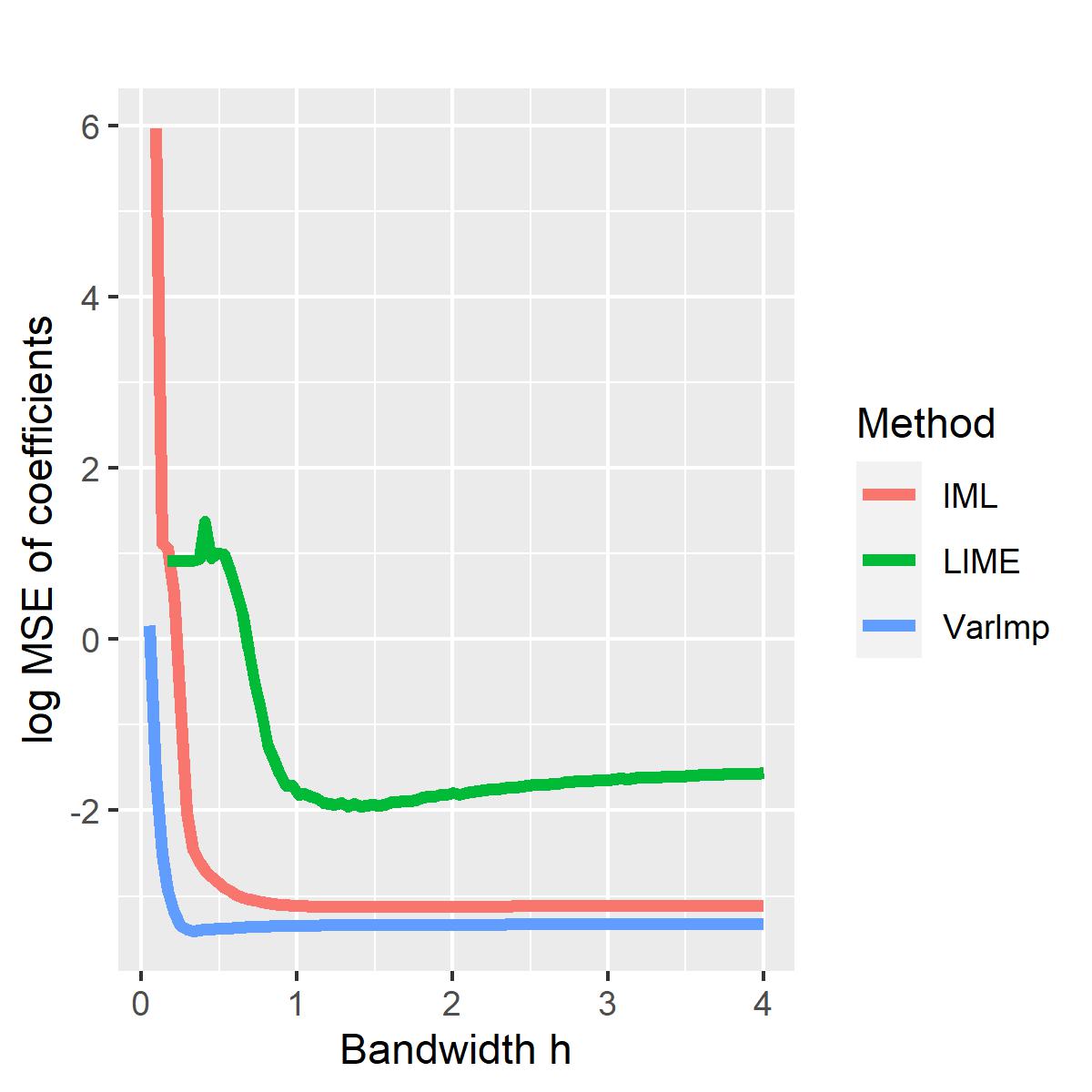}
            \caption[]%
            {{\small Dataset 1}}   
        \end{subfigure}
        \begin{subfigure}[b]{0.32\textwidth}  
            \centering 
            \includegraphics[width=\textwidth]{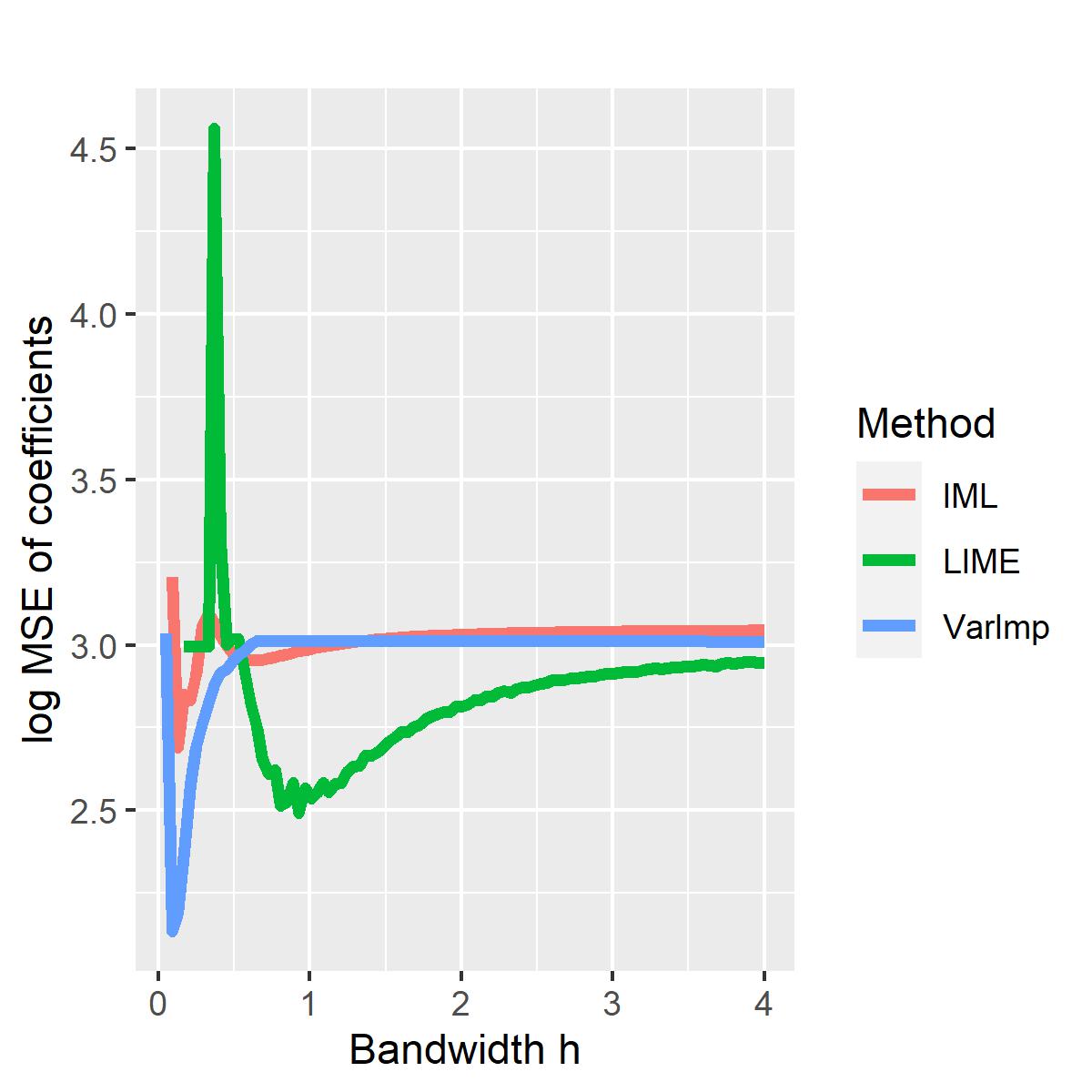}
            \caption[]%
            {{\small Dataset 2}} 
        \end{subfigure}
        \begin{subfigure}[b]{0.32\textwidth}  
            \centering 
            \includegraphics[width=\textwidth]{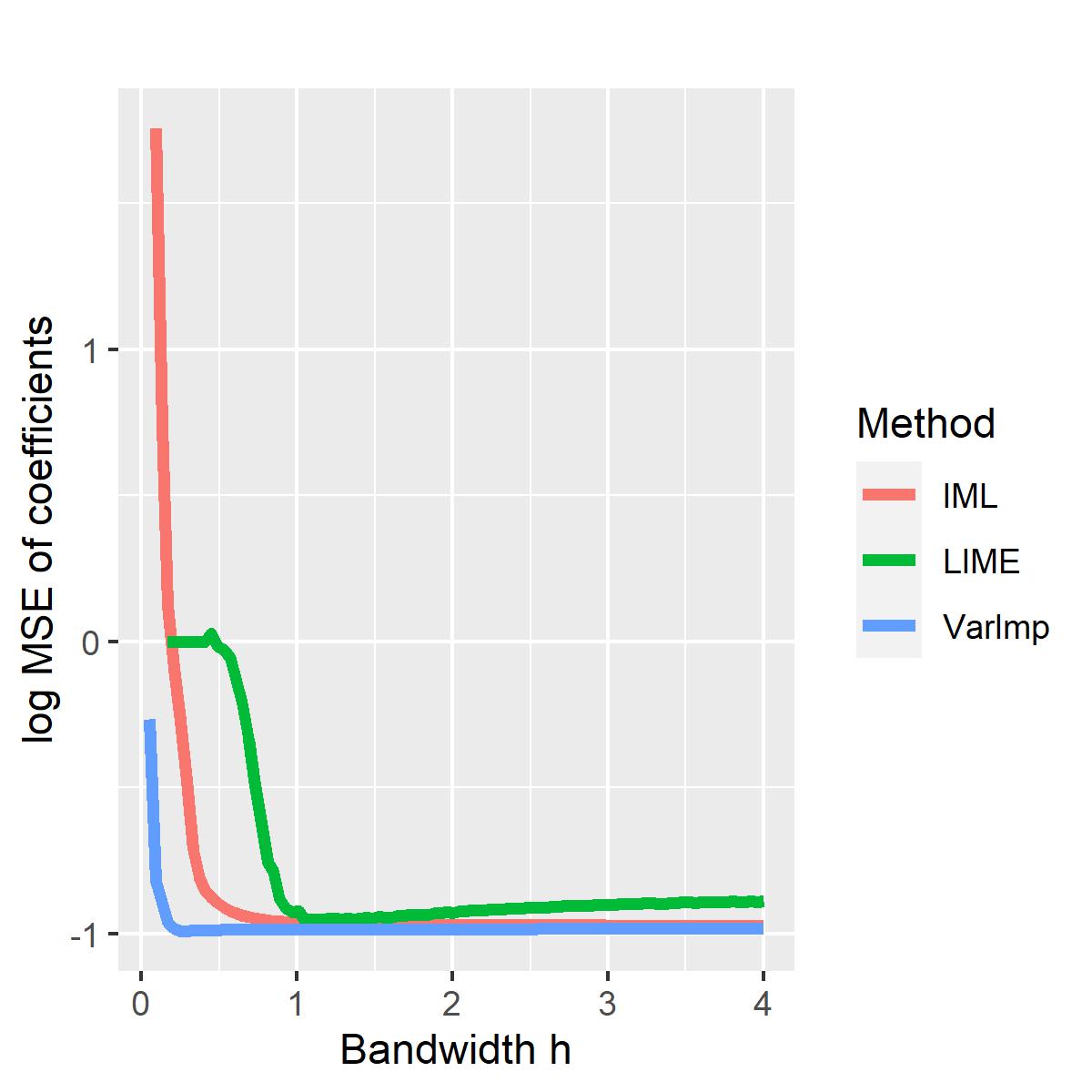}
            \caption[]%
            {{\small Dataset 3}} 
        \end{subfigure}
        \hfill
        \begin{subfigure}[b]{0.32\textwidth}  
            \centering 
            \includegraphics[width=\textwidth]{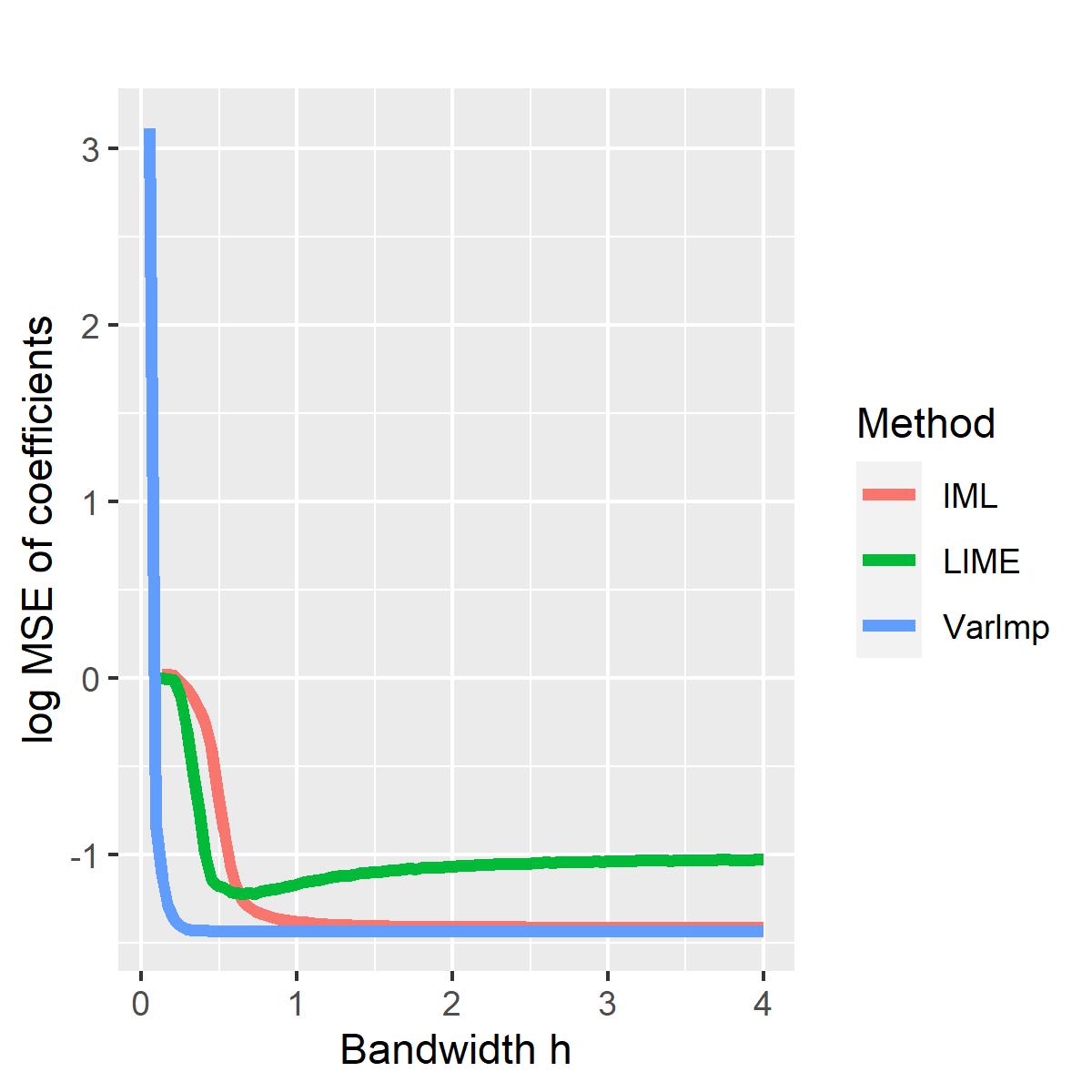}
            \caption[]%
            {{\small Dataset 4}} 
        \end{subfigure}
        \begin{subfigure}[b]{0.32\textwidth}  
            \centering 
            \includegraphics[width=\textwidth]{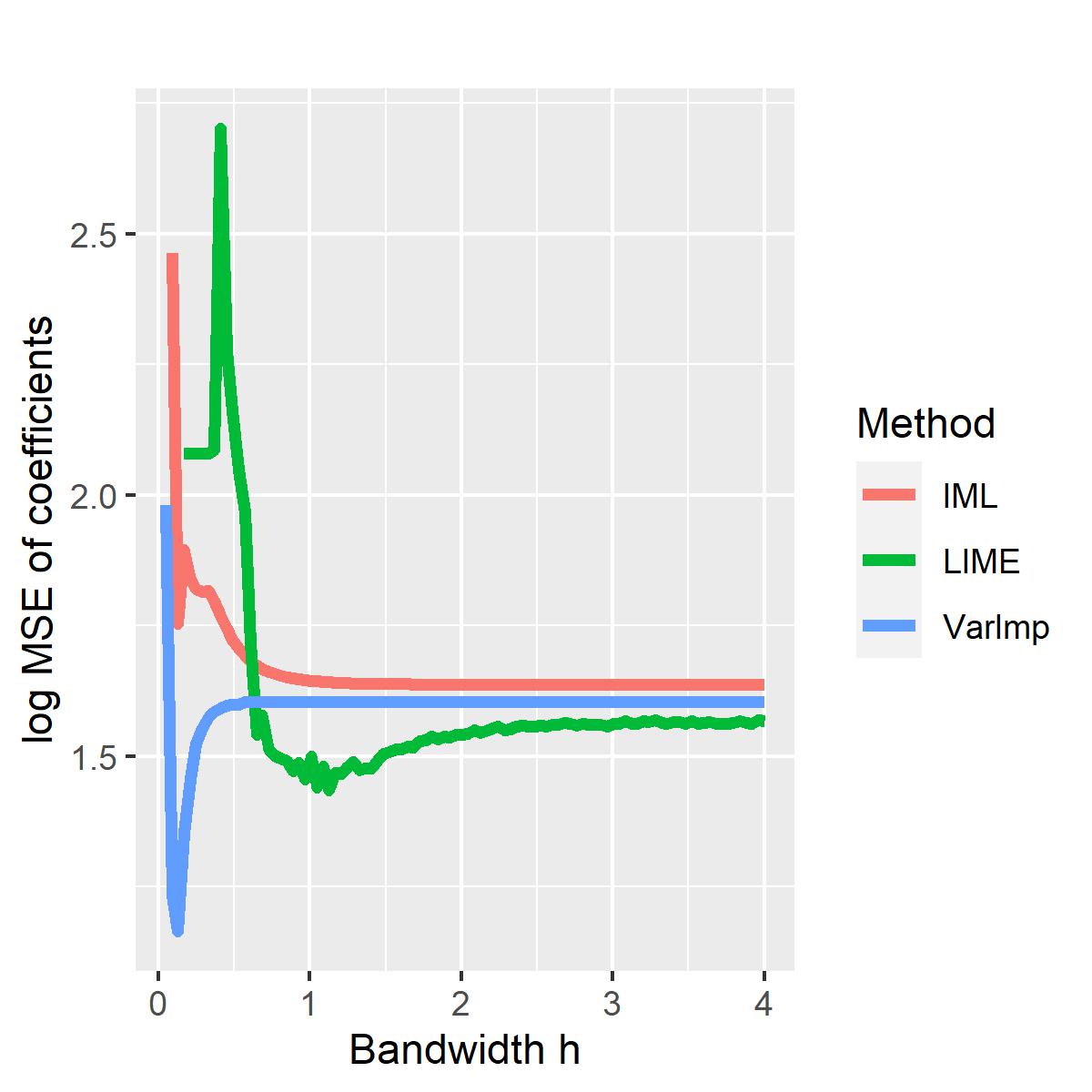} 
            \caption[]%
            {{\small Dataset 5}} 
        \end{subfigure}
            \begin{subfigure}[b]{0.32\textwidth}  
            \centering 
            \includegraphics[width=\textwidth]{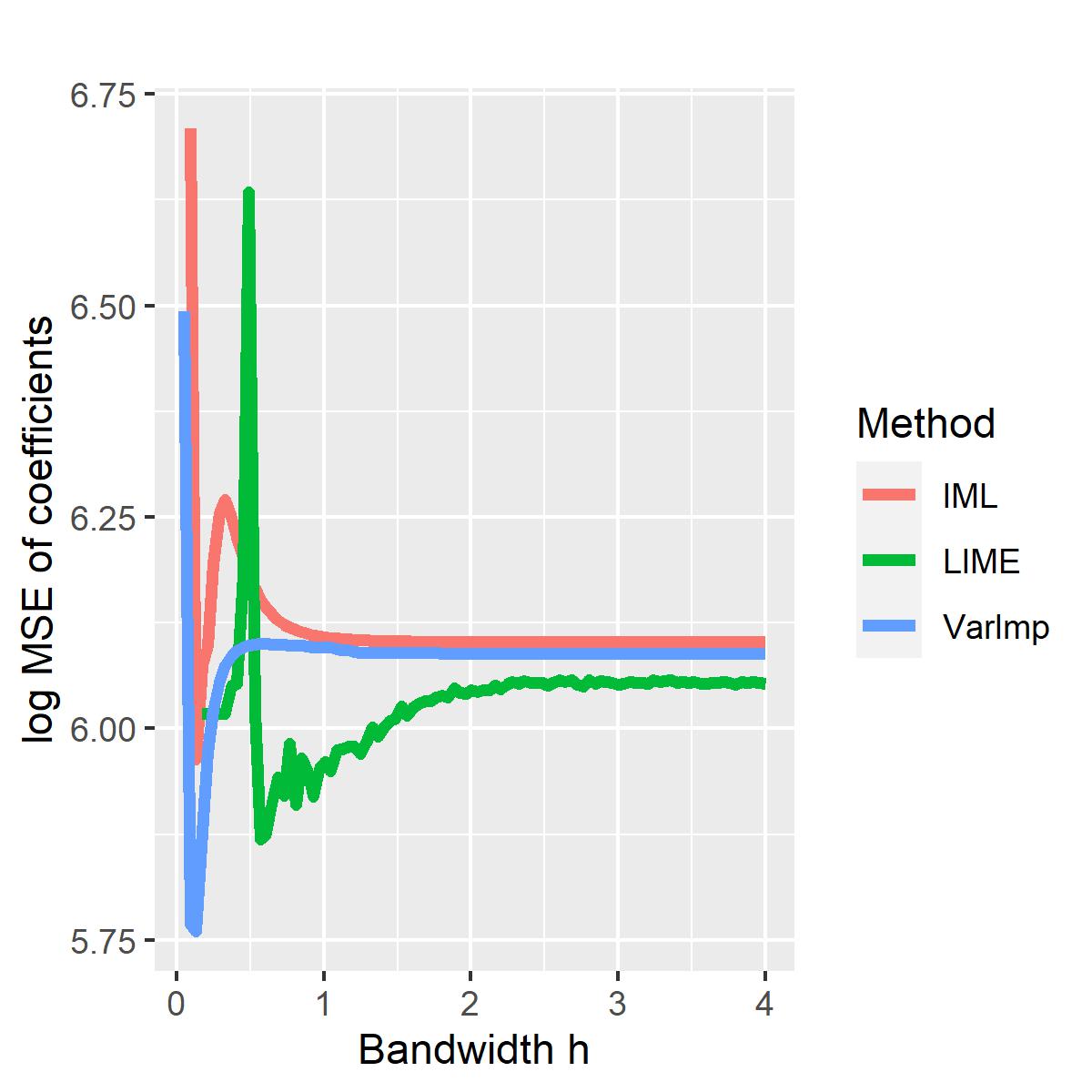} 
            \caption[]%
            {{\small Dataset 6}} 
        \end{subfigure}
        \caption[]
        {\small Influence of bandwidth $h$ in MSE of coefficients for artificial datasets. VarImp has smaller errors compared to LIME and IML even considering different values of bandwidth $h$.} 
        \label{fig:mse_bandwidth}
\end{figure}

In the VarImp, LIME and IML methods, a bandwidth $h$ between $0.05$ and $4.00$ was used. These values were obtained by analyzing local slope plots and mean square error of coefficients (Subsection \ref{sec:measures}). When true coefficients are not known we use coefficients obtained by the ICE method \citep{goldstein2015peeking}. Figure \ref{fig:mse_bandwidth} shows that VarImp has smaller errors compared to LIME and IML even considering different values of bandwidth $h$. The number of clusters $k$ in SupClus varied between $1$ and $5$ and the cluster with highest $R^2$ was chosen.
All available features were analyzed for
all methods, with $M=d$, for comparisons with Shapley values. For the experiments with the real data sets, categorical variables were transformed into indicators and continuous variables were rescaled between $0$ and $1$ by min-max scaling. LIME, IML and Shapley values were applied using packages available in R \citep{molnar2018, Pedersen2021}.

\subsection{Evaluation measures}\label{sec:measures}

Measures for evaluating interpretability methods that involve surveys have the limitation of being subjective in addition to using only a few points from a sample. Other measures involve aspects such as complexity and stability and do not directly assess the quality of explanations. This section presents measures for evaluations and comparisons of the quality of interpretations and the predictive performance of the proposed methods. The measurements take into account all points in a sample $\mathbb{D}$. 


\textbf{Mean square error of coefficients.} For artificial data whose values of local coefficients are known, we compute the mean squared error between real and estimated coefficients, $\text{MSE}_{\mathbf{B},\hat{\mathbf{B}}} = \frac{1}{nd} \sum_{l=1}^n \sum_{j=1}^d \left( \beta_{l,j}-\hat{\beta}_{l,j} \right)^2$. This measure directly assesses the quality of explanations through local coefficients and therefore cannot be used for Shapley values whose local coefficients are not estimated.

\textbf{Effect correlation.} Still for artificial data, we compute the Pearson correlation between real and estimated effects, $r_{\boldsymbol{\phi}, \hat{\boldsymbol{\phi}}}$. This measure assesses the quality of explanations through local effects and
can be applied to Shapley values.

\textbf{Prediction correlation.} We assess the predictive performance of the explanations using the
correlations between predictions from the original model and the local model, $r_{f,g}$. This measure evaluates the fidelity of the local model as an approximator of 
the original prediction algorithm $f$. 

\textbf{ICE effect correlation.} The Individual Conditional Expectation method \citep{goldstein2015peeking}  generates visualizations of the curves $g_{i,j}(x) = f(x_{i,1} +, \dots, x_{i,j-1}, x, x_{i,j+1}, \dots, x_{i,d})$ for a feature $j$ and instance $i$. Derivatives and their effects were obtained by this method and the Pearson correlation between ICE effects and the effects of an interpretability method, $r_{\boldsymbol{\phi}_{\text{ICE}}, \hat{\boldsymbol{\phi}}}$, were calculated. This measure assesses the quality of explanations through local effects and  is especially useful in real data where true effects are not known.

\textbf{Local slope plot.} 
In order to check how well the local coefficients fit the ML algorithm $f$, we add line segments to the scatter plot of the predictions $f(\mathbf{x})$ versus each feature $x_l$. These segments have slope given by the local coefficient associated to $x_l$.

\section{Experimental results}\label{sec:results}

\subsection{Artificial data}

\begin{figure}[ht]
        \centering
        \begin{subfigure}[b]{0.49\textwidth}
            \centering
            \includegraphics[width=\textwidth]{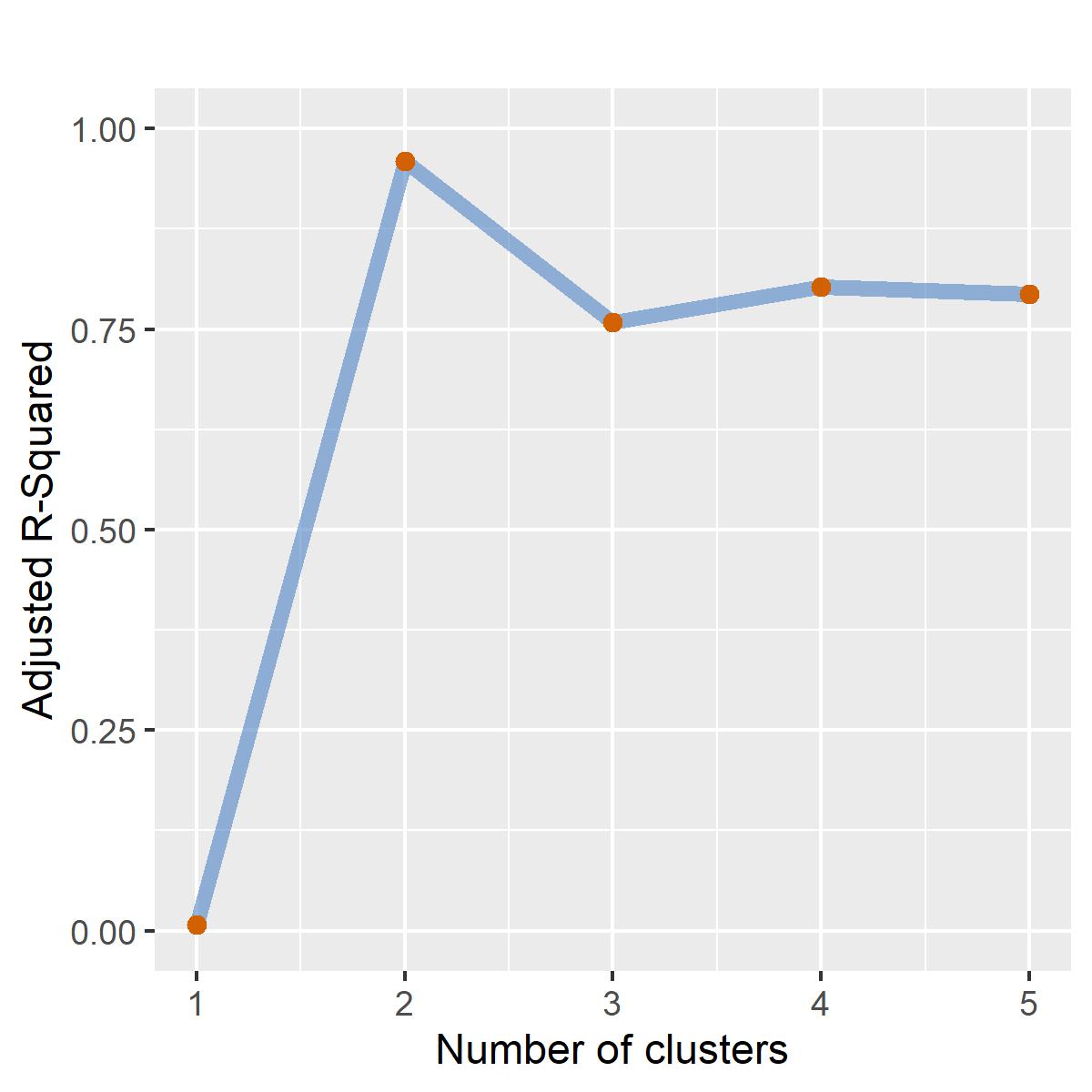}
            \caption[]%
            {{\small Dataset 2}}   
            \label{fig:r2a}
        \end{subfigure}
        \hfill
        \begin{subfigure}[b]{0.49\textwidth}  
            \centering 
            \includegraphics[width=\textwidth]{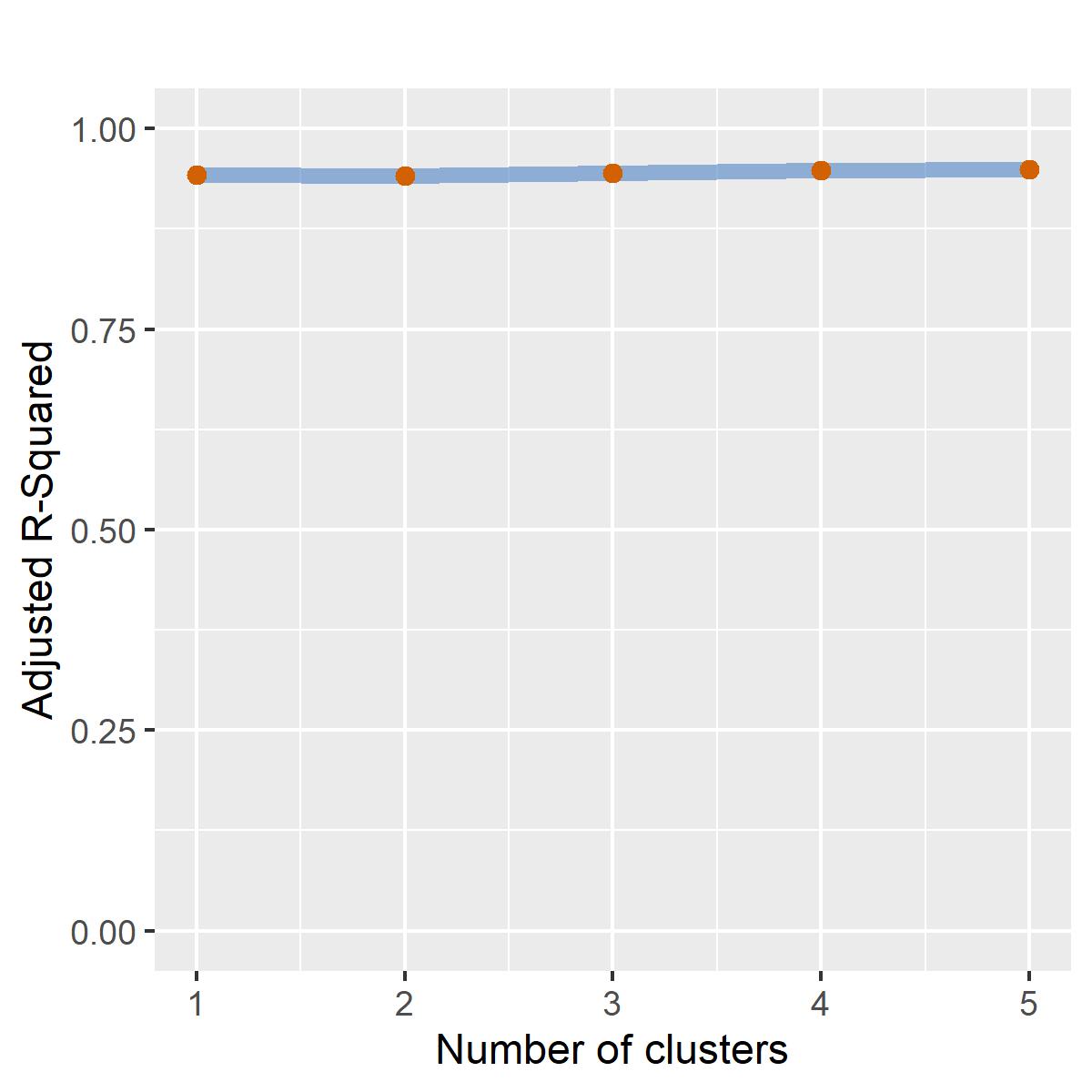}
            \caption[]%
            {{\small Dataset 4}} 
            \label{fig:r2b}
        \end{subfigure}
        \caption[]
        {\small Coefficient of determination with a variable number of clusters for datasets 2 and 4. Dataset 2 can be interpreted with only 2 clusters and dataset 4 can be interpreted with only 1 cluster.} 
        \label{fig:r2}
\end{figure}

\textbf{Number of clusters.} As previously mentioned, the main advantage of SupClus when compared with the other methods is that
it can identify simpler datasets whose
instances can be interpreted by using clusters. As an
example, datasets 1, 3, and 4 can be interpreted with only 1 cluster and dataset 2 (or dataset 5) can be interpreted with only 2 or 3 clusters (Figure \ref{fig:clusterb}). On the other hand, dataset 6 requires explanations for each instance due to its complexity. A weighted coefficient of determination obtained the optimal number of clusters in each dataset (Figure \ref{fig:r2}).

\textbf{Mean square error of coefficients.} Since the true local coefficients are known for those datasets, the proposed methods can be more accurately compared with other available methods. Table \ref{tab:simmse} shows that, in general, VarImp presented lower MSE values,
especially for more complex datasets, such as dataset 6. All methods have low errors in datasets 3 and 4, where all variables are relevant and the relationships of $X_1$ with the target are linear. SupClus clearly has the smallest coefficient error in datasets 2 and 5, which contain well-defined clusters and irrelevant variables. LIME and IML have higher errors for datasets with irrelevant variables and when the relationships are not linear. LIME has the largest error in dataset 1. We conclude that our methods do a good job in approximating true local slopes.

\begin{table}[ht]
\caption{Mean square error of coefficients for artificial data. VarImp presented lower MSE values for more complex datasets. SupClus has the smallest coefficient error in dataset 2, which contains well-defined clusters and irrelevant variables.}
\label{tab:simmse}
\resizebox{\columnwidth}{!}{%
\begin{tabular}{lrrrr}
\hline
                                                          & \multicolumn{4}{c}{$\textbf{MSE}_{\mathbf{B},\hat{\mathbf{B}}}$}                                                                                                  \\ \cline{2-5} 
\multicolumn{1}{c}{\textbf{Dataset}}                      & \multicolumn{1}{c}{\textbf{VarImp}} & \multicolumn{1}{c}{\textbf{SupClus}} & \multicolumn{1}{c}{\textbf{LIME}} & \multicolumn{1}{c}{\textbf{IML}} \\ \hline
Linear regression w/ irrelevant features (1)              & \textbf{0.085}                               & 0.094                                & 0.229                             & 0.093                            \\
Linear regressions combination w/ irrelevant features (2) & 4.484                               & \textbf{1.862}                                & 13.303                            & 14.851                           \\
Relevant continuous features (3)                          & \textbf{0.412}                               & 0.414                                & 0.429                             & 0.417                            \\
Relevant binary features (4)                              & 0.290                               & \textbf{0.289}                                & 0.347                             & 0.293                            \\
Step function w/ irrelevant features (5)                  & 1.609                               & \textbf{0.923}                                & 4.236                             & 4.735                            \\
Sine function w/ irrelevant features (6)                  & \textbf{61.726}                              & 186.507                              & 376.778                           & 411.008                          \\ \hline
\end{tabular}%
}
\end{table}

\begin{figure}[ht]
        \centering
        \begin{subfigure}[b]{0.49\textwidth}
            \centering
            \includegraphics[width=\textwidth]{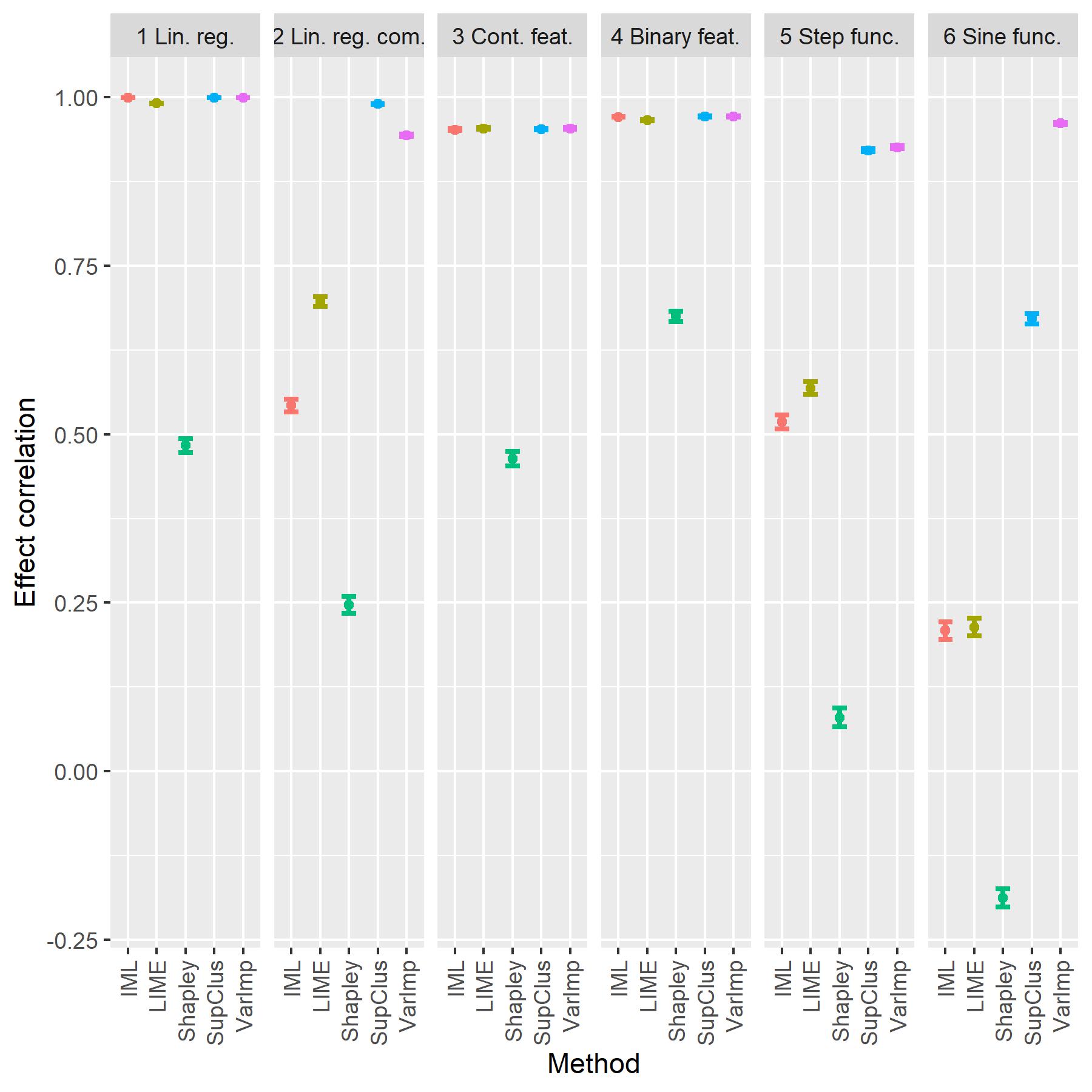}
            \caption[]%
            {{\small Effect correlation}}   
            \label{fig:simcora}
        \end{subfigure}
        \hfill
        \begin{subfigure}[b]{0.49\textwidth}  
            \centering 
            \includegraphics[width=\textwidth]{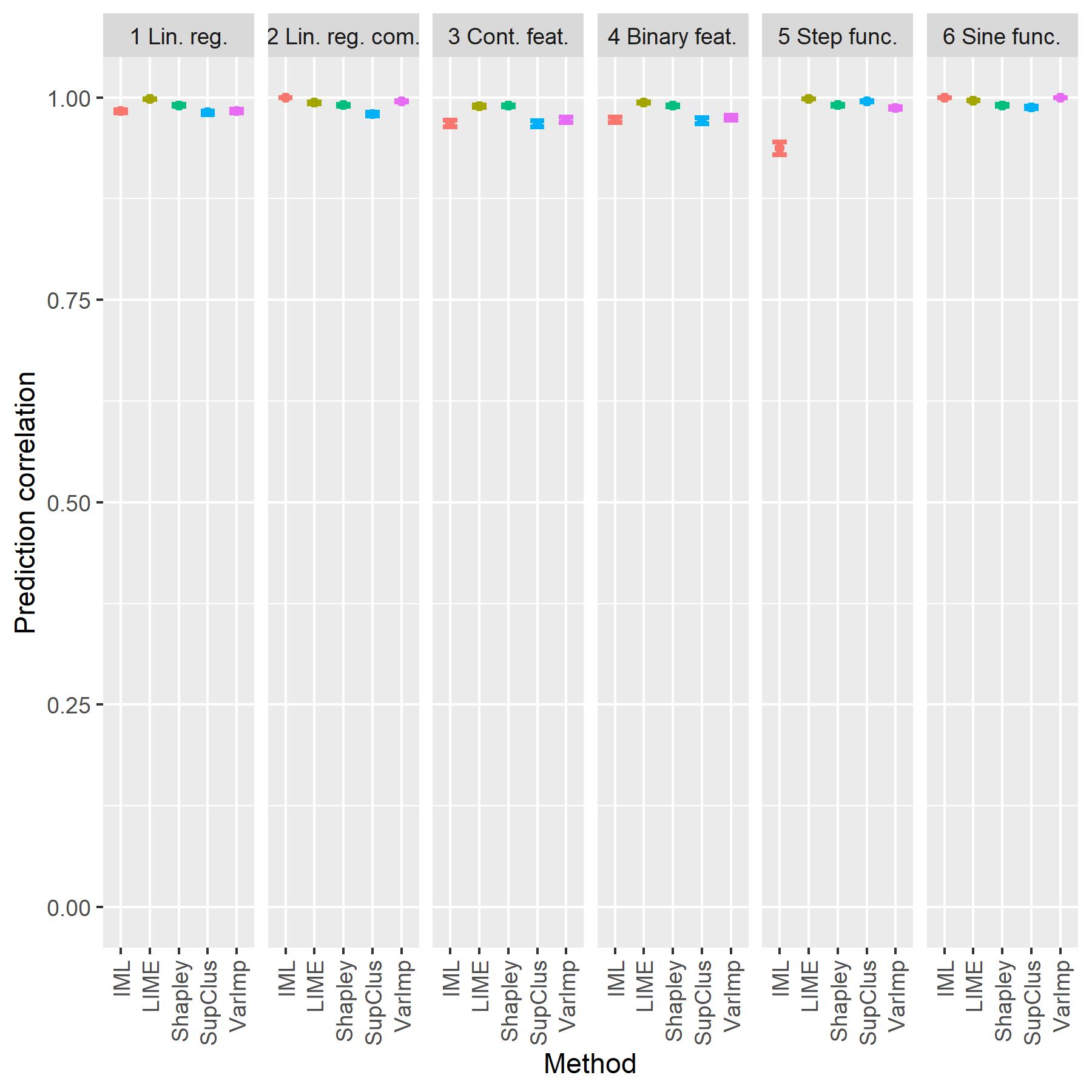}
            \caption[]%
            {{\small Prediction correlation}} 
            \label{fig:simcorb}
        \end{subfigure}
        \caption[]
        {\small Correlations of effects and predictions for artificial data with $95\%$ confidence level. VarImp and SupClus provide better or equal measures than other methods.} 
        \label{fig:simcor}
\end{figure}

\textbf{Effect and prediction correlation.} Another way of comparing interpretation methods in artificial data is through \emph{effect correlation} from subsection \ref{sec:measures} (Figure \ref{fig:simcora}). All conclusions with MSE are reinforced with this measure. Shapley value does not show good correlations of effects in most datasets, although it yields a good indicator of which variables are the most important (regardless of the sign of the relationship).
All methods show good correlations between local and global predictions (Figure \ref{fig:simcorb}). 

\begin{figure}[ht]
        \centering
        \begin{subfigure}[b]{0.32\textwidth}
            \centering
            \includegraphics[width=\textwidth]{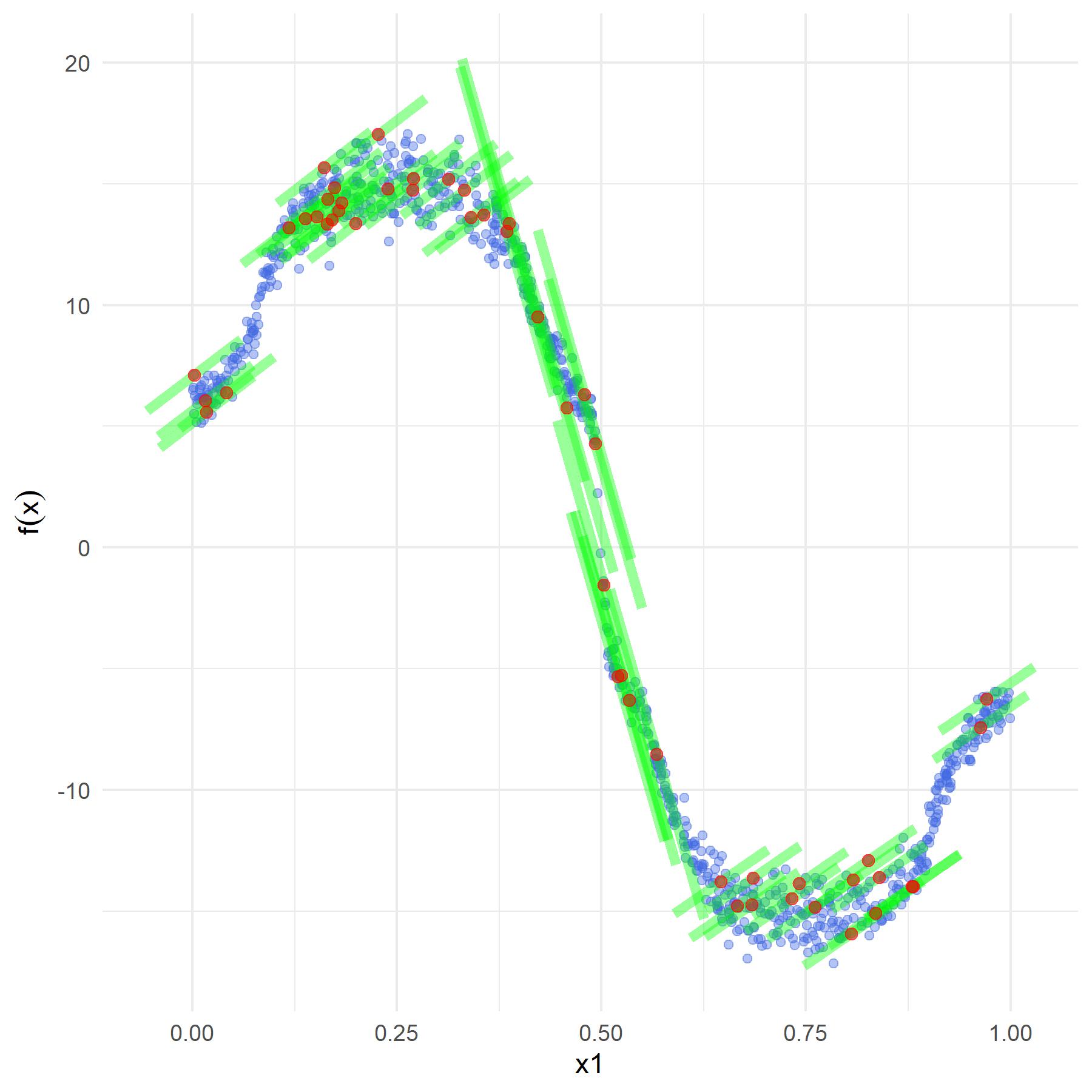}
            \caption[]%
            {{\small SupClus}}   
            \label{fig:simslopea}
        \end{subfigure}
        \begin{subfigure}[b]{0.32\textwidth}  
            \centering 
            \includegraphics[width=\textwidth]{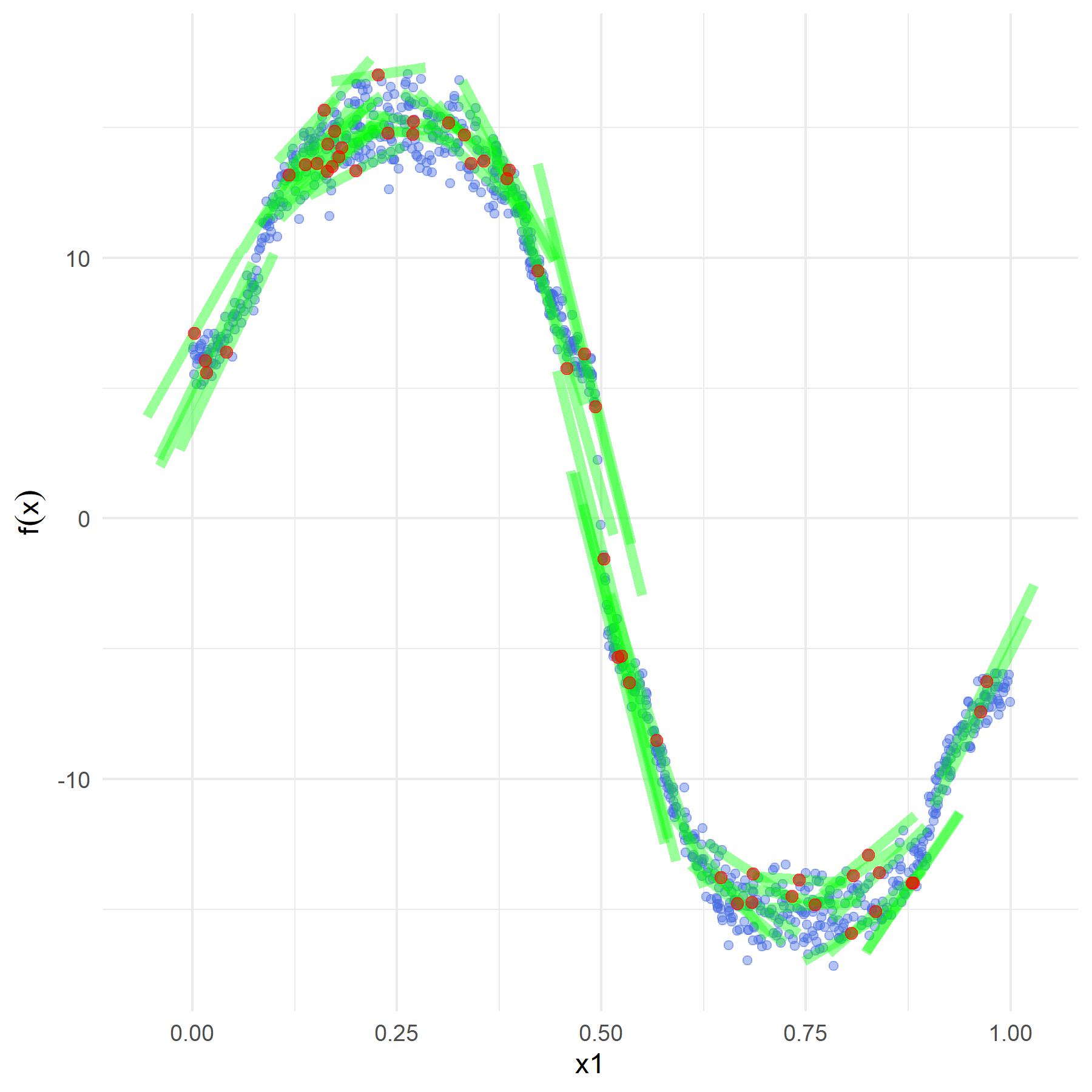}
            \caption[]%
            {{\small VarImp}} 
            \label{fig:simslopeb}
        \end{subfigure}
        \begin{subfigure}[b]{0.32\textwidth}  
            \centering 
            \includegraphics[width=\textwidth]{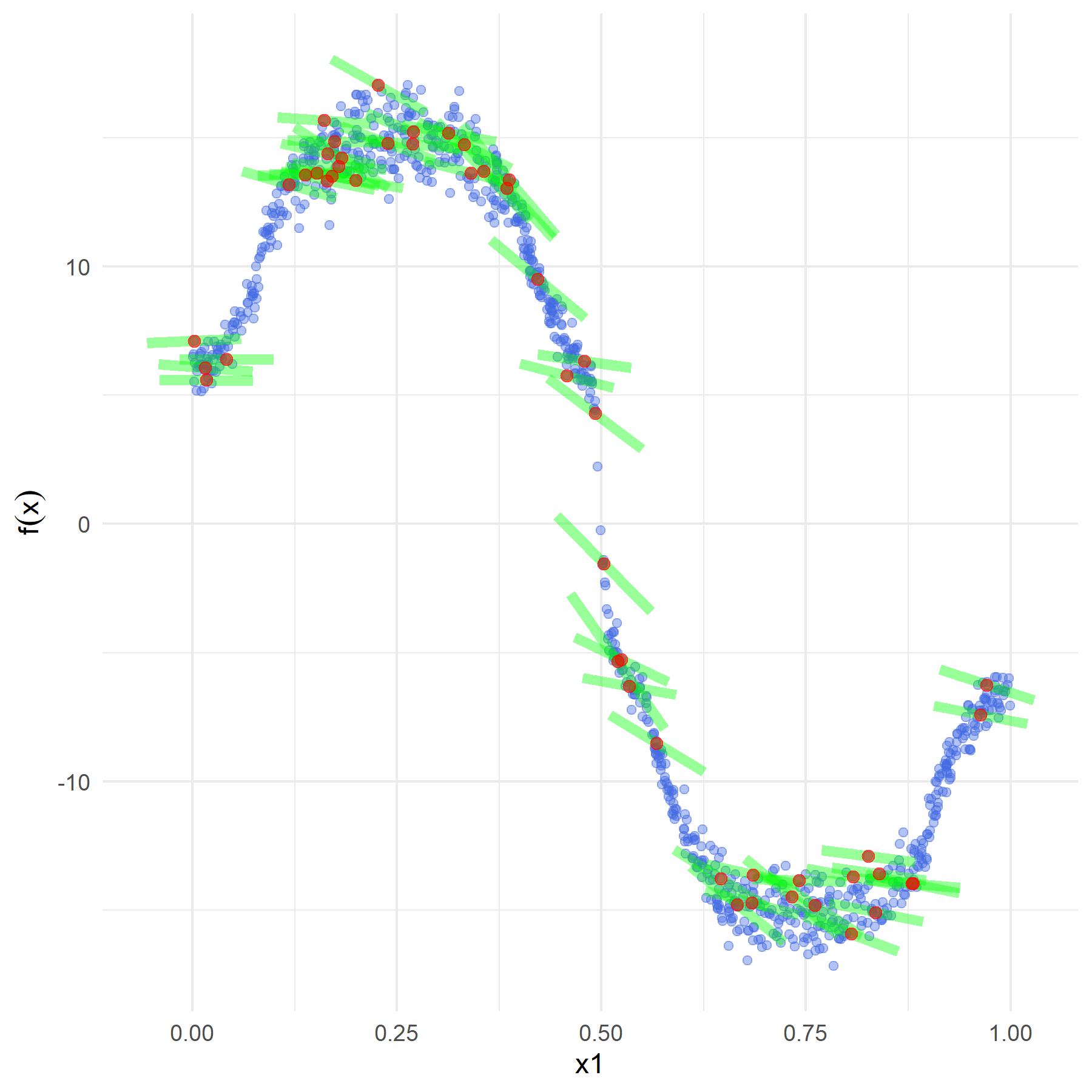}
            \caption[]%
            {{\small LIME}} 
            \label{fig:simslopec}
        \end{subfigure}
        \hfill
        \begin{subfigure}[b]{0.32\textwidth}  
            \centering 
            \includegraphics[width=\textwidth]{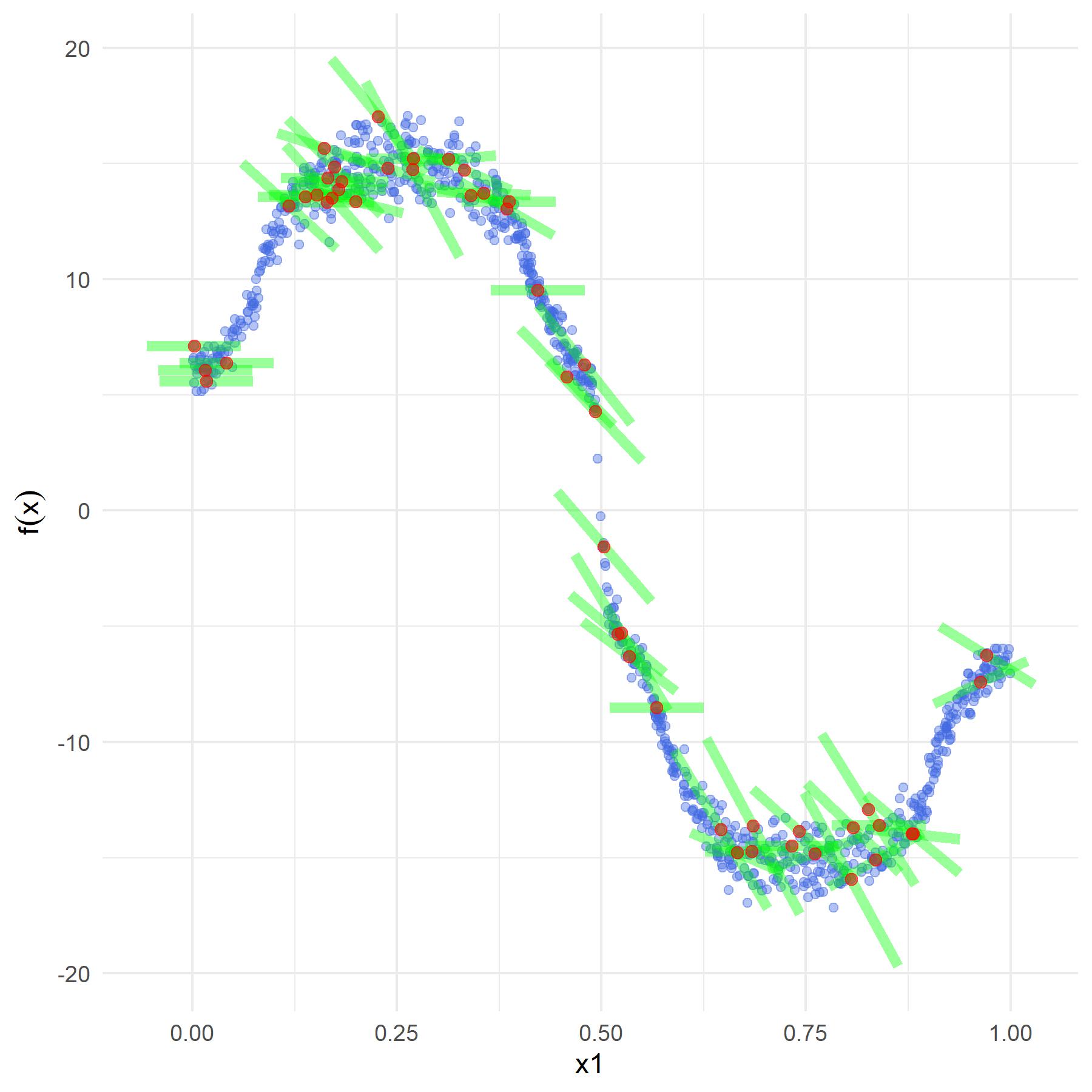}
            \caption[]%
            {{\small IML}} 
            \label{fig:simsloped}
        \end{subfigure}
        \begin{subfigure}[b]{0.32\textwidth}  
            \centering 
            \includegraphics[width=\textwidth]{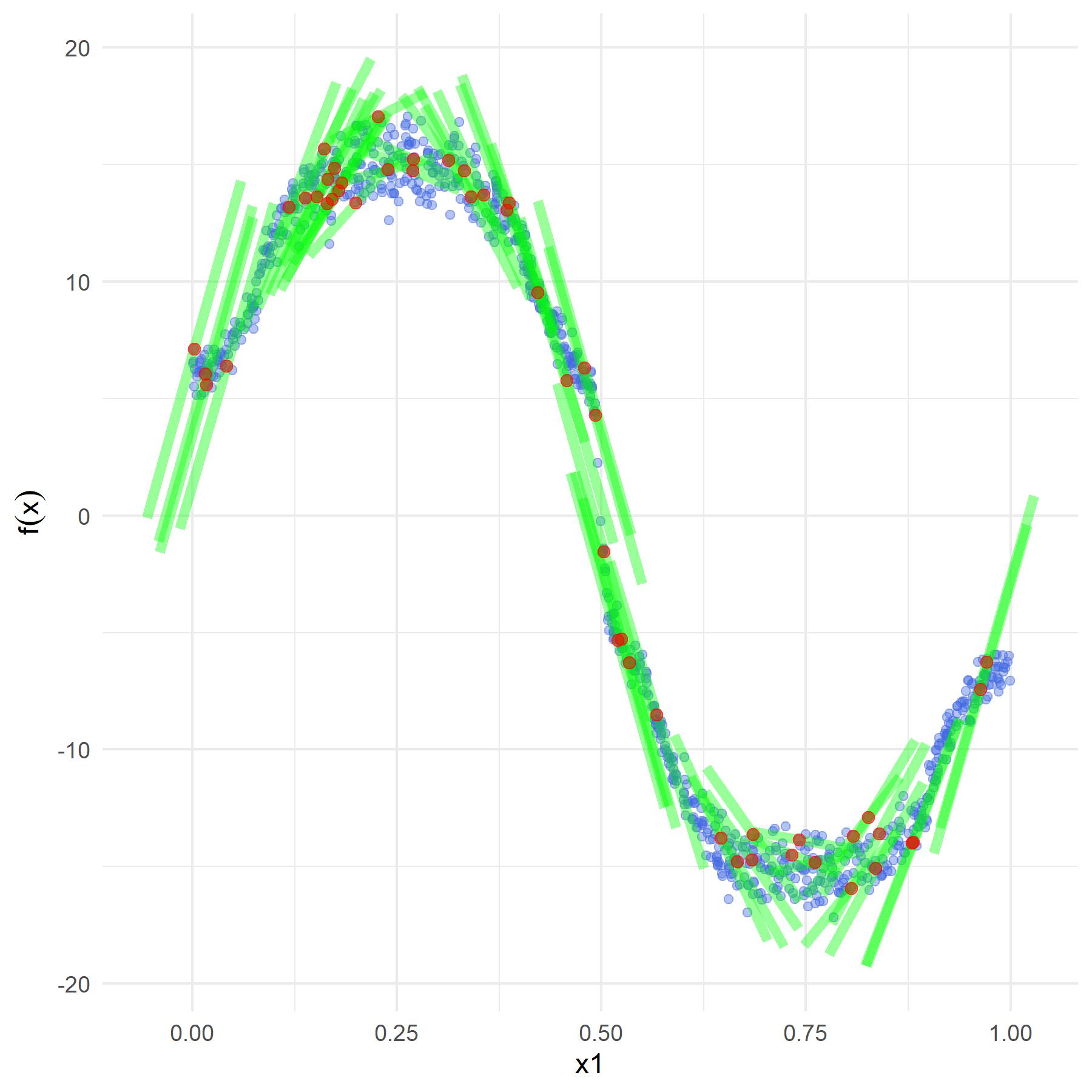} 
            \caption[]%
            {{\small True}} 
            \label{fig:simslopee}
        \end{subfigure}
        \caption[]
        {\small Local slope plots (green) for sample points (red) from the dataset 6. Coefficients estimated by VarImp are closer to the true coefficients.} 
        \label{fig:simslope}
\end{figure}

\textbf{Local slope plots.} Figure \ref{fig:simslope} shows that the coefficients estimated by VarImp are closer to the true coefficients for dataset 6 when compared to the other approaches. In general, VarImp has better coefficients for data with non-linear relationships between target and feature (see Appendix \ref{sec:slopeplots} for the other plots).

\begin{figure}[ht]
        \centering
        \begin{subfigure}[b]{0.32\textwidth}
            \centering
            \includegraphics[width=\textwidth]{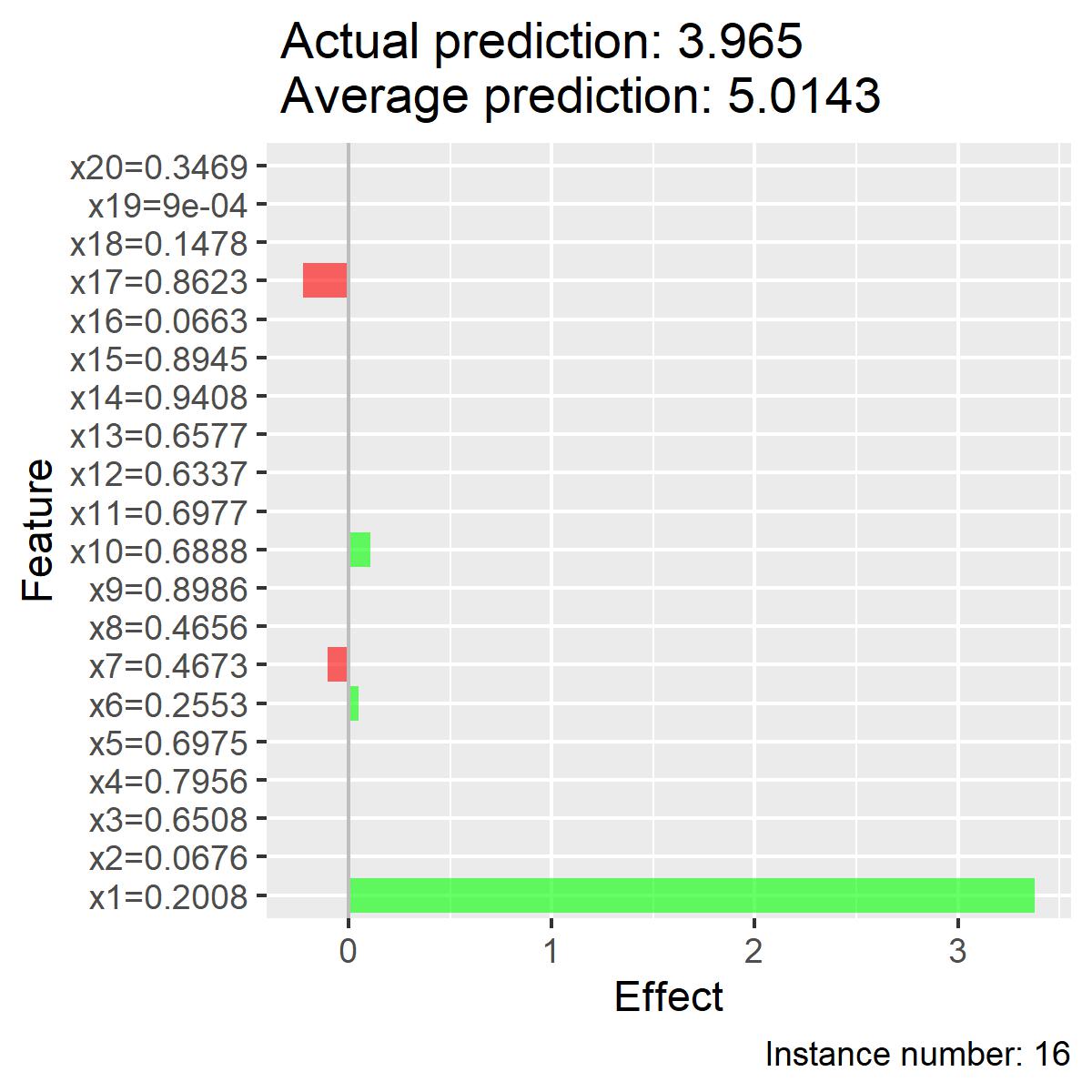}
            \caption[]%
            {{\small VarImp}}   
            \label{fig:simeffecta}
        \end{subfigure}
        \begin{subfigure}[b]{0.32\textwidth}  
            \centering 
            \includegraphics[width=\textwidth]{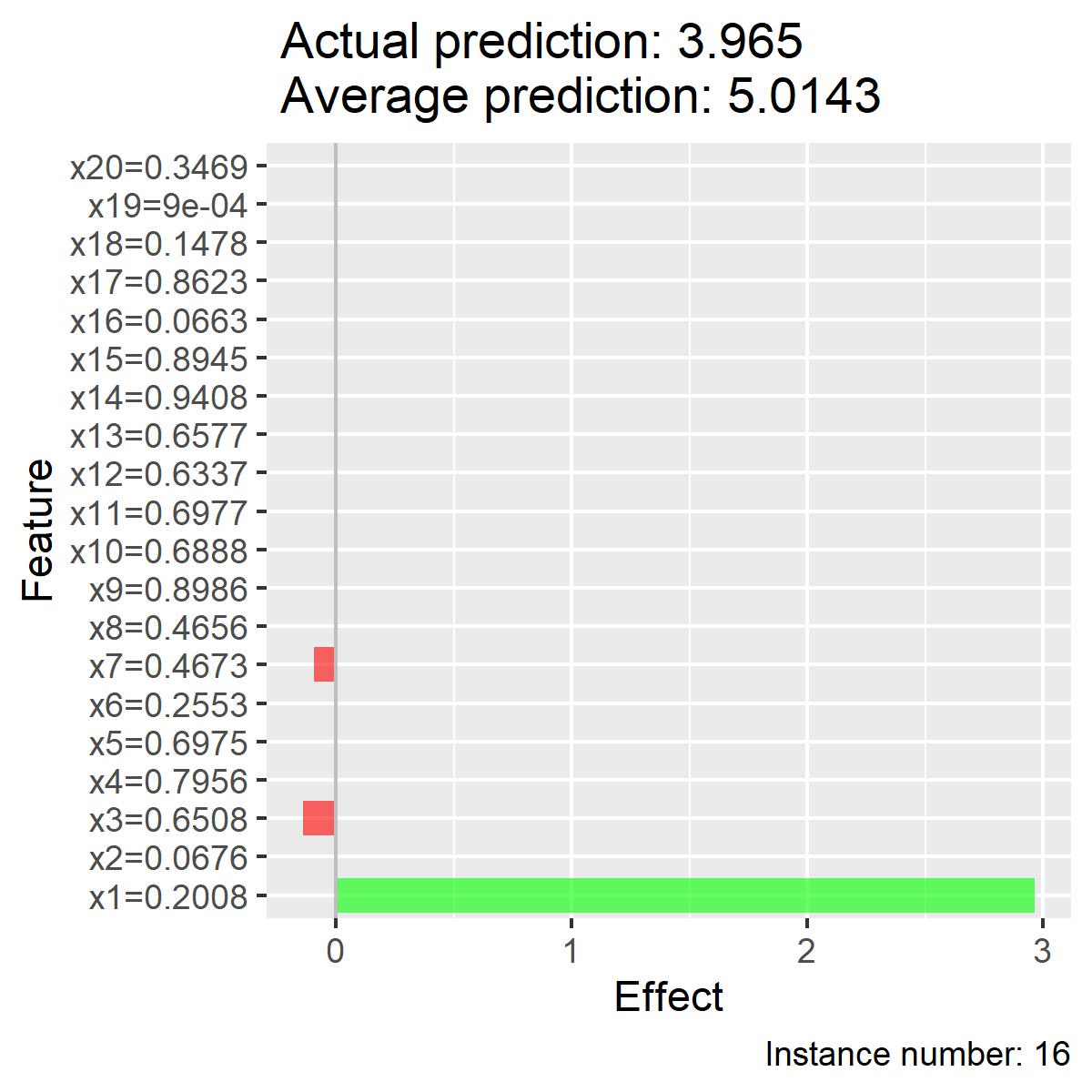}
            \caption[]%
            {{\small SupClus}} 
            \label{fig:simeffectb}
        \end{subfigure}
        \begin{subfigure}[b]{0.32\textwidth}  
            \centering 
            \includegraphics[width=\textwidth]{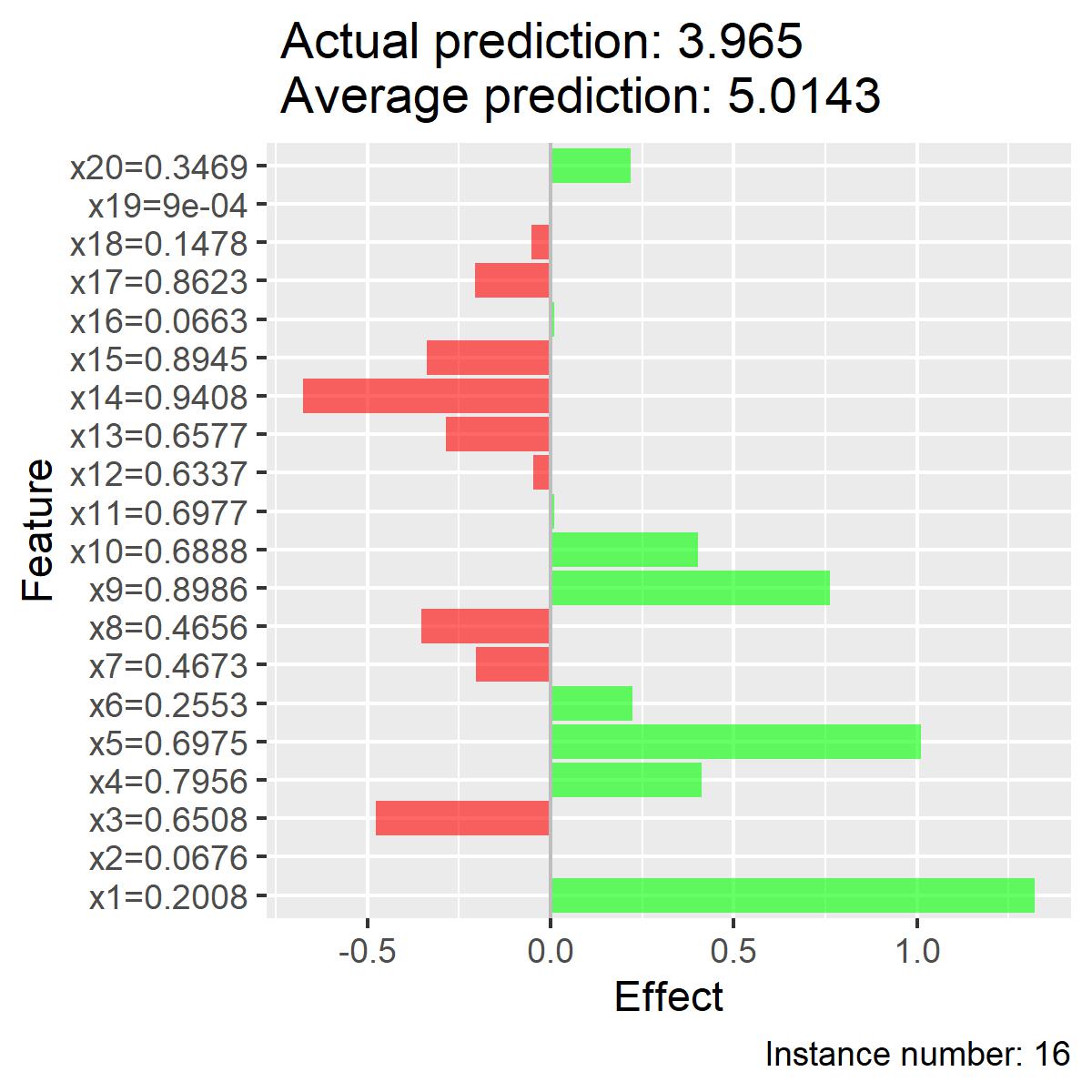}
            \caption[]%
            {{\small LIME}} 
            \label{fig:simeffectc}
        \end{subfigure}
        \hfill
        \begin{subfigure}[b]{0.32\textwidth}  
            \centering 
            \includegraphics[width=\textwidth]{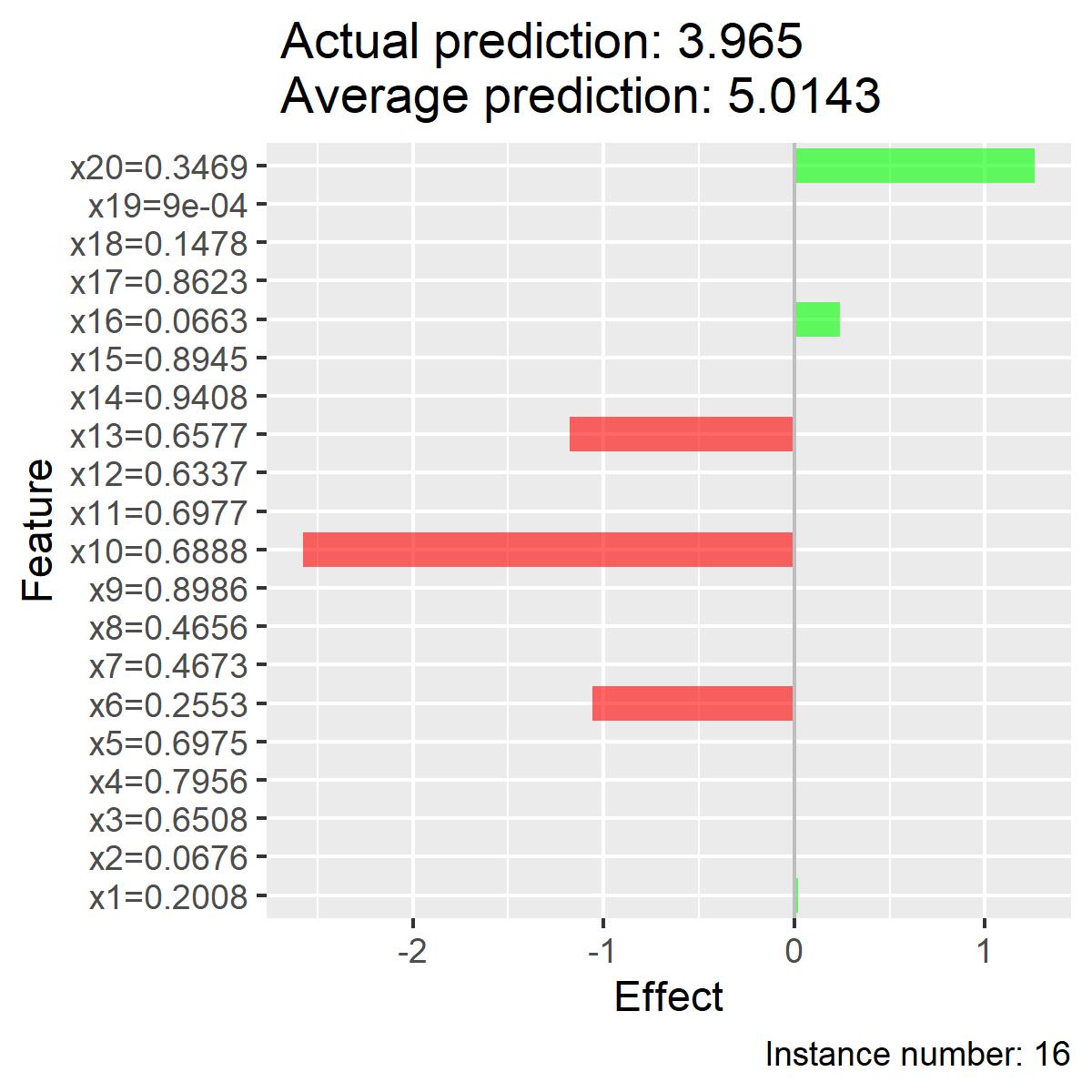}
            \caption[]%
            {{\small IML}} 
            \label{fig:simeffectd}
        \end{subfigure}
        \begin{subfigure}[b]{0.32\textwidth}  
            \centering 
            \includegraphics[width=\textwidth]{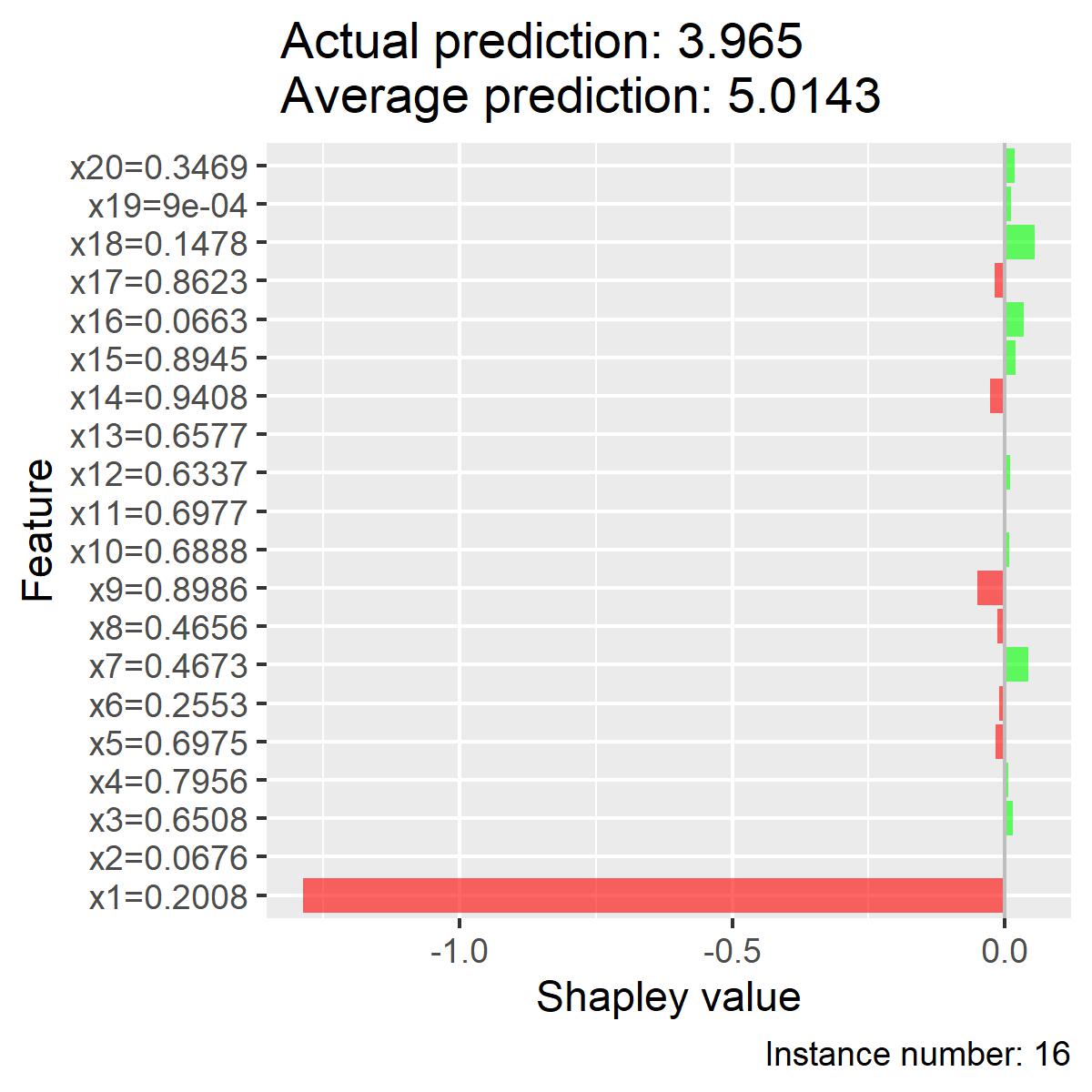}
            \caption[]%
            {{\small Shapley values}} 
            \label{fig:simeffecte}
        \end{subfigure}
        \caption[]
        {\small Estimated effects of an instance for dataset 2. VarImp and SupClus provided better explanations.} 
        \label{fig:simeffect}
\end{figure}

\textbf{Instance analysis.} Finally, Figure \ref{fig:simeffect} shows a comparison of the estimated effects of a specific instance made by VarImp, SupClus, LIME, IML, and Shapley values for dataset 2, where only $X_1$ is shown to be relevant ($Y = 20X_1 \mathbf{1}_{X_1 \leq 0.5}(X_1) + (20-20X_1)\mathbf{1}_{X_1 > 0.5}(X_1) + \epsilon$). We analyze an instance with $X_1 \leq 0.5$, for which a positive effect is expected for $X_1$. VarImp and SupClus provided better explanations, whereas LIME and IML estimated large effects for irrelevant variables and Shapley value estimated a negative effect for $X_1$. The same conclusions also hold for the   other data sets (see Appendix \ref{sec:effectinstance}).

\subsection{Real data}

\begin{figure}[ht]
        \centering
        \begin{subfigure}[b]{0.49\textwidth}
            \centering
            \includegraphics[width=\textwidth]{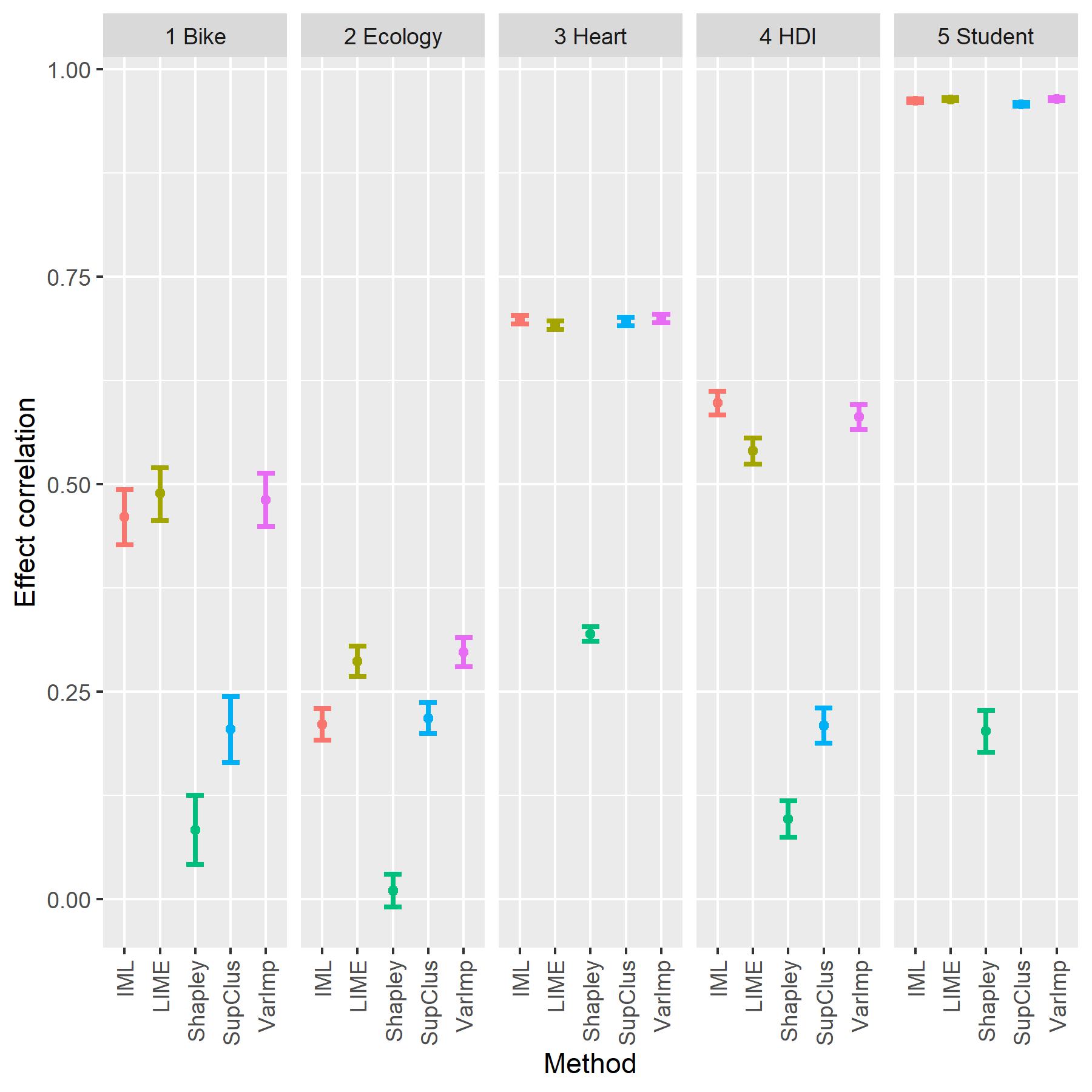}
            \caption[]%
            {{\small Effect correlation}}   
            \label{fig:realcora}
        \end{subfigure}
        \hfill
        \begin{subfigure}[b]{0.49\textwidth}  
            \centering 
            \includegraphics[width=\textwidth]{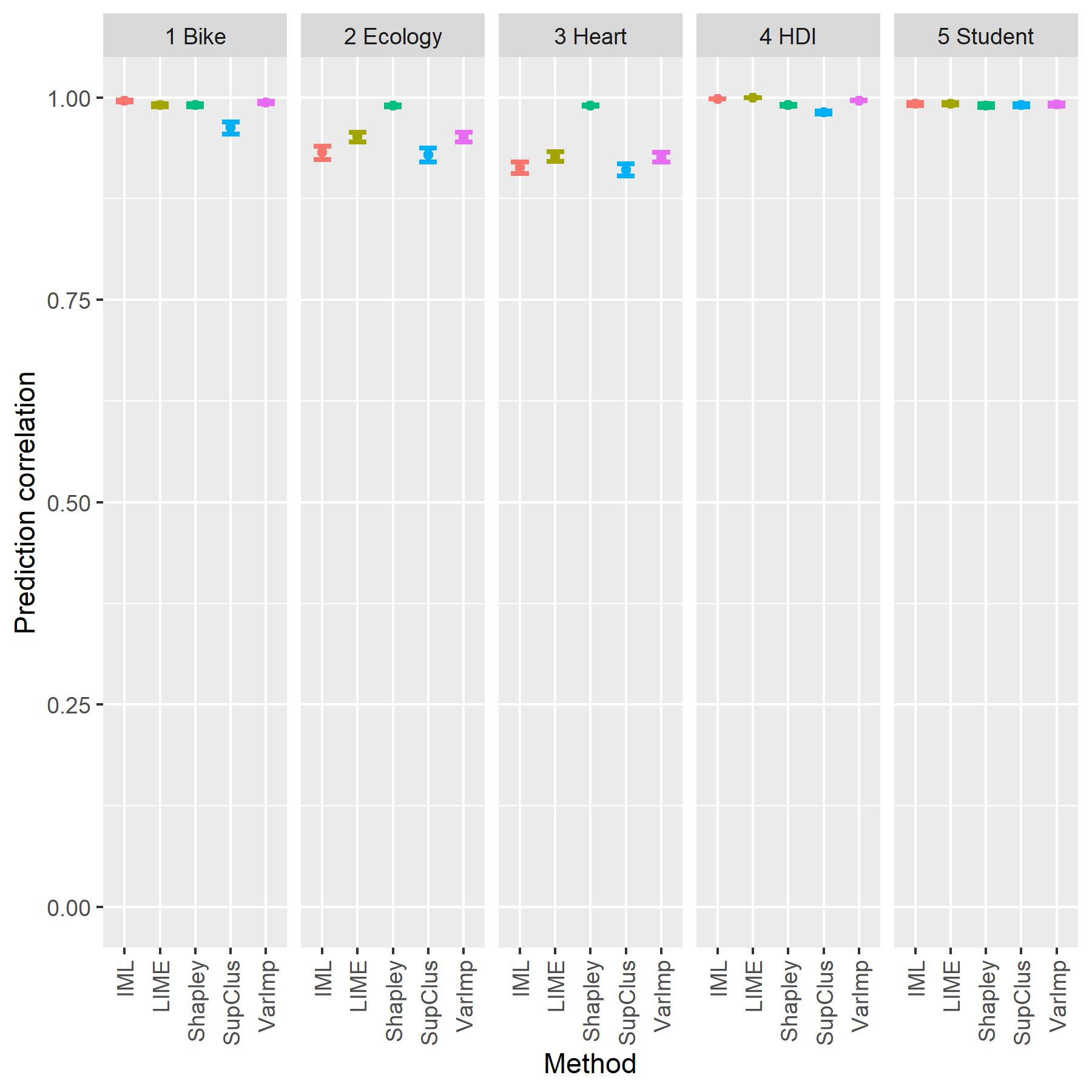}
            \caption[]%
            {{\small Prediction correlation}} 
            \label{fig:realcorb}
        \end{subfigure}
        \caption[]
        {\small Correlations of effects and predictions for real data with $95\%$ confidence level. Overall, VarImp has better ICE effect correlations than the other methods.} 
        \label{fig:realcor}
\end{figure}

Comparing interpretability methods using real datasets is  more challenging since the true local coefficients are not known. Thus, instead of computing the correlations between estimated and true effects, we compared the estimated effects of the methods with the estimated effects using ICE, as described in Section \ref{sec:experiments} (Figure \ref{fig:realcora}).  Correlations between local and global predictions were also calculated (Figure \ref{fig:realcorb}).

\textbf{ICE effect and Prediction correlation.}
 Shapley values have the worst correlations of ICE effects in all datasets. VarImp, SupClus, LIME, and IML have the best correlations of effects for datasets 3 and 5. On the other hand, VarImp, LIME, and IML have the best correlations for dataset 1, VarImp and LIME have the best correlations for dataset 2 and VarImp and IML have the best correlations for dataset 4. All methods show good correlations of predictions for all datasets. Overall, VarImp has better ICE effect correlations than the other methods.

\begin{figure}[ht]
        \centering
        \begin{subfigure}[b]{0.49\textwidth}
            \centering
            \includegraphics[width=\textwidth]{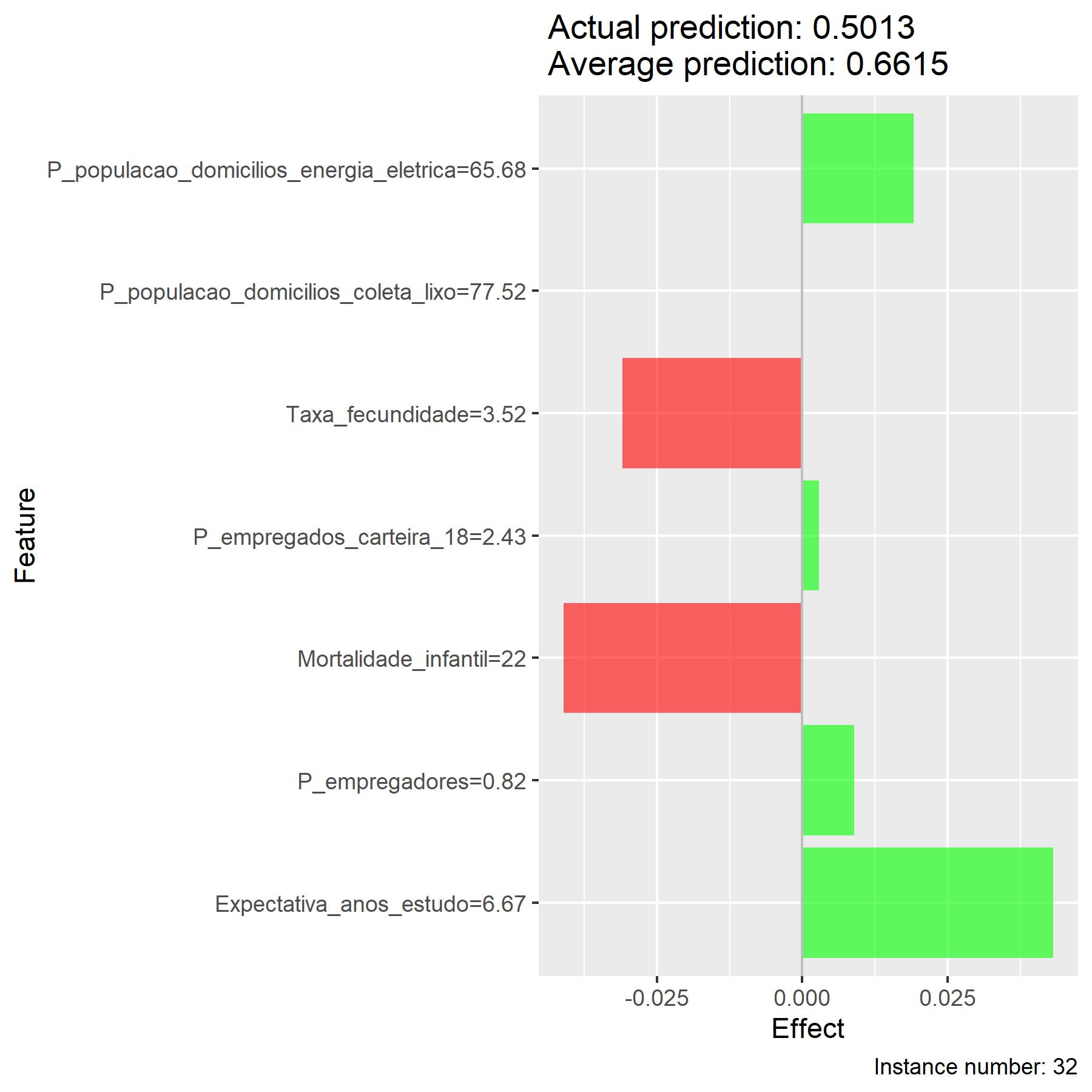}
            \caption[]%
            {{\small Low prediction}}   
            \label{fig:dhiextremea}
        \end{subfigure}
        \hfill
        \begin{subfigure}[b]{0.49\textwidth}  
            \centering 
            \includegraphics[width=\textwidth]{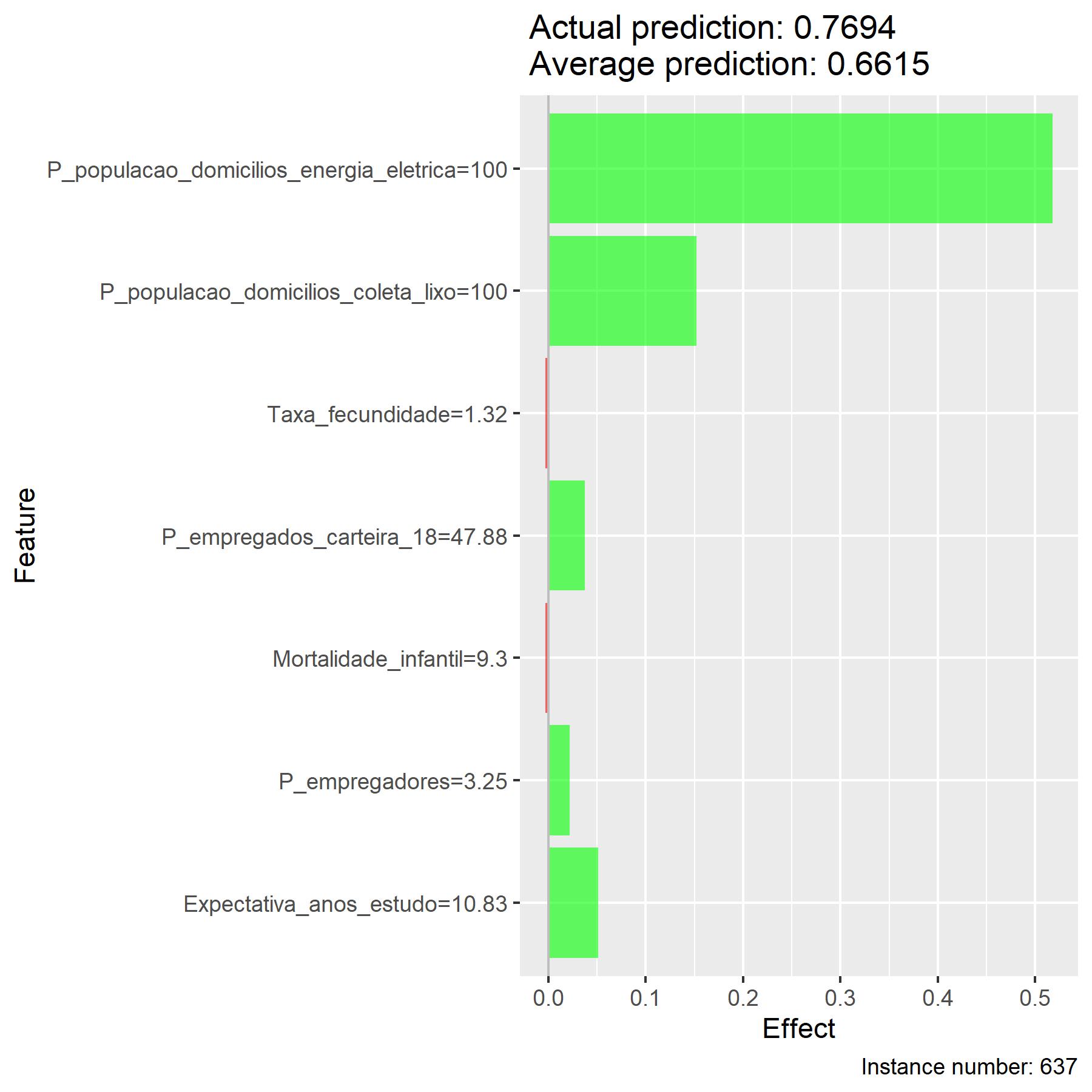}
            \caption[]%
            {{\small High prediction}} 
            \label{fig:dhiextremeb}
        \end{subfigure}
        \caption[]
        {\small Explanations of instances with extreme predictions from the HDI dataset using VarImp. The prediction of the low HDI city is mainly due to the high infant mortality rate. The prediction of the city with a high HDI can be explained by the high percentage of households with electricity and garbage collection.} 
        \label{fig:dhiextreme}
\end{figure}

\textbf{Extreme cases.} When we analyze instances individually, it is interesting to compare extreme cases (low and high predictions) to evaluate if the explanations make sense. Using the HDI dataset we compared two cities with extreme predictions using VarImp (Figure \ref{fig:dhiextreme}). The prediction of the low HDI city is mainly due to the high infant mortality rate. On the other hand, the prediction of the city with a high HDI can be explained by the high percentage of households with electricity and garbage collection. Thus, this analysis reinforces the quality of the explanations.

\section{Conclusion}\label{sec:conclusion}

This article proposed two new methods of local and agnostic interpretability, namely VarImp and SupClus, which can better deal with high-dimensional situations with irrelevant variables and better work when the relationship between input variables and target is not linear. Both are based on a local linear regression improved through the weights of local variables included in the Euclidean distance. While SupClus interprets instances in groups with similar explanations, VarImp can be applied to data with more complex relationships and that require individual interpretations per instance.

Measures for assessments of the quality of VarImp and SupClus interpretations, and to compare then with other methods, were used. These measures consider not one instance only, but a complete sample of points, and use non-subjective criteria more directly related to the quality of explanations. A way to visualize the explanations of a method for each variable is used where the local coefficients for a sample of points are plotted.

Applications in artificial and real data indicated that the proposed methods achieve better performance than known interpretability methods, LIME, IML, and Shapley values. In most datasets, VarImp showed better quality measures, especially for datasets with nonlinear relationships or irrelevant variables. SupClus performed well for simpler datasets, where clusters of instances with similar explanations can be identified. Both proposed methods, together with LIME and IML provided good results in datasets with all relevant variables and linear relationships (which is common in practice). Despite its good properties and good local predictions, Shapley values method did not provide good quality measures in the investigated applications and LIME and IML showed poor-quality interpretations in datasets with nonlinear relationships.

The R package with the implementation of VarImp and SupClus is available at: https://github.com/gilsonshimizu/interpretability.

\backmatter


\bmhead*{Acknowledgments}
Gilson Y. Shimizu is grateful for the financial support of Pró-Reitoria de Pesquisa da USP - Brazil.
Rafael Izbicki is grateful for the financial support of FAPESP (grant 2019/11321-9) and CNPq (grant 306943/2017-4). Andre C. P. L. F. de Carvalho is grateful for the financial support of CNPq (grants 309858/2021-6 and 422705/2021-7).

\section*{Declarations}

\textbf{Conflict of Interest} The authors declare that they have no conflict of interest.







\begin{appendices}

\section{Local slope plots for artificial datasets.}\label{sec:slopeplots}

\begin{figure}[H]
        \centering
        \begin{subfigure}[b]{0.32\textwidth}
            \centering
            \includegraphics[width=\textwidth]{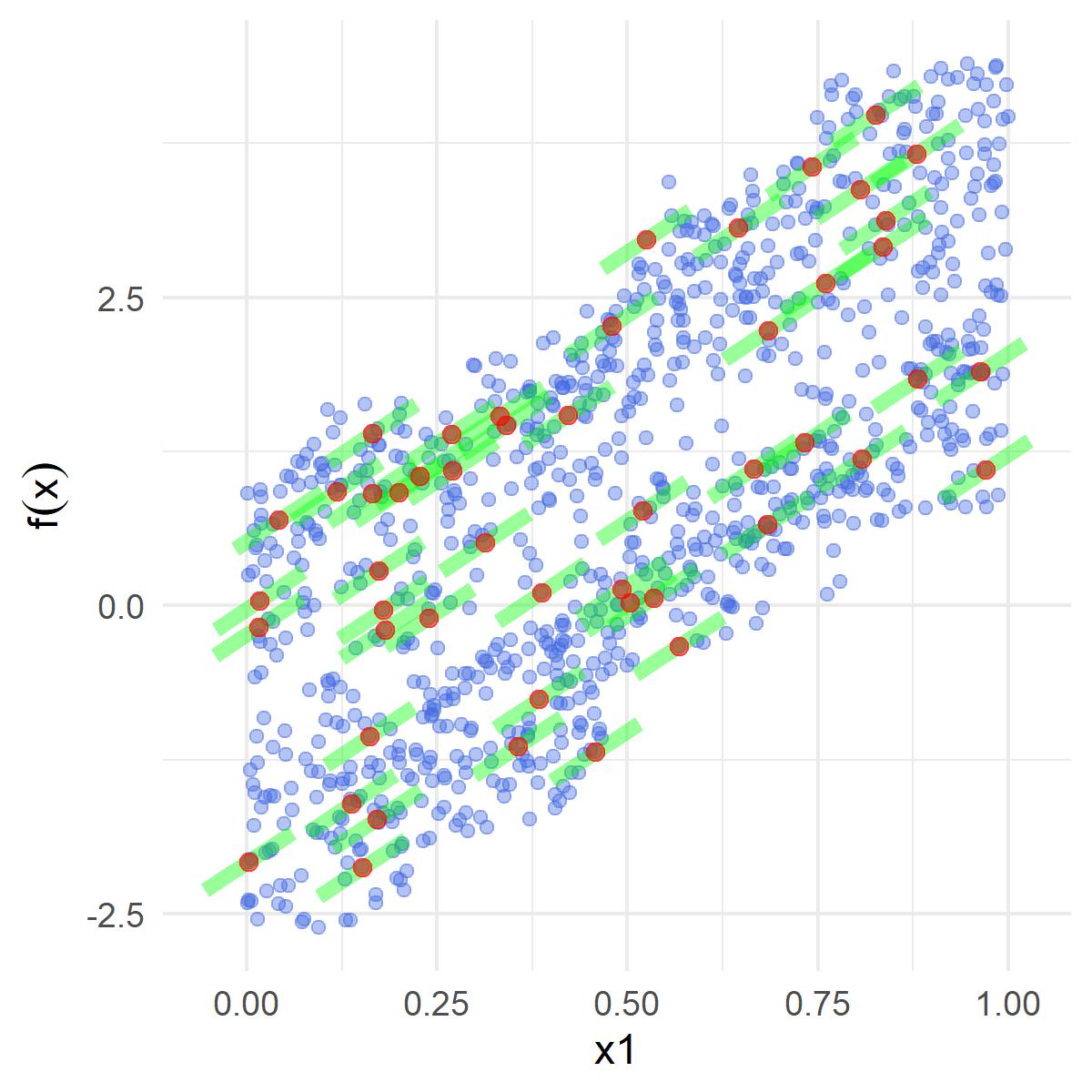}
            \caption[]%
            {{\small SupClus}}   
        \end{subfigure}
        \begin{subfigure}[b]{0.32\textwidth}  
            \centering 
            \includegraphics[width=\textwidth]{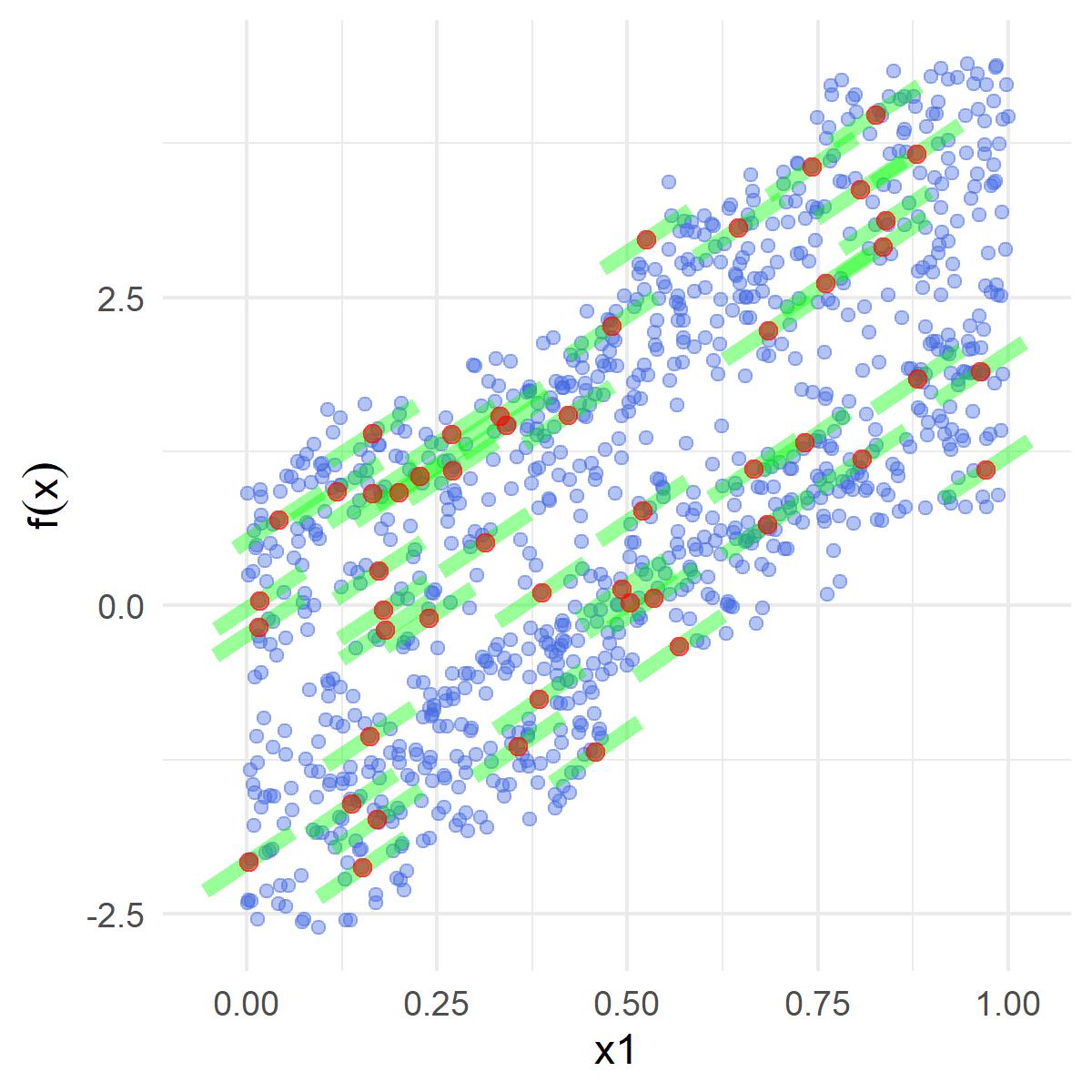}
            \caption[]%
            {{\small VarImp}} 
        \end{subfigure}
        \begin{subfigure}[b]{0.32\textwidth}  
            \centering 
            \includegraphics[width=\textwidth]{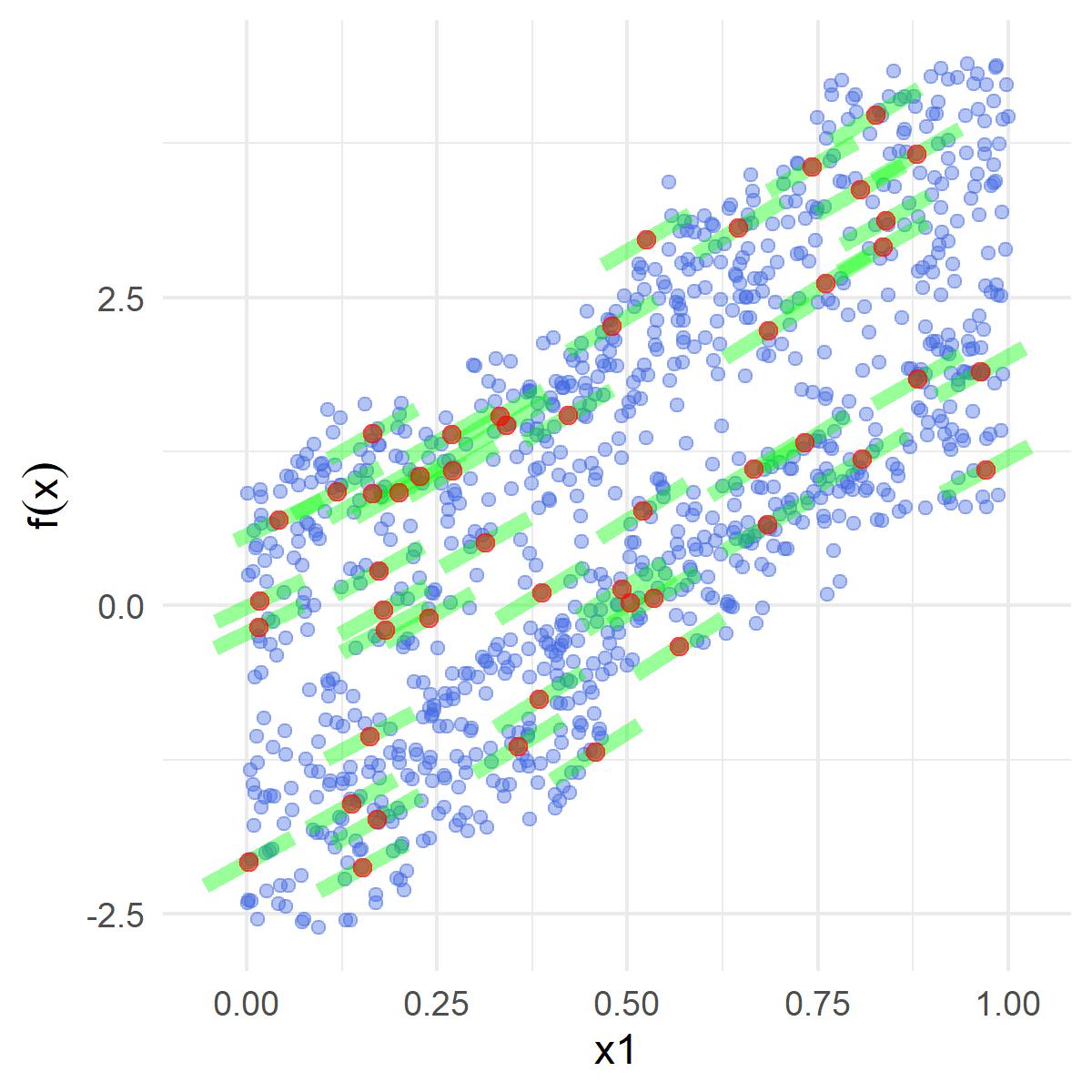}
            \caption[]%
            {{\small LIME}} 
        \end{subfigure}
        \hfill
        \begin{subfigure}[b]{0.32\textwidth}  
            \centering 
            \includegraphics[width=\textwidth]{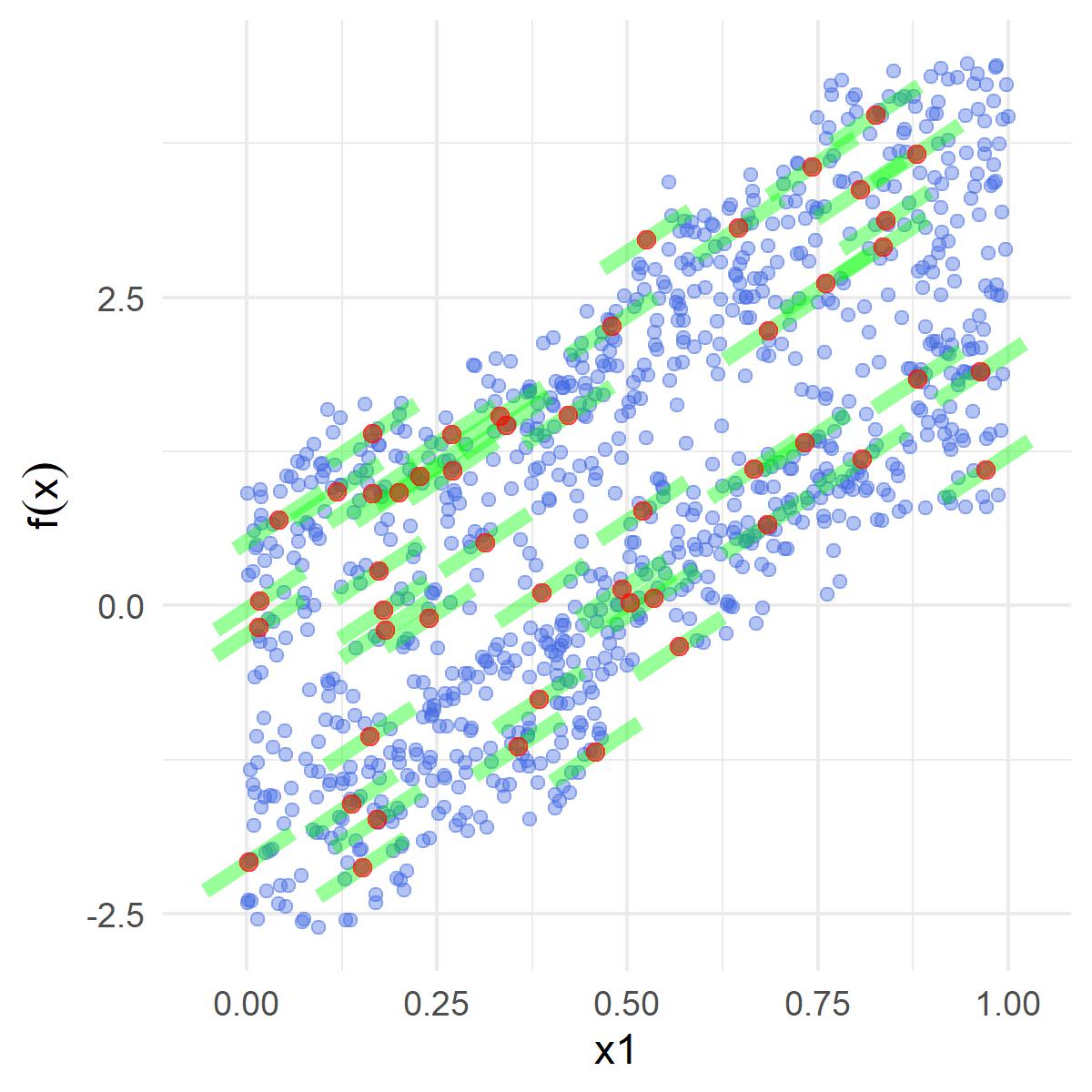}
            \caption[]%
            {{\small IML}} 
        \end{subfigure}
        \begin{subfigure}[b]{0.32\textwidth}  
            \centering 
            \includegraphics[width=\textwidth]{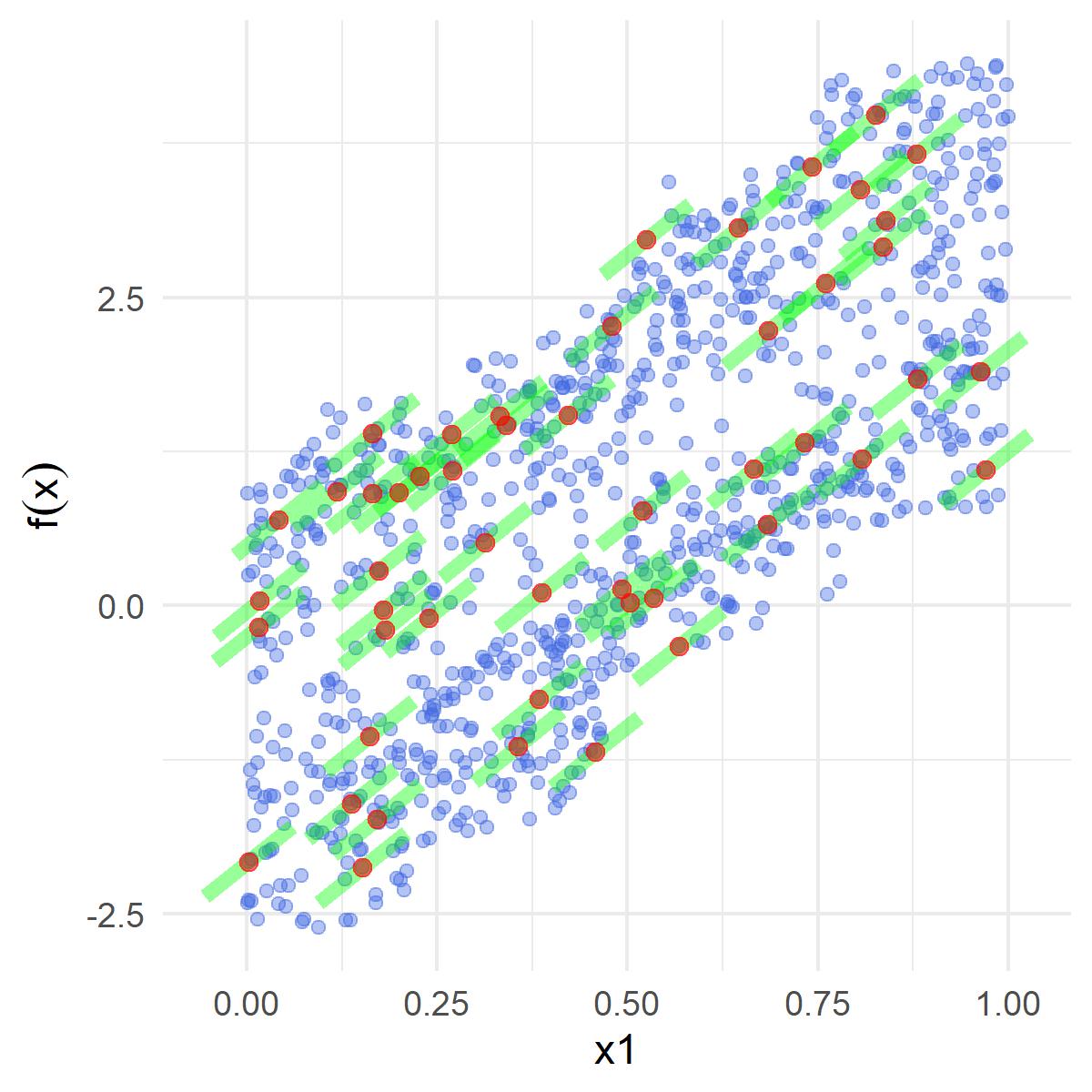} 
            \caption[]%
            {{\small True}} 
        \end{subfigure}
        \caption[]
        {\small Local slope plots (green) for sample points (red) from the dataset 1.} 
\end{figure}

\begin{figure}[H]
        \centering
        \begin{subfigure}[b]{0.32\textwidth}
            \centering
            \includegraphics[width=\textwidth]{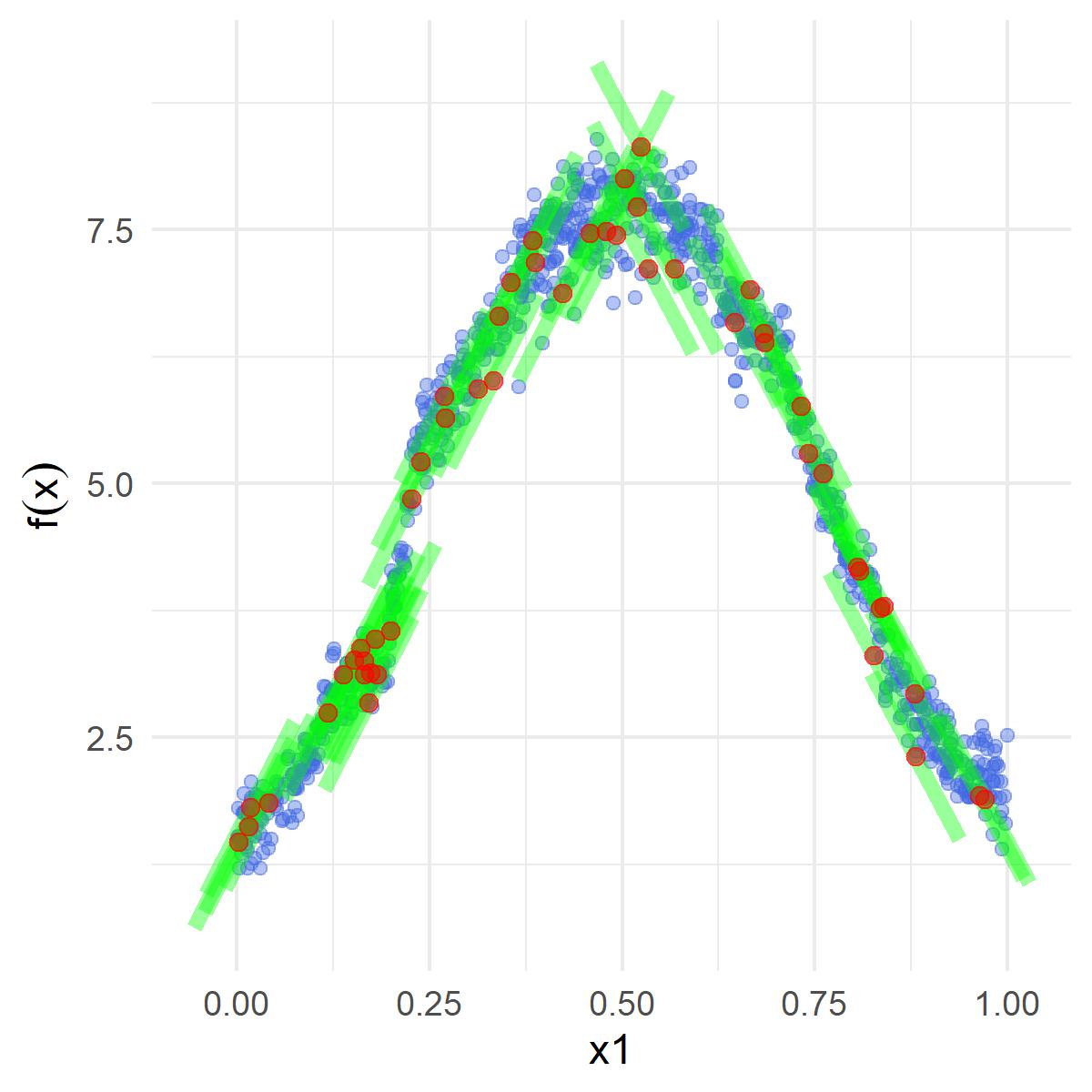}
            \caption[]%
            {{\small SupClus}}   
        \end{subfigure}
        \begin{subfigure}[b]{0.32\textwidth}  
            \centering 
            \includegraphics[width=\textwidth]{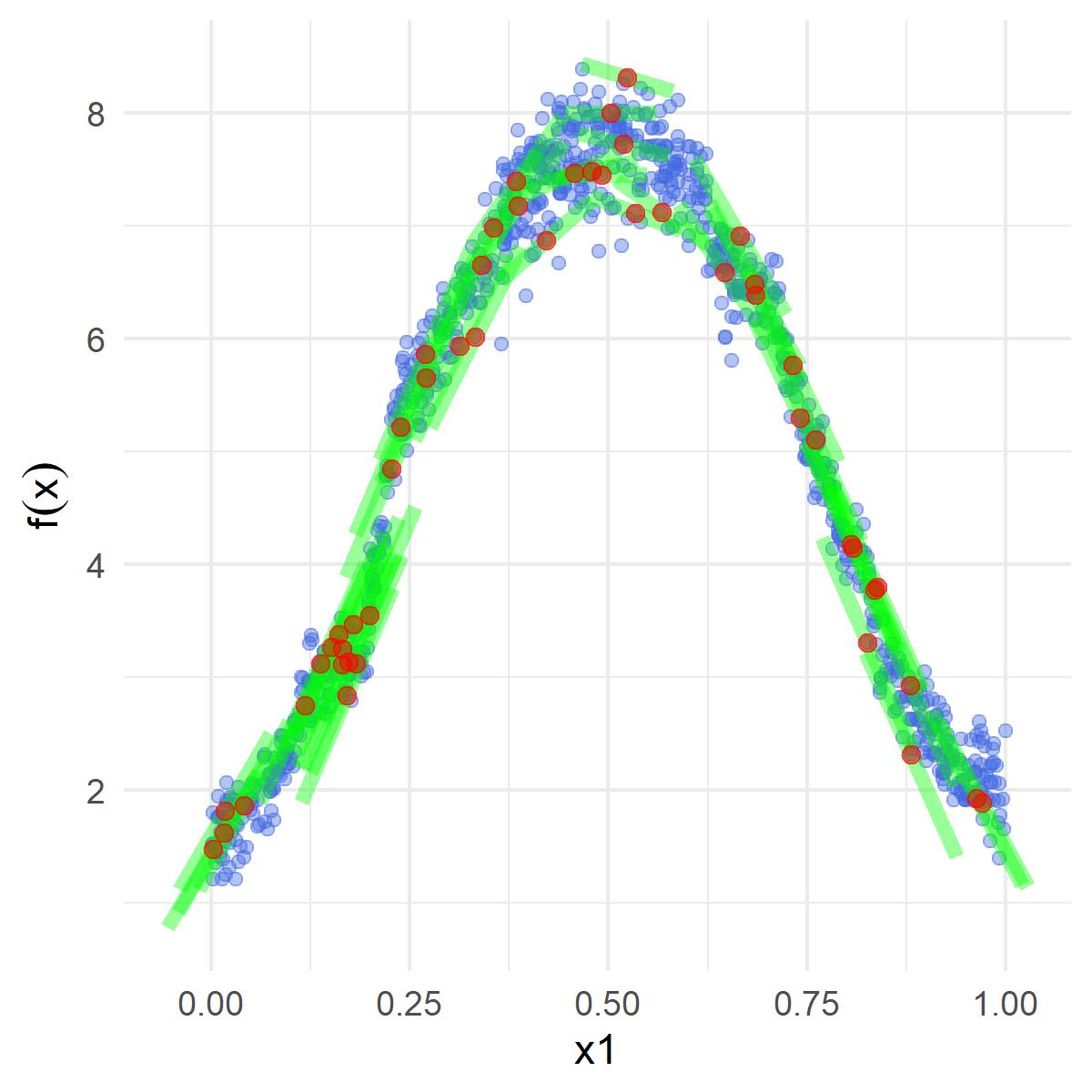}
            \caption[]%
            {{\small VarImp}} 
        \end{subfigure}
        \begin{subfigure}[b]{0.32\textwidth}  
            \centering 
            \includegraphics[width=\textwidth]{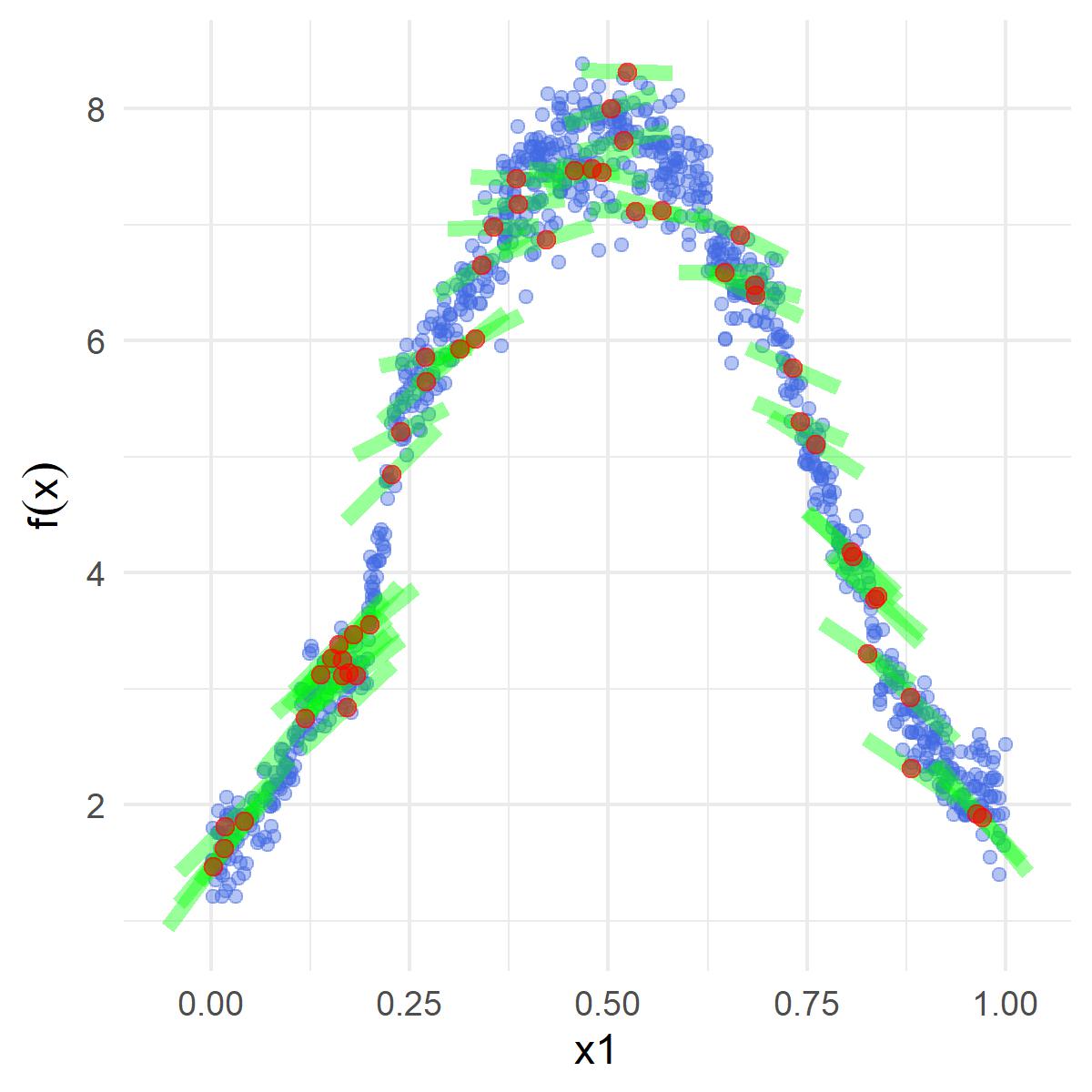}
            \caption[]%
            {{\small LIME}} 
        \end{subfigure}
        \hfill
        \begin{subfigure}[b]{0.32\textwidth}  
            \centering 
            \includegraphics[width=\textwidth]{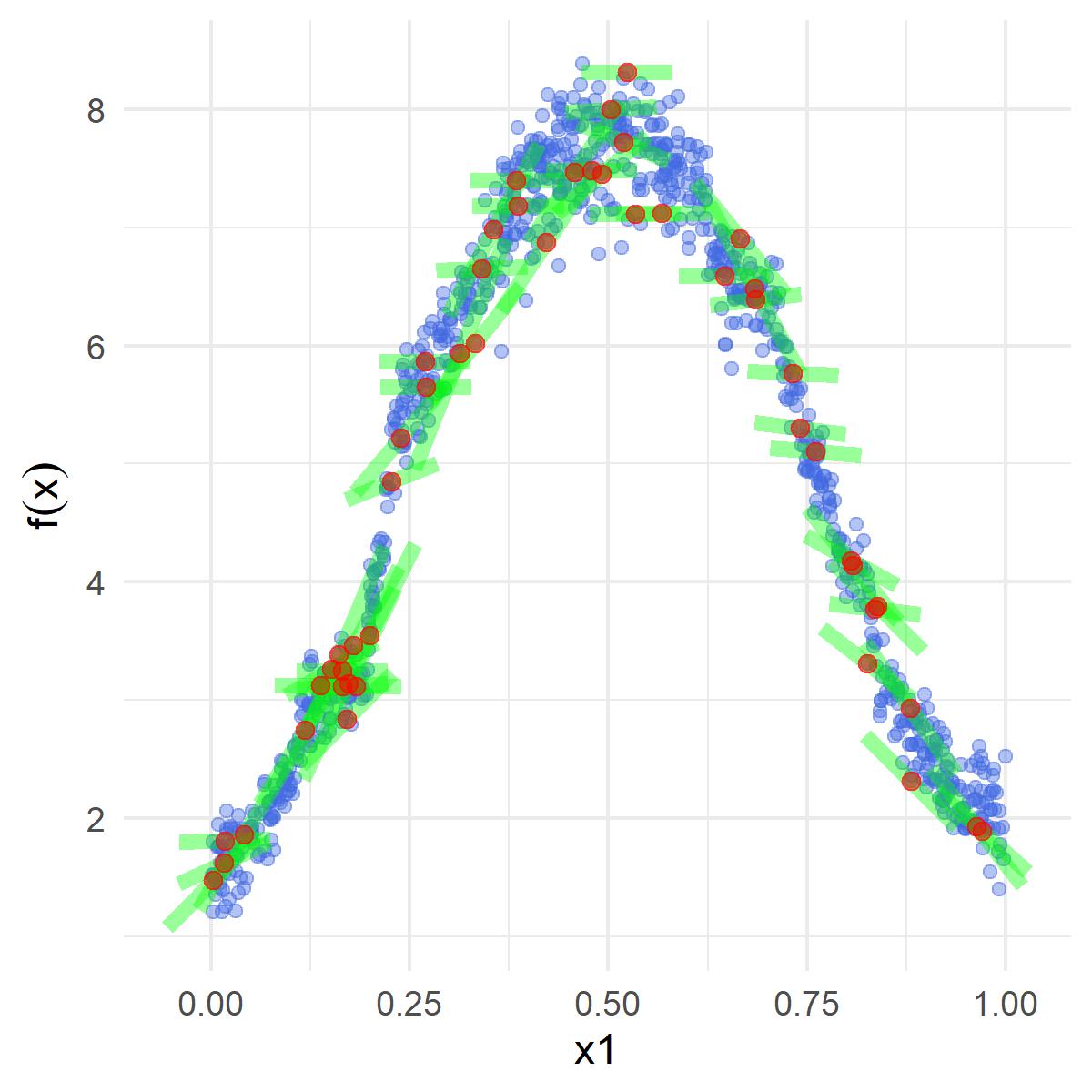}
            \caption[]%
            {{\small IML}} 
        \end{subfigure}
        \begin{subfigure}[b]{0.32\textwidth}  
            \centering 
            \includegraphics[width=\textwidth]{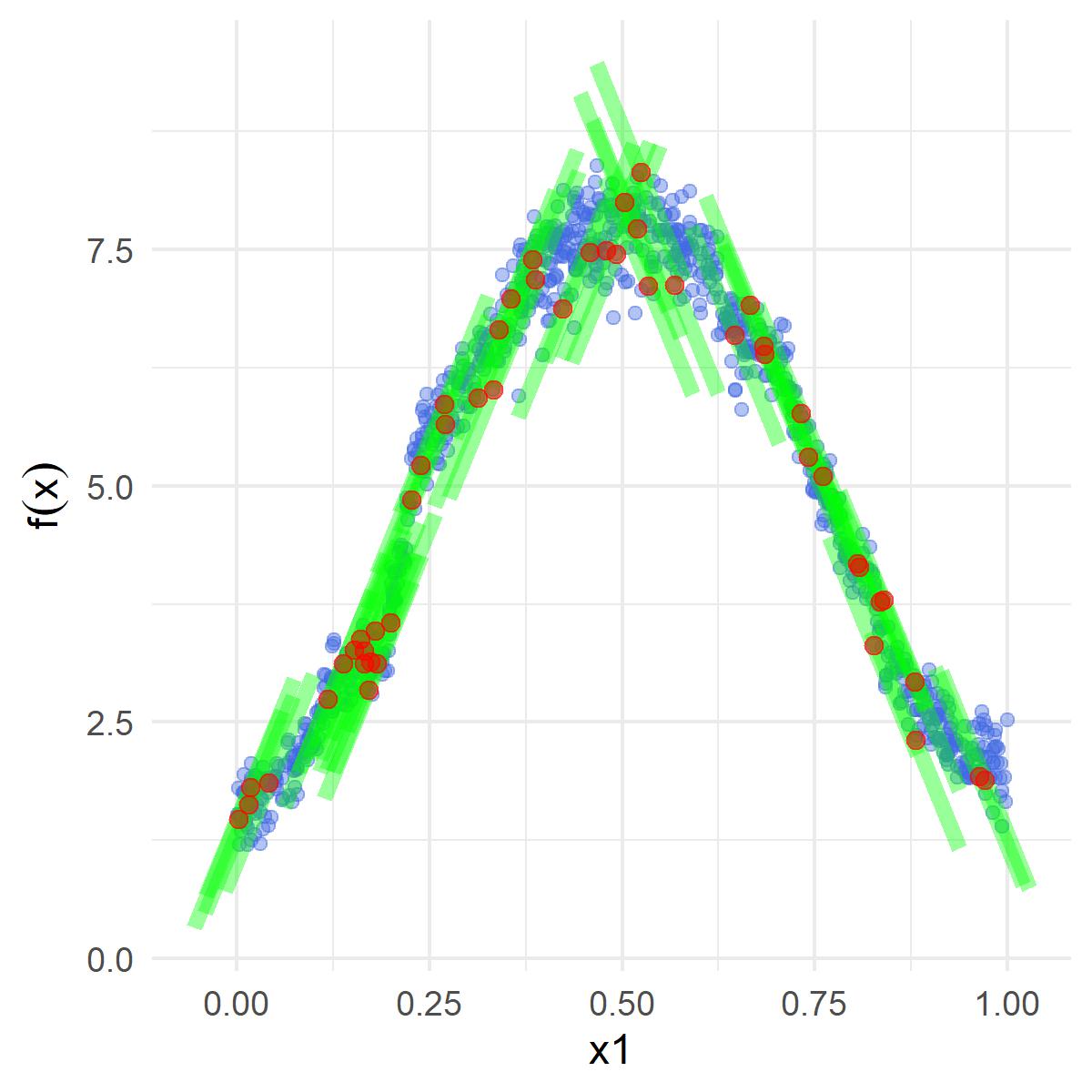} 
            \caption[]%
            {{\small True}} 
        \end{subfigure}
        \caption[]
        {\small Local slope plots (green) for sample points (red) from the dataset 2.} 
\end{figure}

\begin{figure}[H]
        \centering
        \begin{subfigure}[b]{0.32\textwidth}
            \centering
            \includegraphics[width=\textwidth]{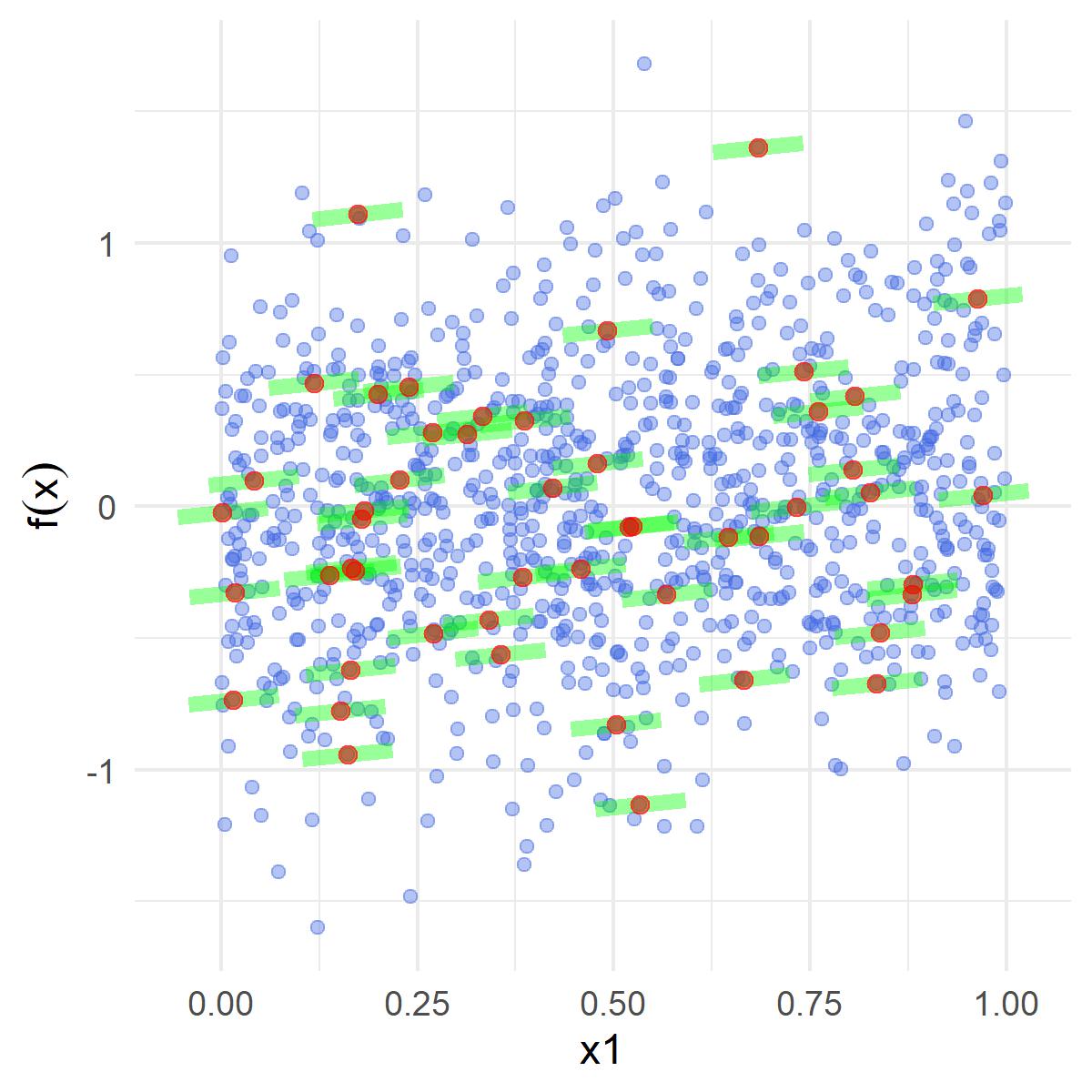}
            \caption[]%
            {{\small SupClus}}   
        \end{subfigure}
        \begin{subfigure}[b]{0.32\textwidth}  
            \centering 
            \includegraphics[width=\textwidth]{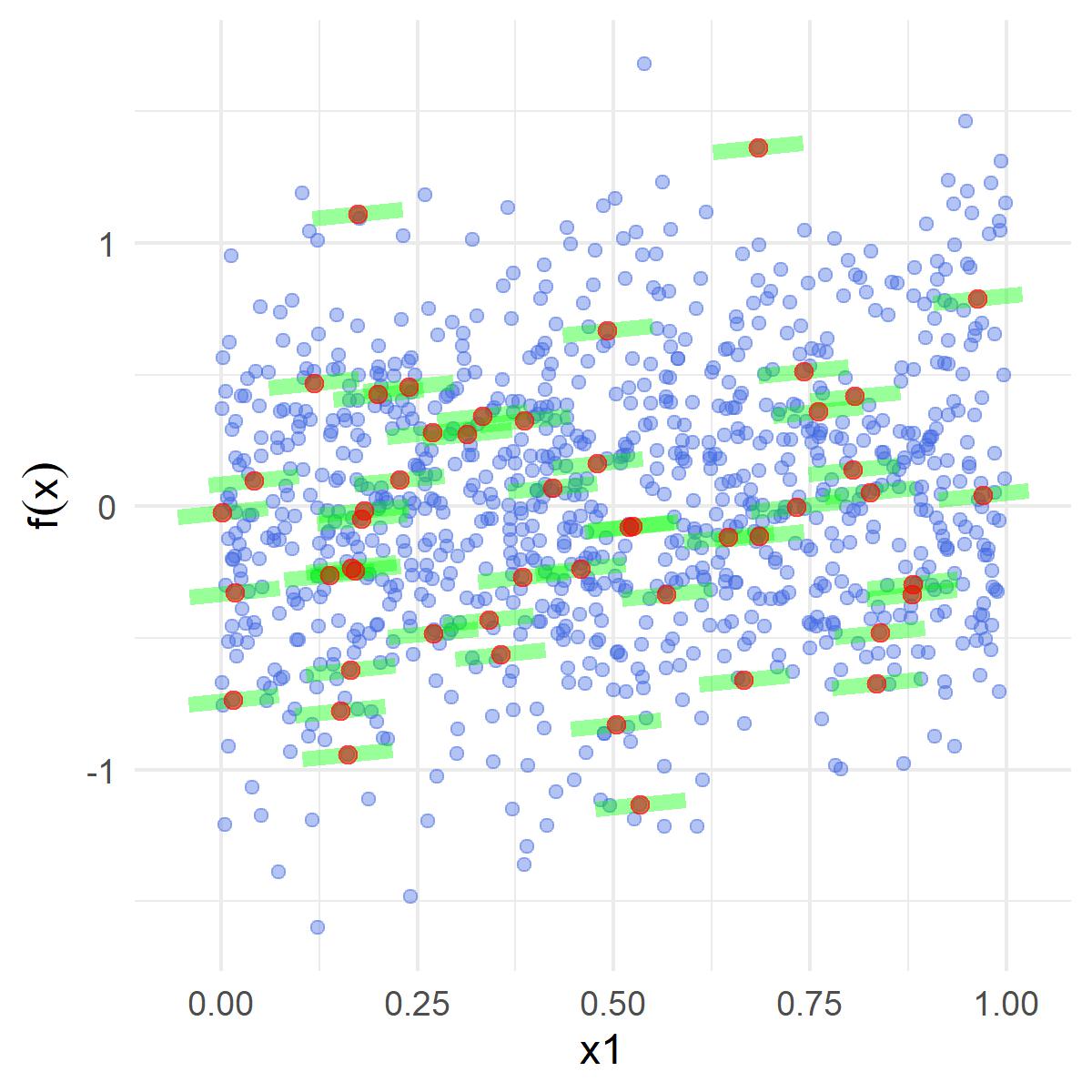}
            \caption[]%
            {{\small VarImp}} 
        \end{subfigure}
        \begin{subfigure}[b]{0.32\textwidth}  
            \centering 
            \includegraphics[width=\textwidth]{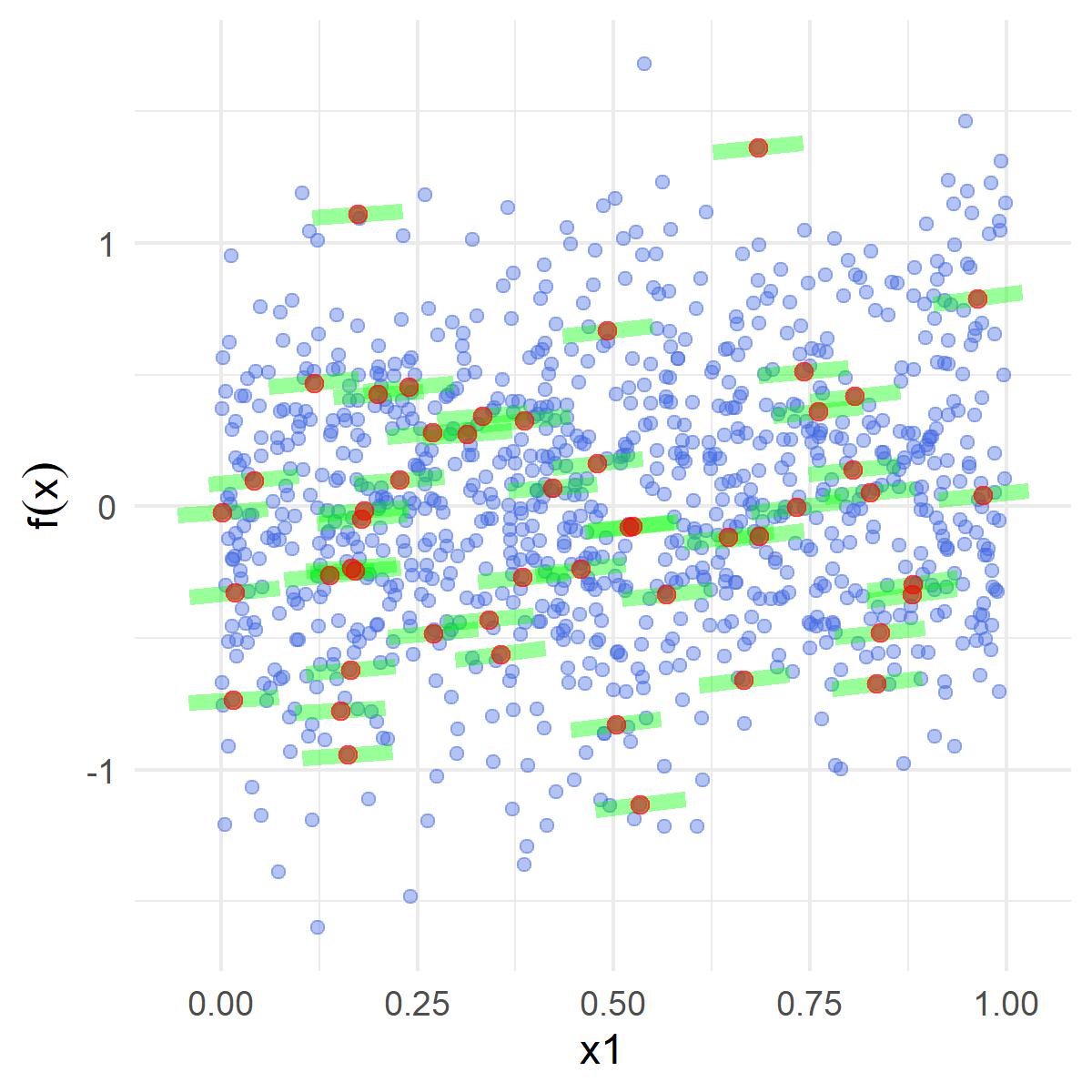}
            \caption[]%
            {{\small LIME}} 
        \end{subfigure}
        \hfill
        \begin{subfigure}[b]{0.32\textwidth}  
            \centering 
            \includegraphics[width=\textwidth]{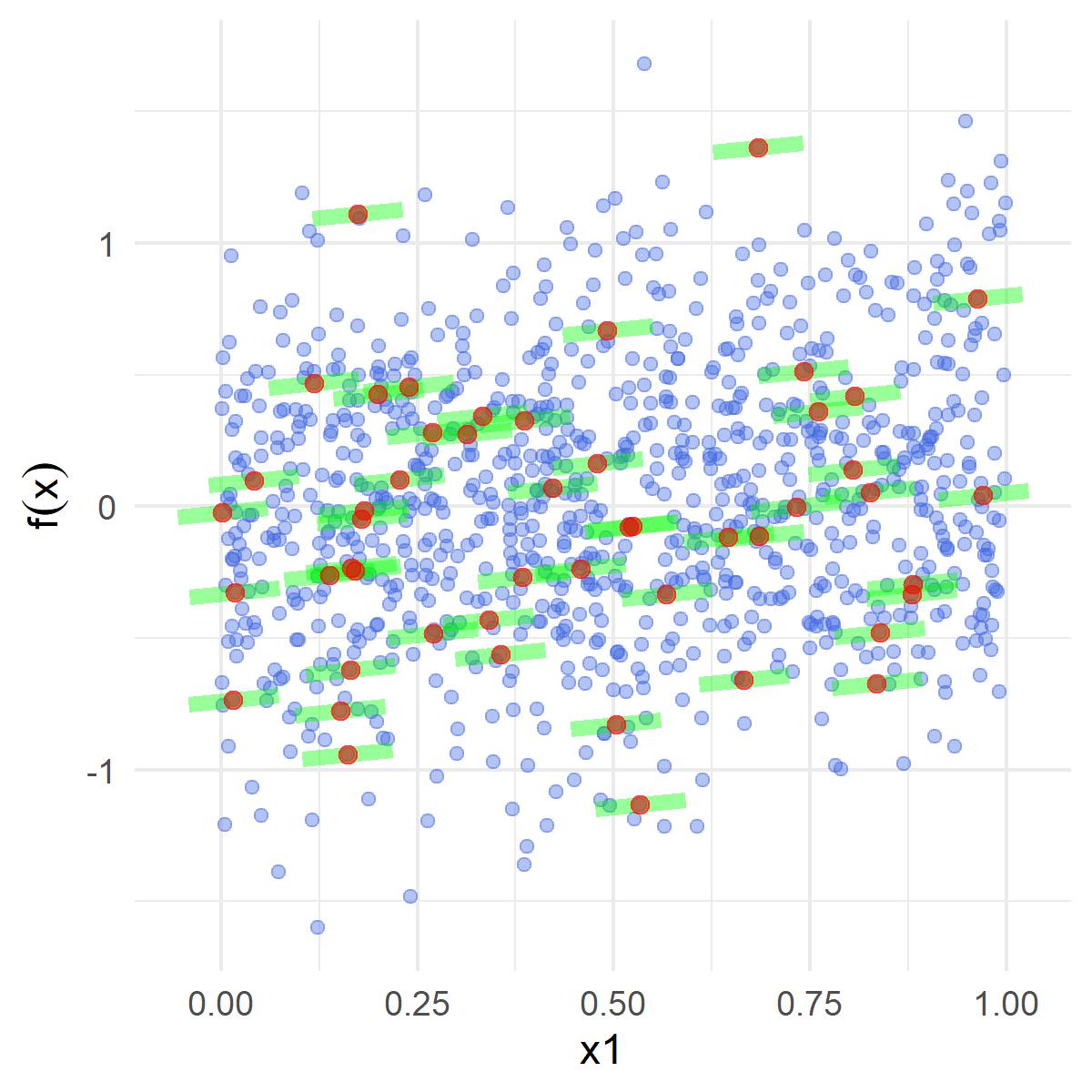}
            \caption[]%
            {{\small IML}} 
        \end{subfigure}
        \begin{subfigure}[b]{0.32\textwidth}  
            \centering 
            \includegraphics[width=\textwidth]{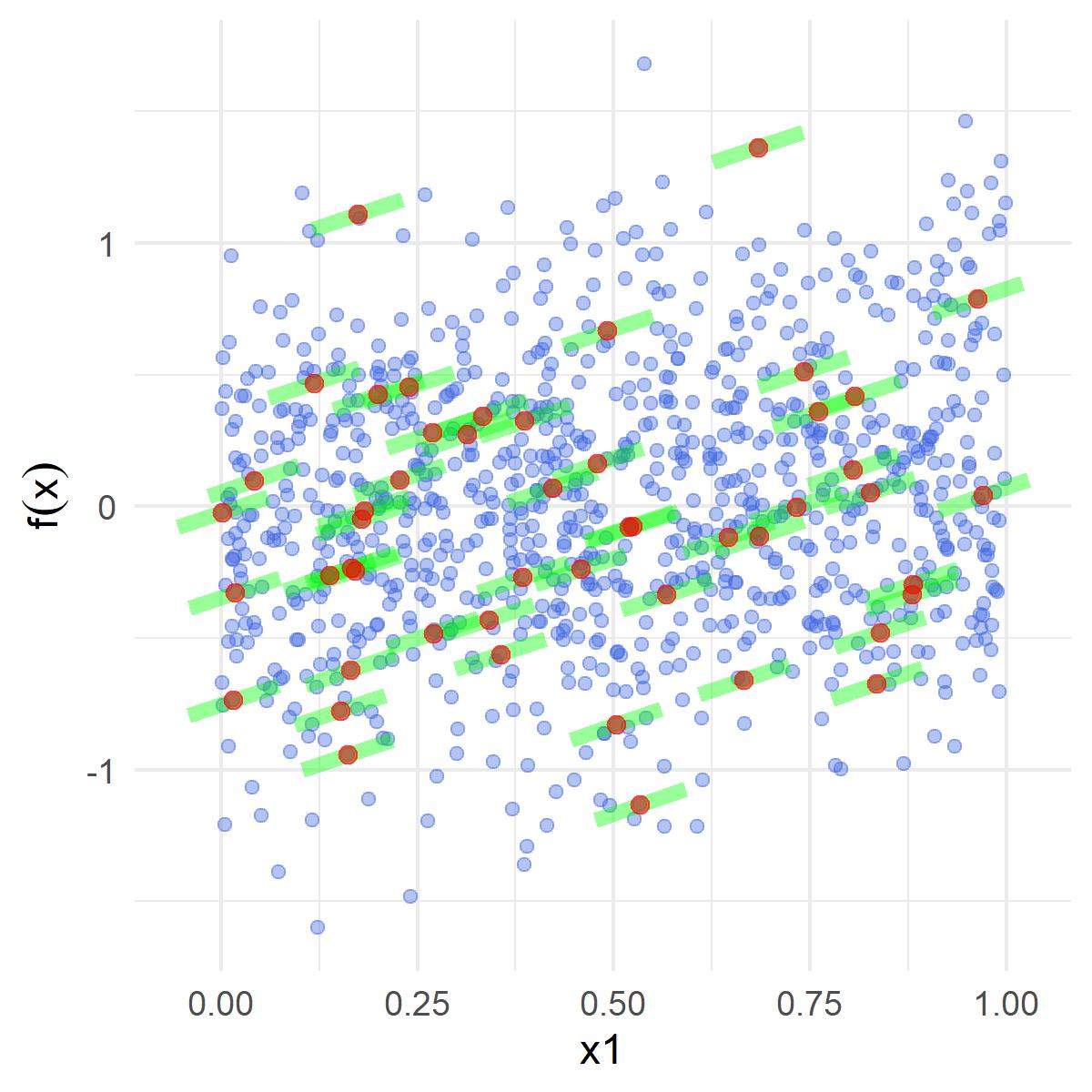} 
            \caption[]%
            {{\small True}} 
        \end{subfigure}
        \caption[]
        {\small Local slope plots (green) for sample points (red) from the dataset 3.} 
\end{figure}

\begin{figure}[H]
        \centering
        \begin{subfigure}[b]{0.32\textwidth}
            \centering
            \includegraphics[width=\textwidth]{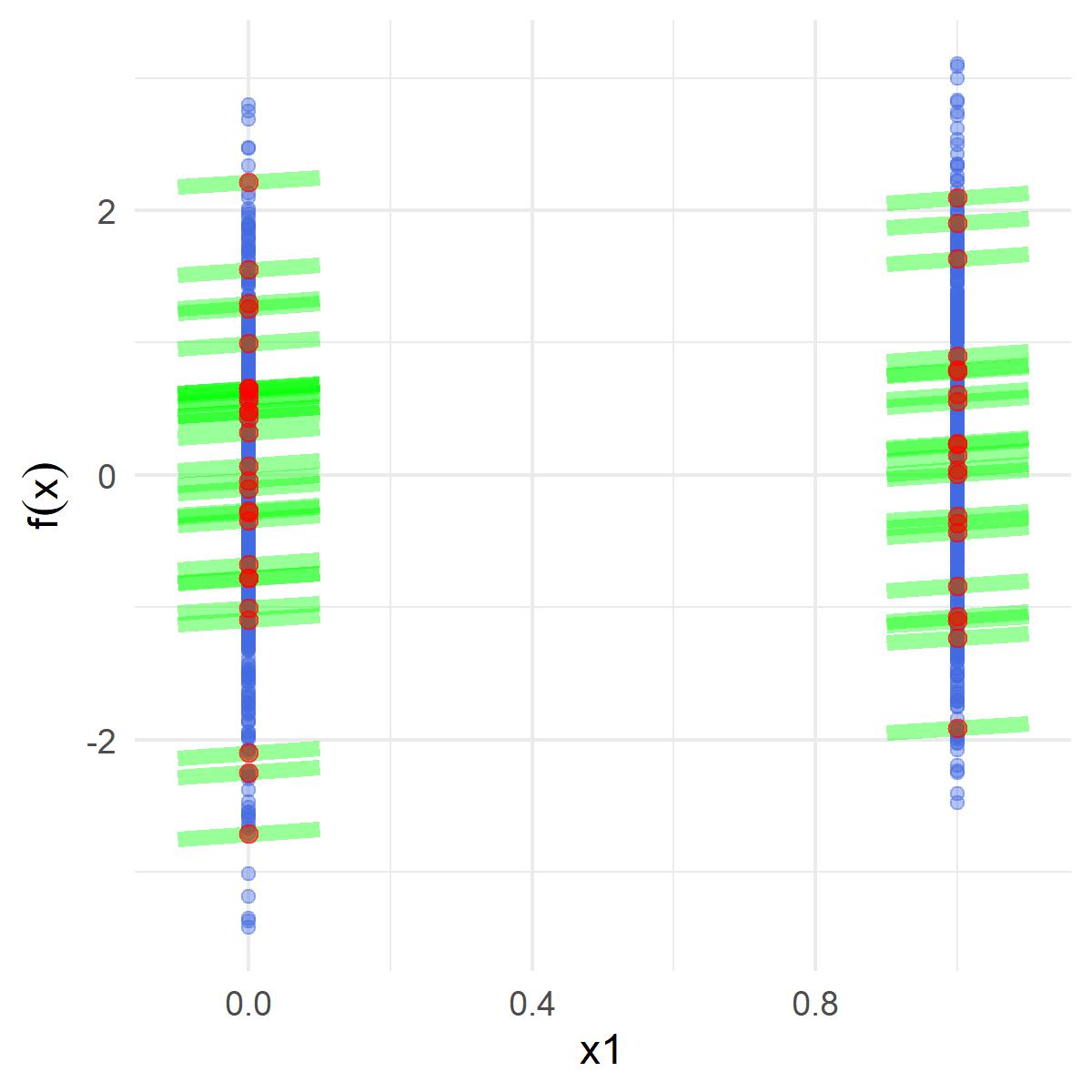}
            \caption[]%
            {{\small SupClus}}   
        \end{subfigure}
        \begin{subfigure}[b]{0.32\textwidth}  
            \centering 
            \includegraphics[width=\textwidth]{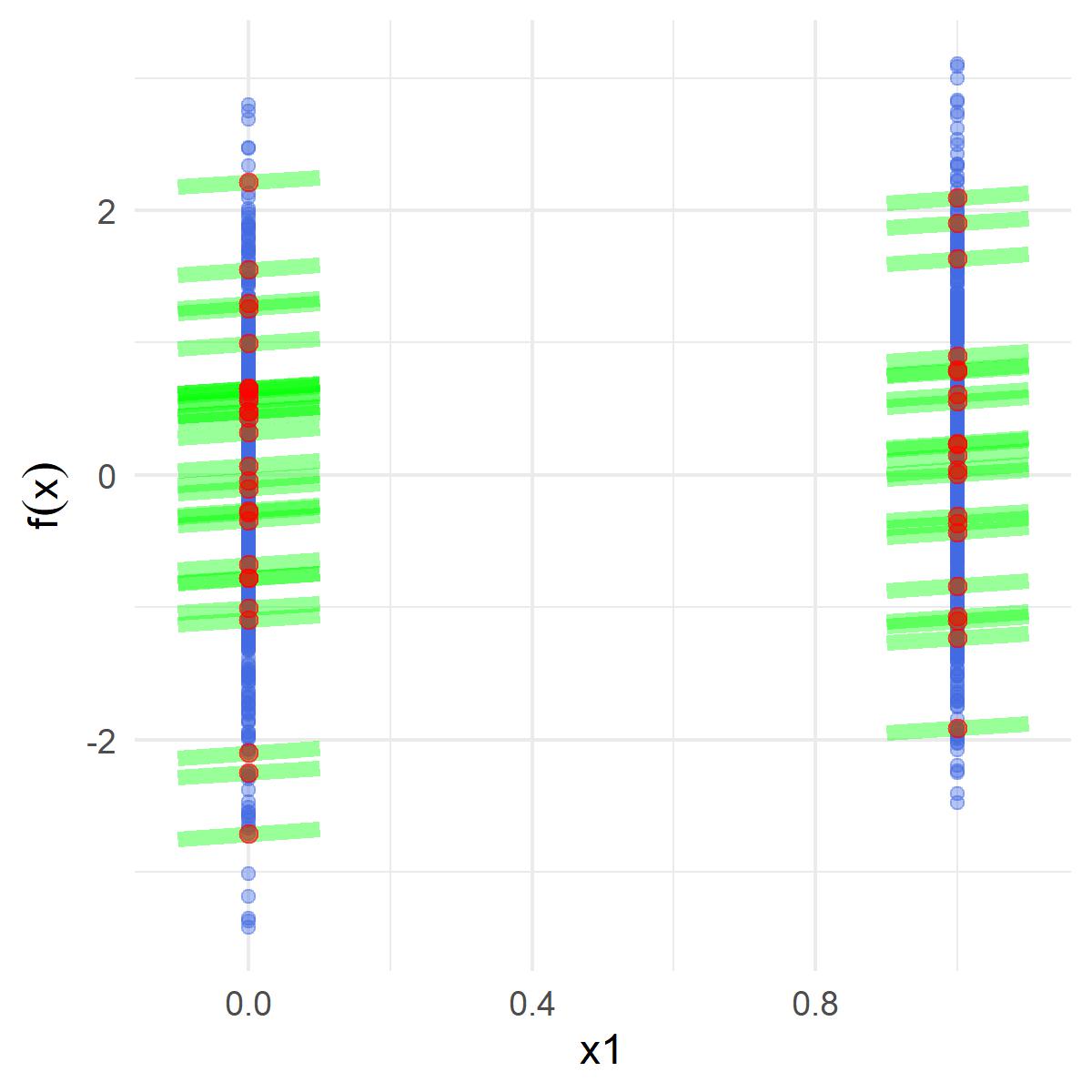}
            \caption[]%
            {{\small VarImp}} 
        \end{subfigure}
        \begin{subfigure}[b]{0.32\textwidth}  
            \centering 
            \includegraphics[width=\textwidth]{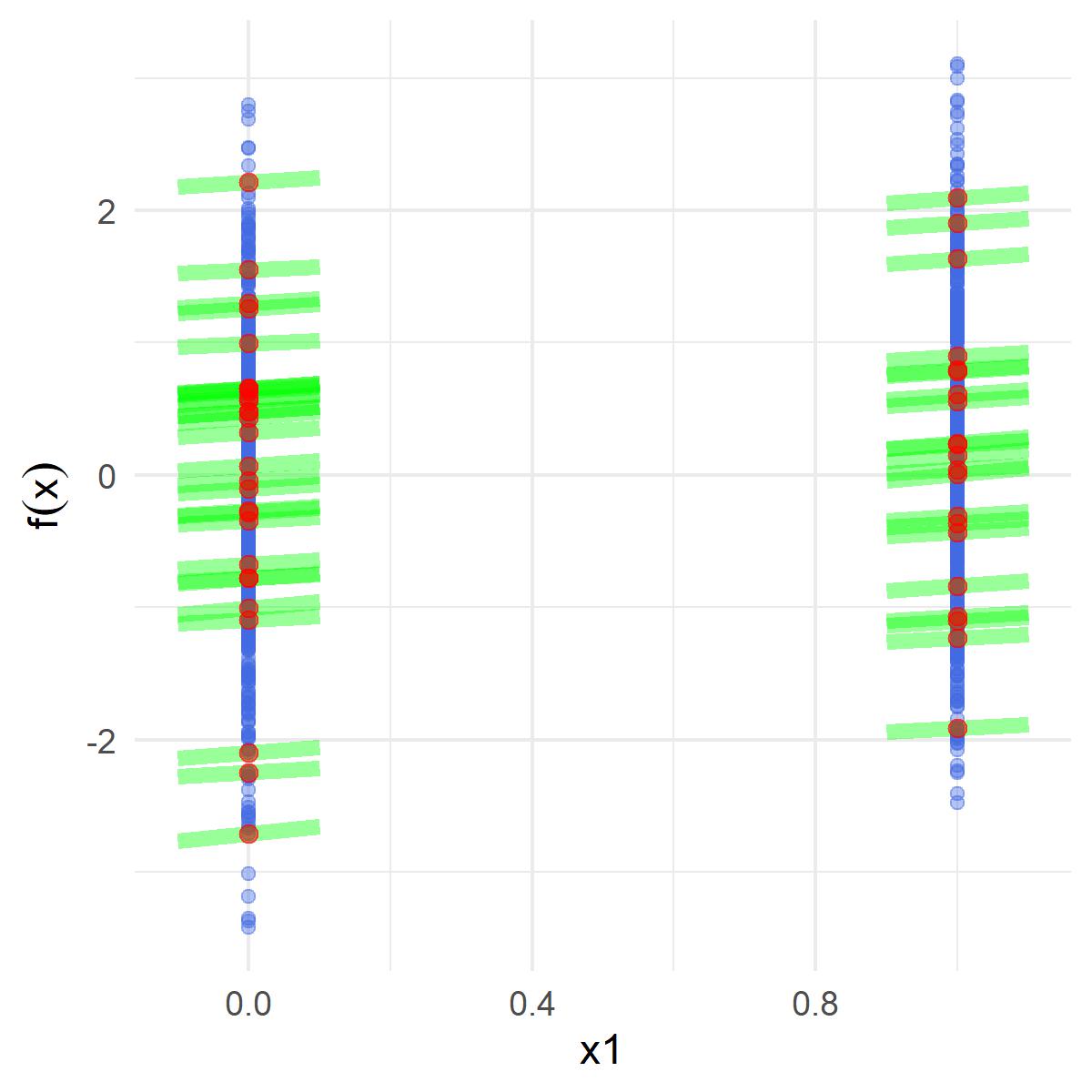}
            \caption[]%
            {{\small LIME}} 
        \end{subfigure}
        \hfill
        \begin{subfigure}[b]{0.32\textwidth}  
            \centering 
            \includegraphics[width=\textwidth]{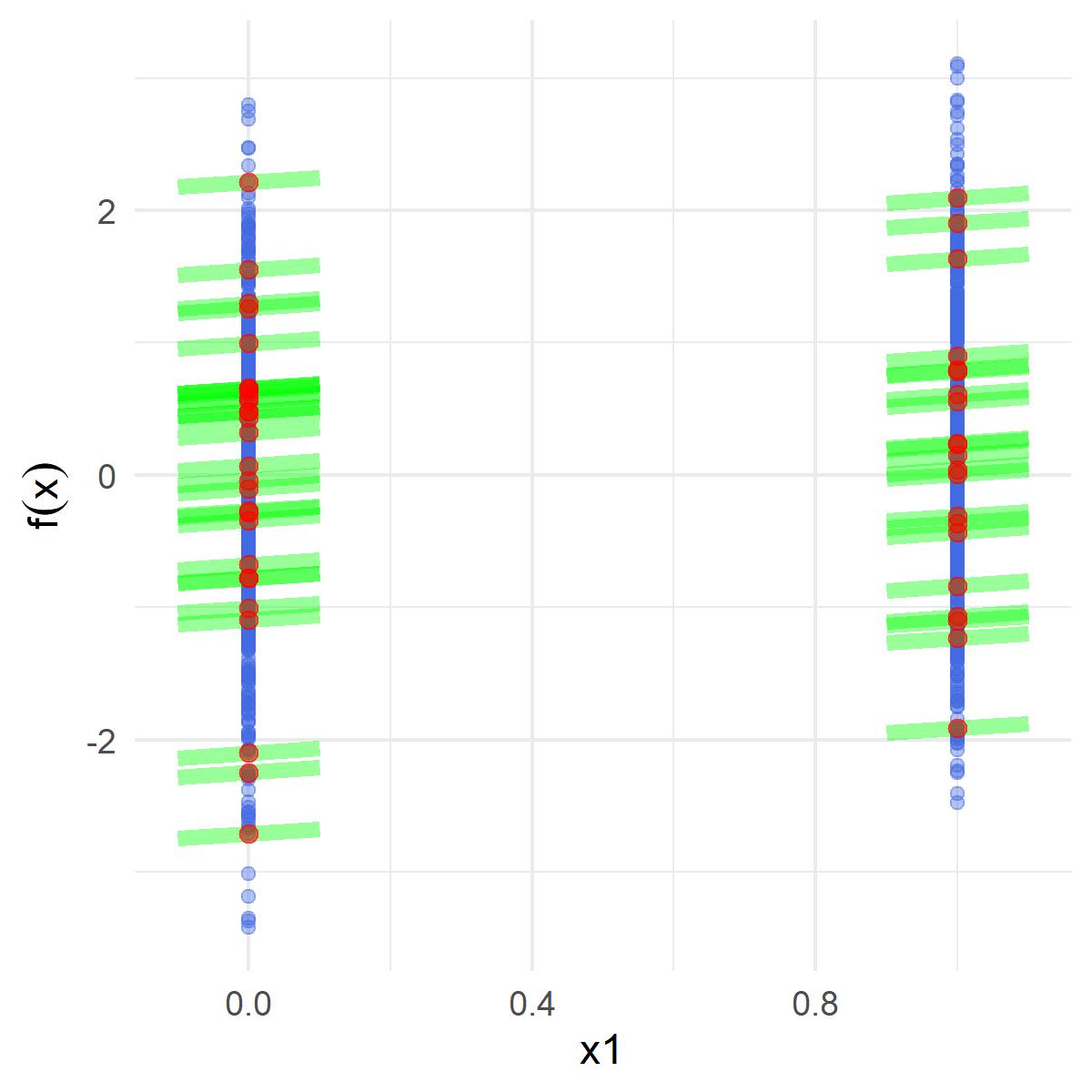}
            \caption[]%
            {{\small IML}} 
        \end{subfigure}
        \begin{subfigure}[b]{0.32\textwidth}  
            \centering 
            \includegraphics[width=\textwidth]{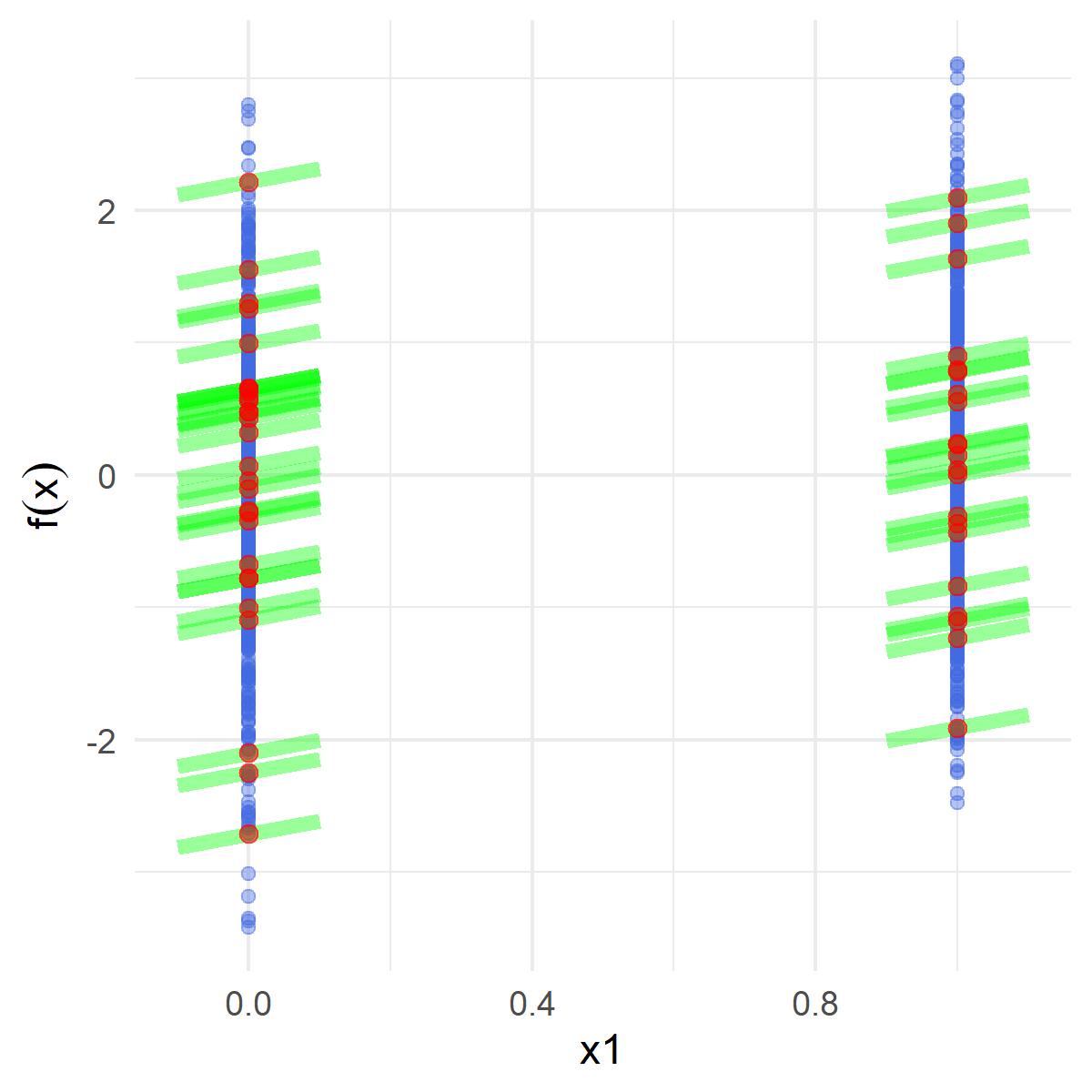} 
            \caption[]%
            {{\small True}} 
        \end{subfigure}
        \caption[]
        {\small Local slope plots (green) for sample points (red) from the dataset 4.} 
\end{figure}

\begin{figure}[H]
        \centering
        \begin{subfigure}[b]{0.32\textwidth}
            \centering
            \includegraphics[width=\textwidth]{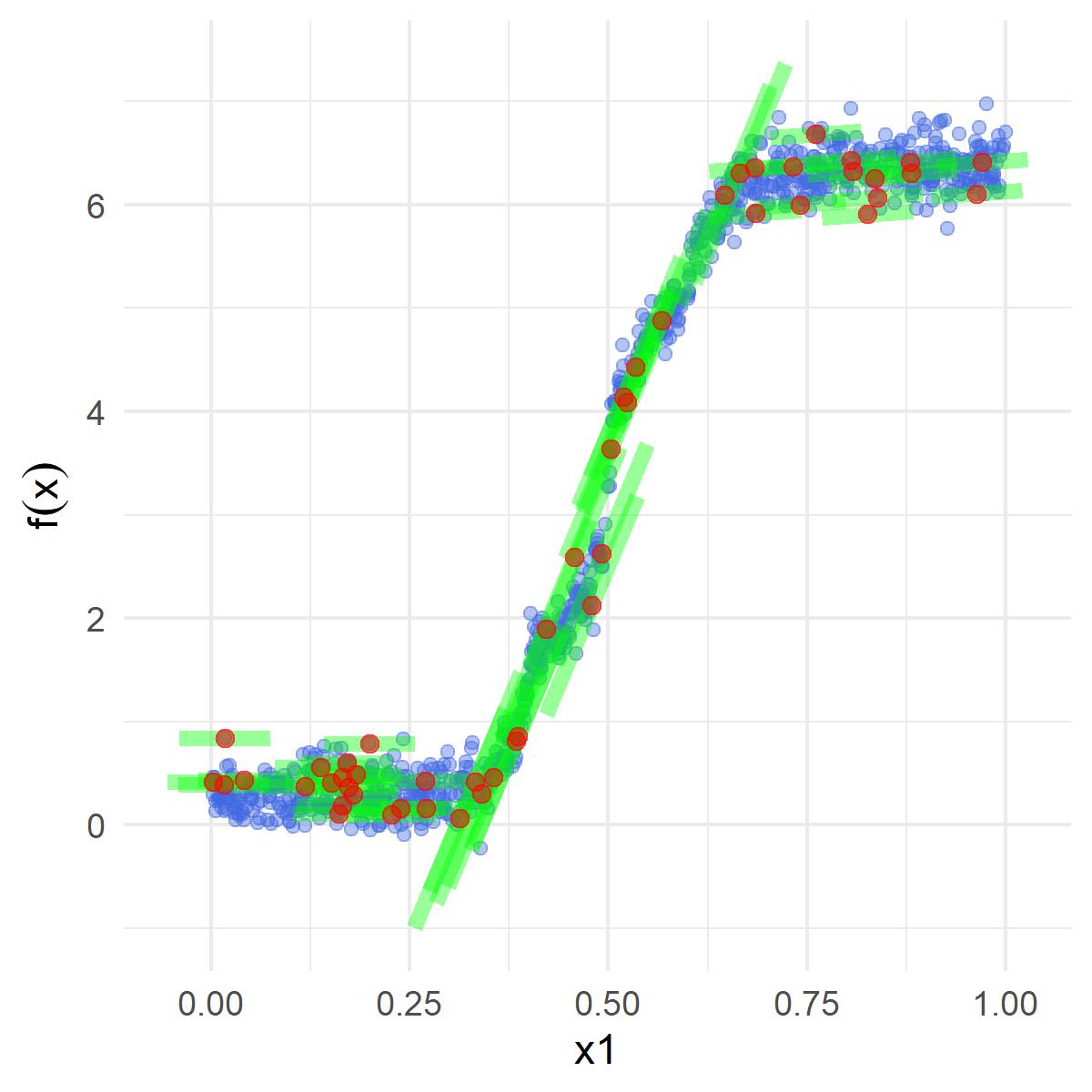}
            \caption[]%
            {{\small SupClus}}   
        \end{subfigure}
        \begin{subfigure}[b]{0.32\textwidth}  
            \centering 
            \includegraphics[width=\textwidth]{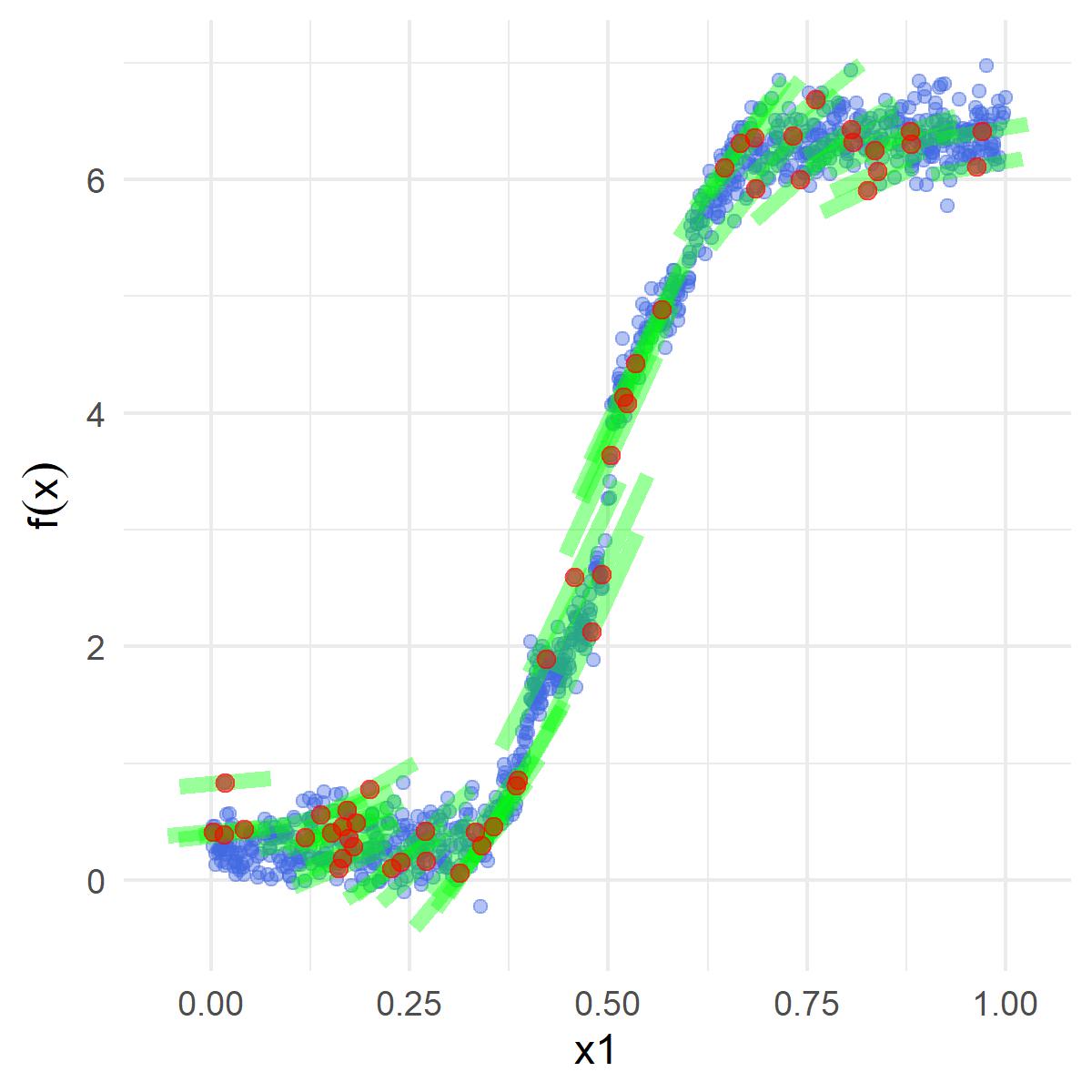}
            \caption[]%
            {{\small VarImp}} 
        \end{subfigure}
        \begin{subfigure}[b]{0.32\textwidth}  
            \centering 
            \includegraphics[width=\textwidth]{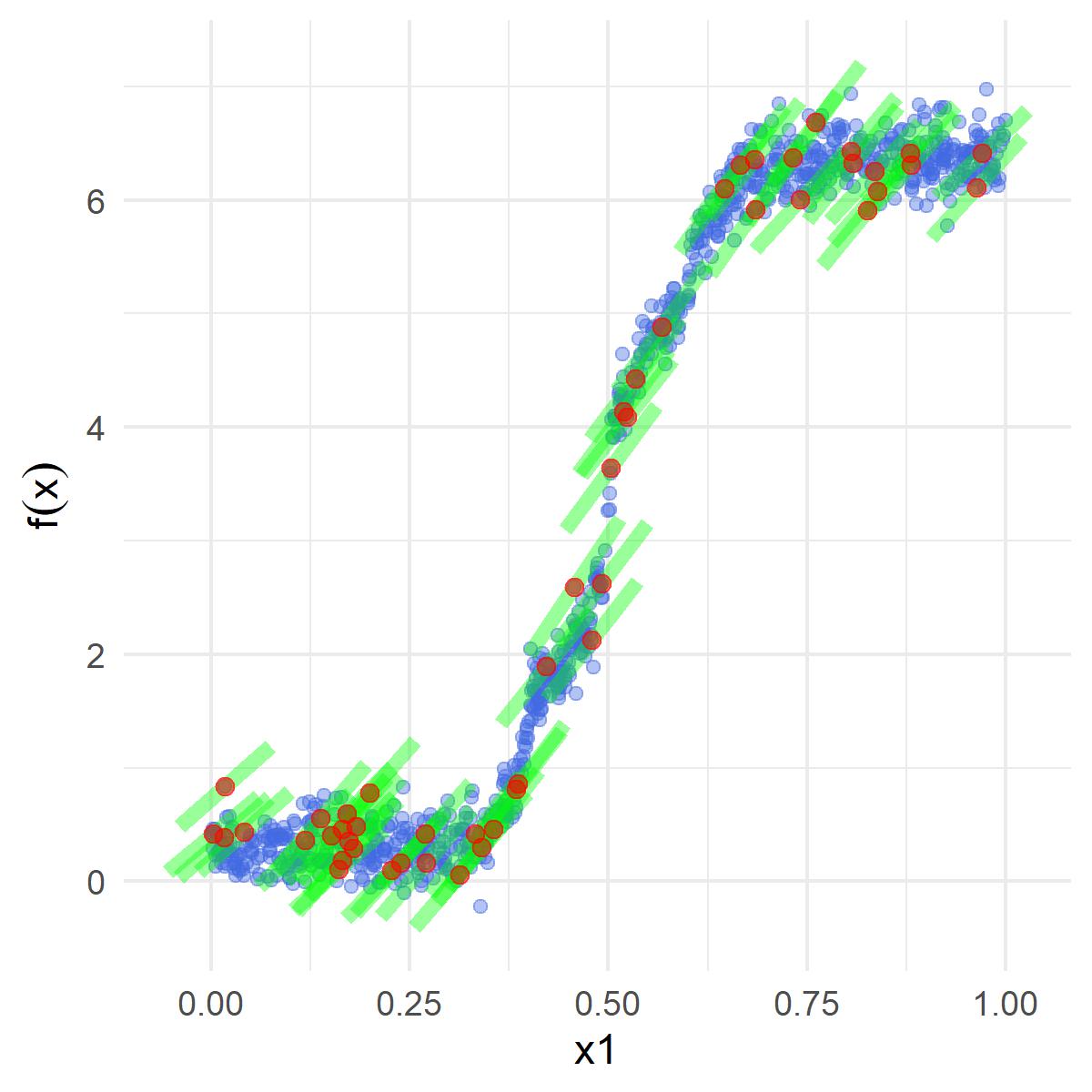}
            \caption[]%
            {{\small LIME}} 
        \end{subfigure}
        \hfill
        \begin{subfigure}[b]{0.32\textwidth}  
            \centering 
            \includegraphics[width=\textwidth]{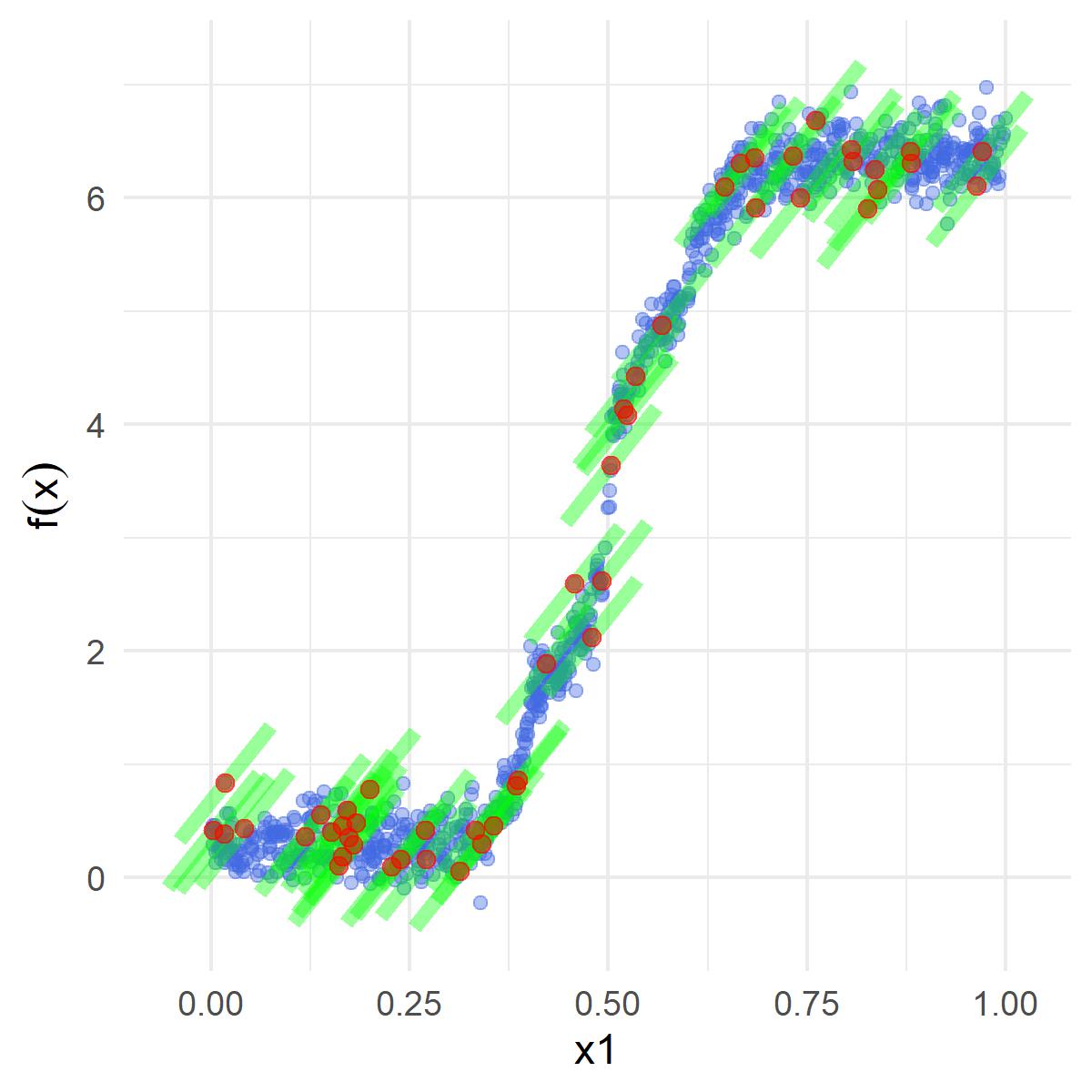}
            \caption[]%
            {{\small IML}} 
        \end{subfigure}
        \begin{subfigure}[b]{0.32\textwidth}  
            \centering 
            \includegraphics[width=\textwidth]{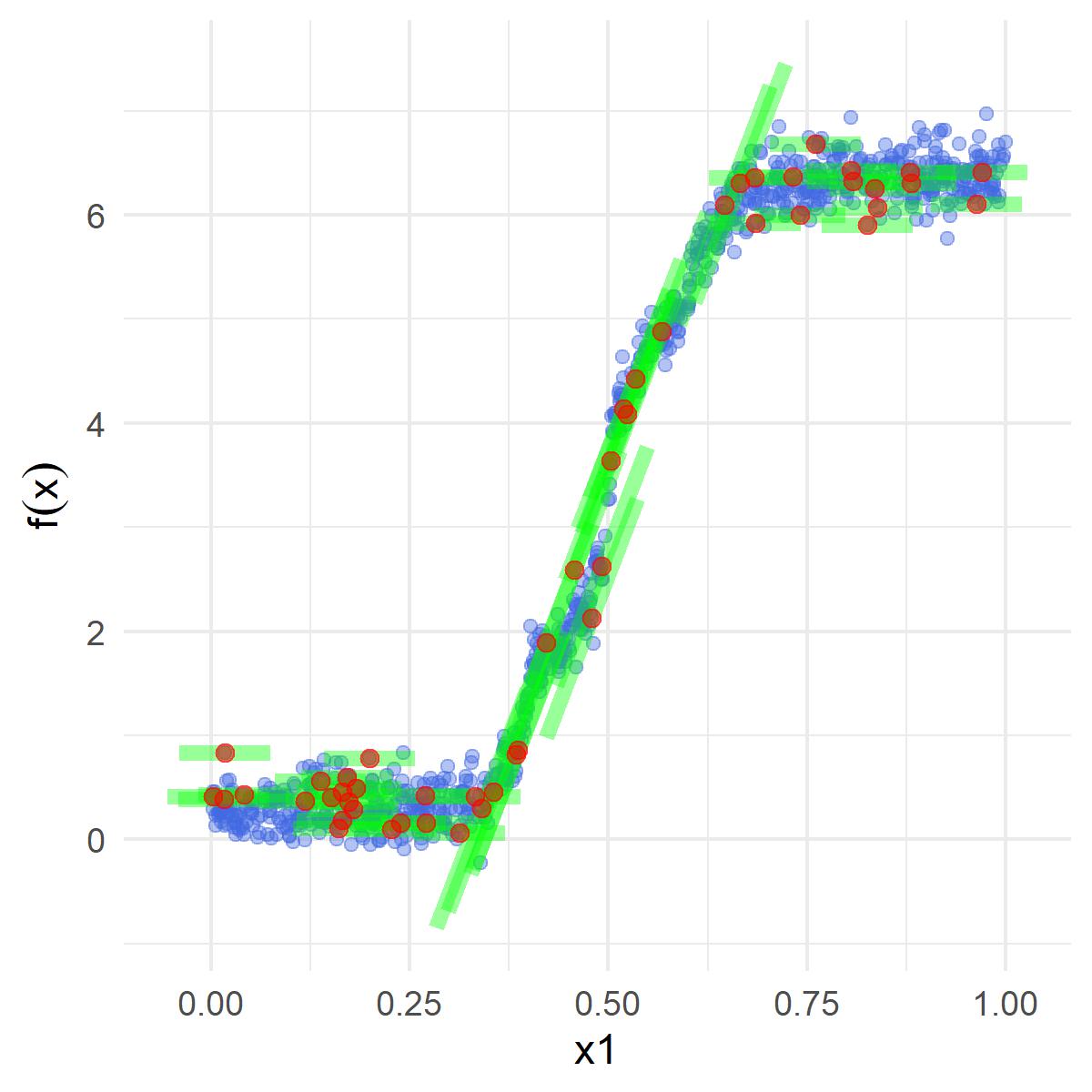} 
            \caption[]%
            {{\small True}} 
        \end{subfigure}
        \caption[]
        {\small Local slope plots (green) for sample points (red) from the dataset 5.} 
\end{figure}

\section{Estimated effects of an instance for artificial datasets.}\label{sec:effectinstance}

\begin{figure}[H]
        \centering
        \begin{subfigure}[b]{0.32\textwidth}
            \centering
            \includegraphics[width=\textwidth]{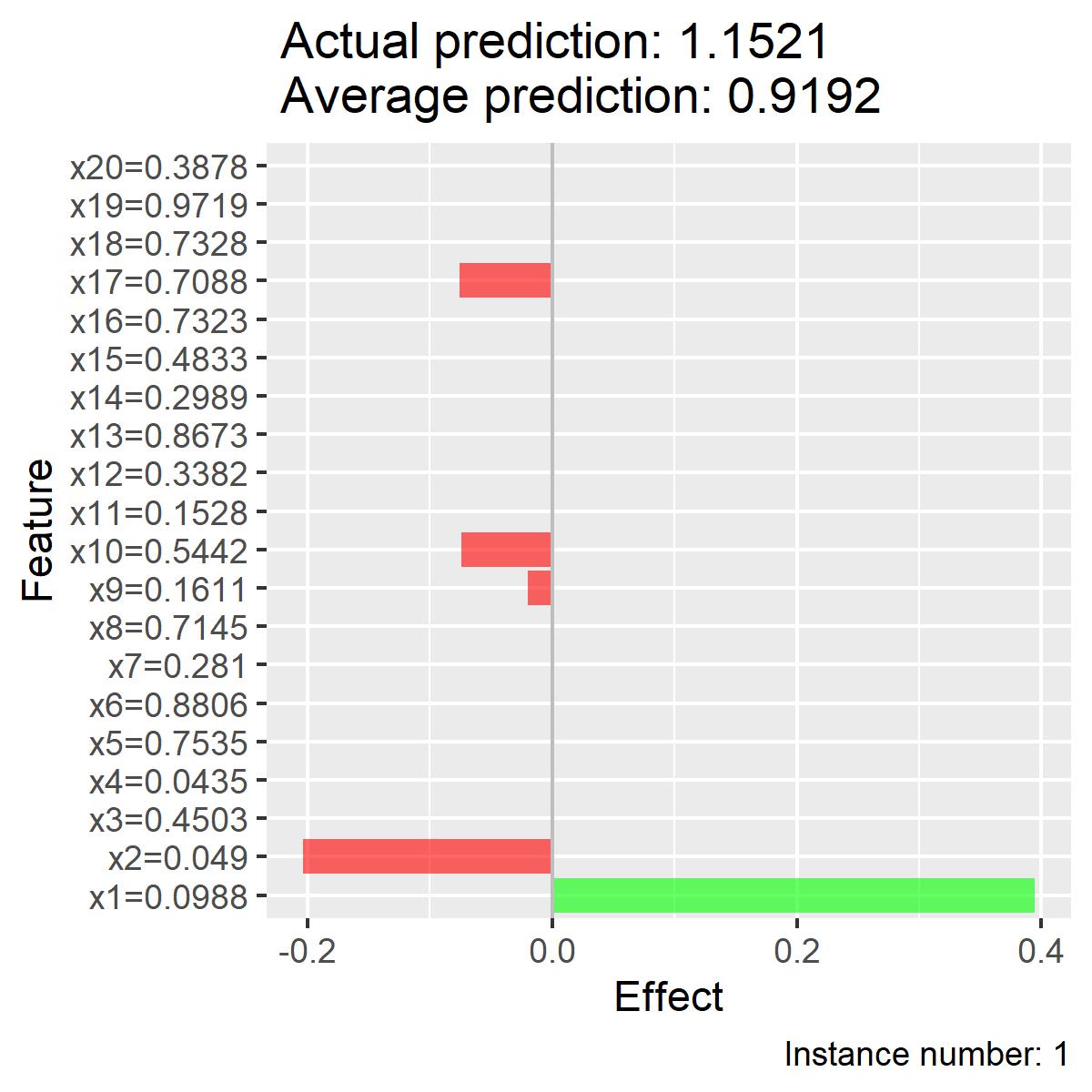}
            \caption[]%
            {{\small VarImp}}   
        \end{subfigure}
        \begin{subfigure}[b]{0.32\textwidth}  
            \centering 
            \includegraphics[width=\textwidth]{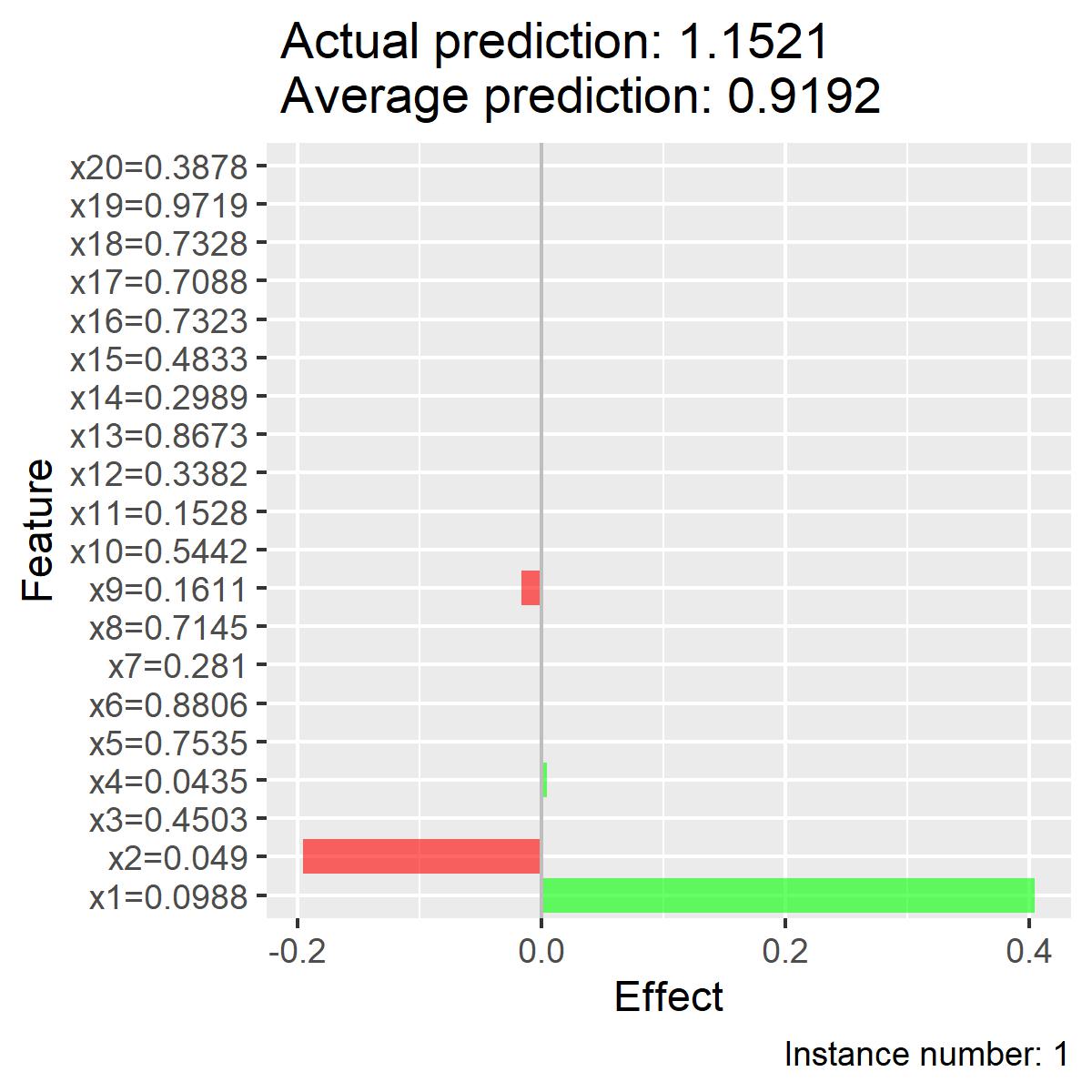}
            \caption[]%
            {{\small SupClus}} 
        \end{subfigure}
        \begin{subfigure}[b]{0.32\textwidth}  
            \centering 
            \includegraphics[width=\textwidth]{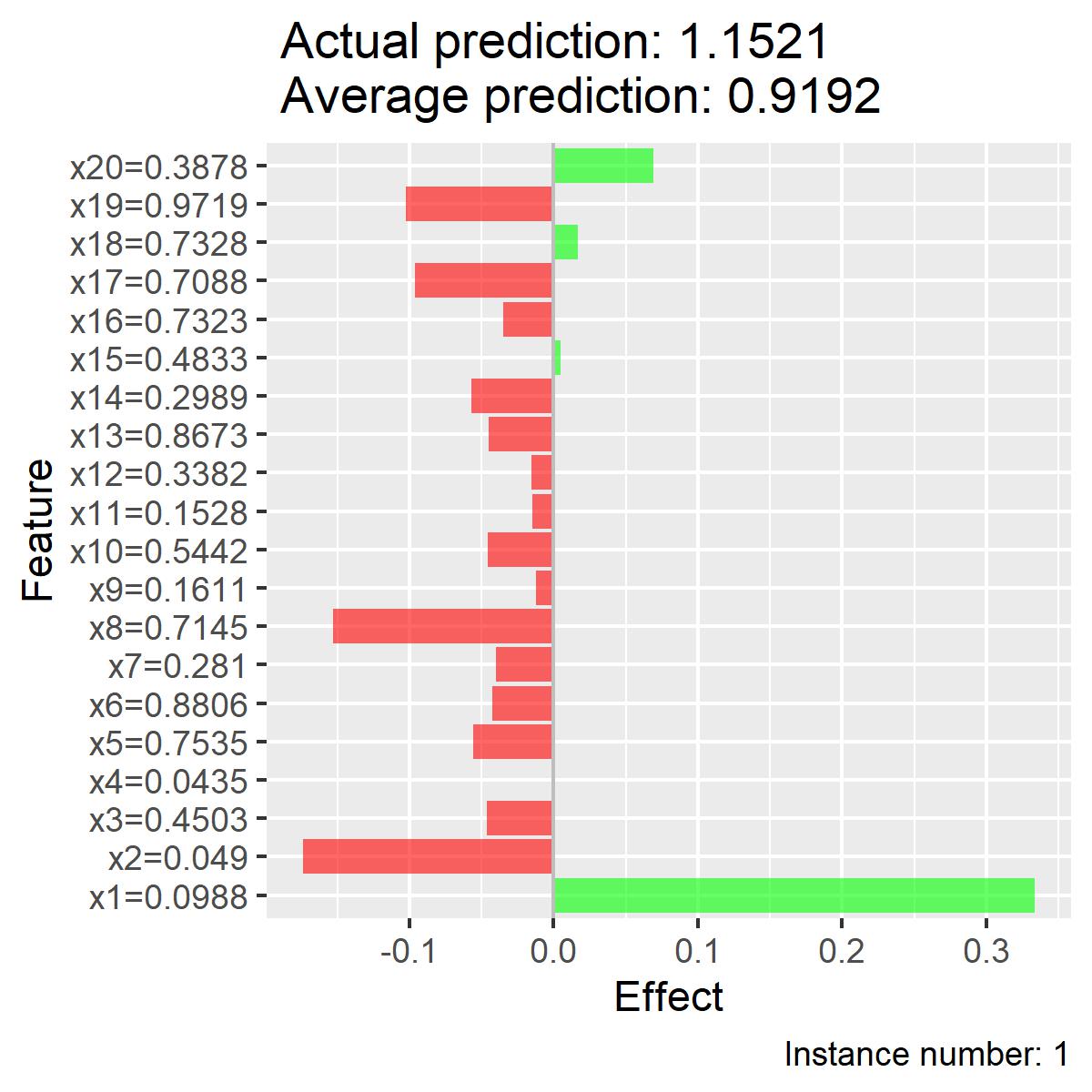}
            \caption[]%
            {{\small LIME}} 
        \end{subfigure}
        \hfill
        \begin{subfigure}[b]{0.32\textwidth}  
            \centering 
            \includegraphics[width=\textwidth]{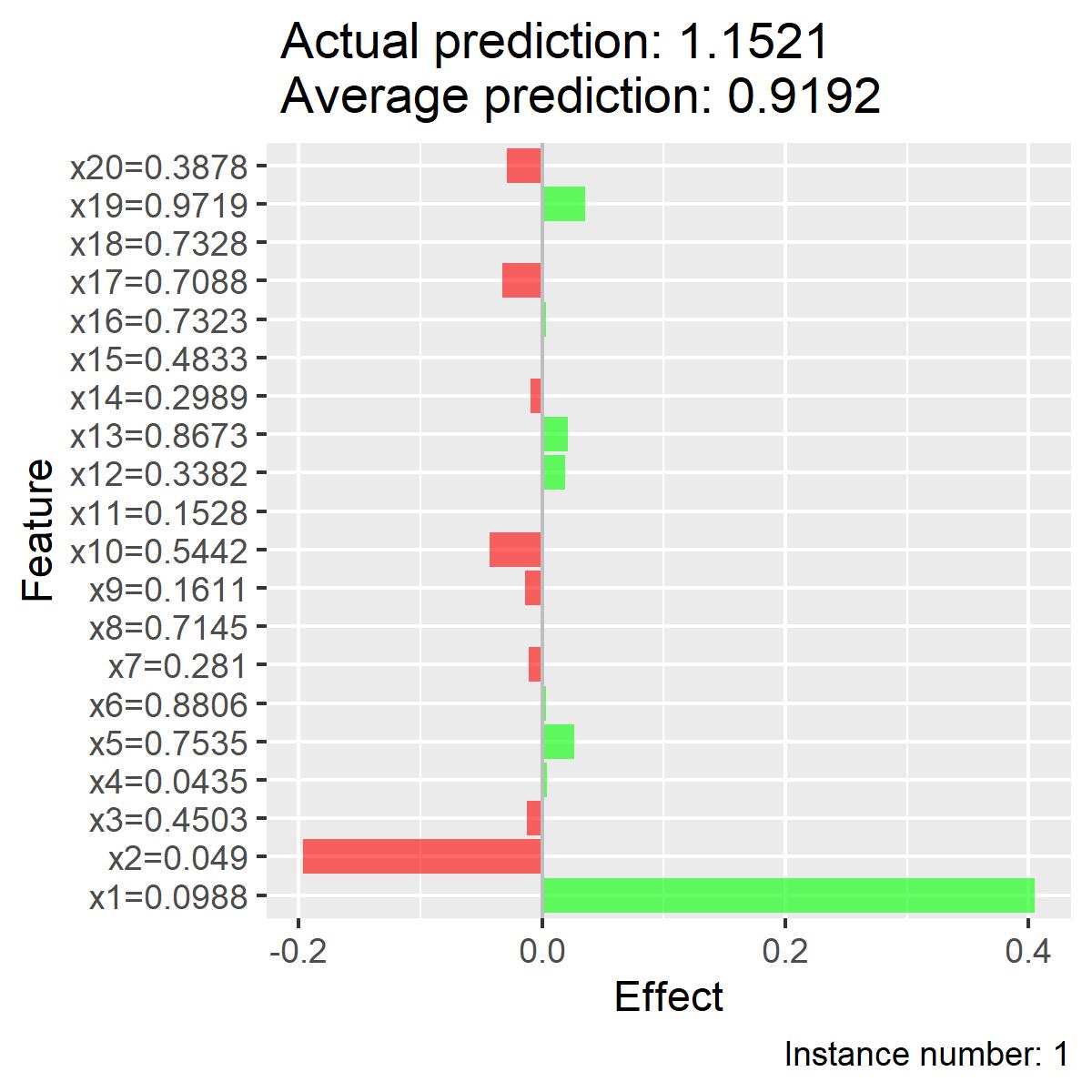}
            \caption[]%
            {{\small IML}} 
        \end{subfigure}
        \begin{subfigure}[b]{0.32\textwidth}  
            \centering 
            \includegraphics[width=\textwidth]{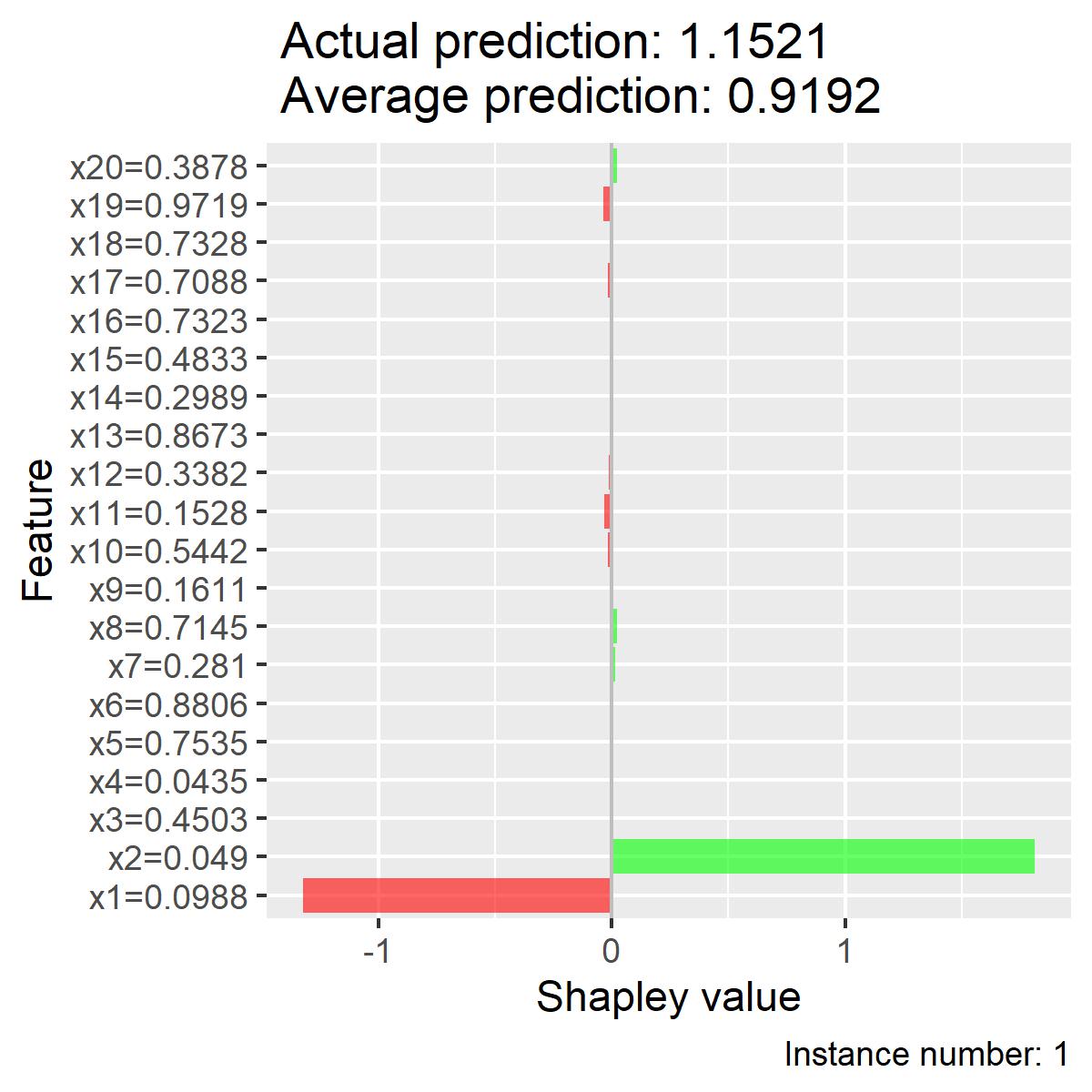}
            \caption[]%
            {{\small Shapley values}} 
        \end{subfigure}
        \caption[]
        {\small Estimated effects of an instance for dataset 1.} 
\end{figure}

\begin{figure}[H]
        \centering
        \begin{subfigure}[b]{0.32\textwidth}
            \centering
            \includegraphics[width=\textwidth]{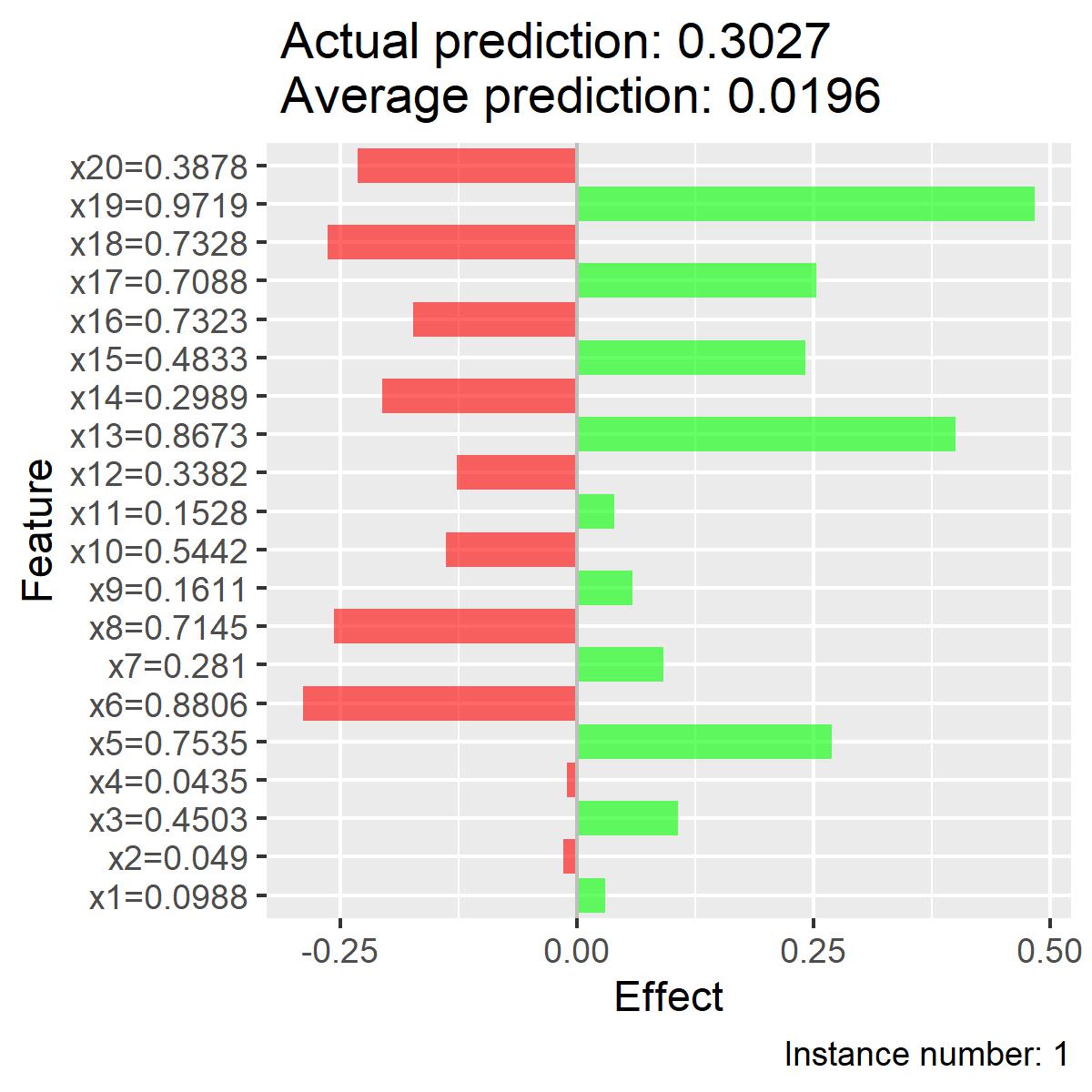}
            \caption[]%
            {{\small VarImp}}   
        \end{subfigure}
        \begin{subfigure}[b]{0.32\textwidth}  
            \centering 
            \includegraphics[width=\textwidth]{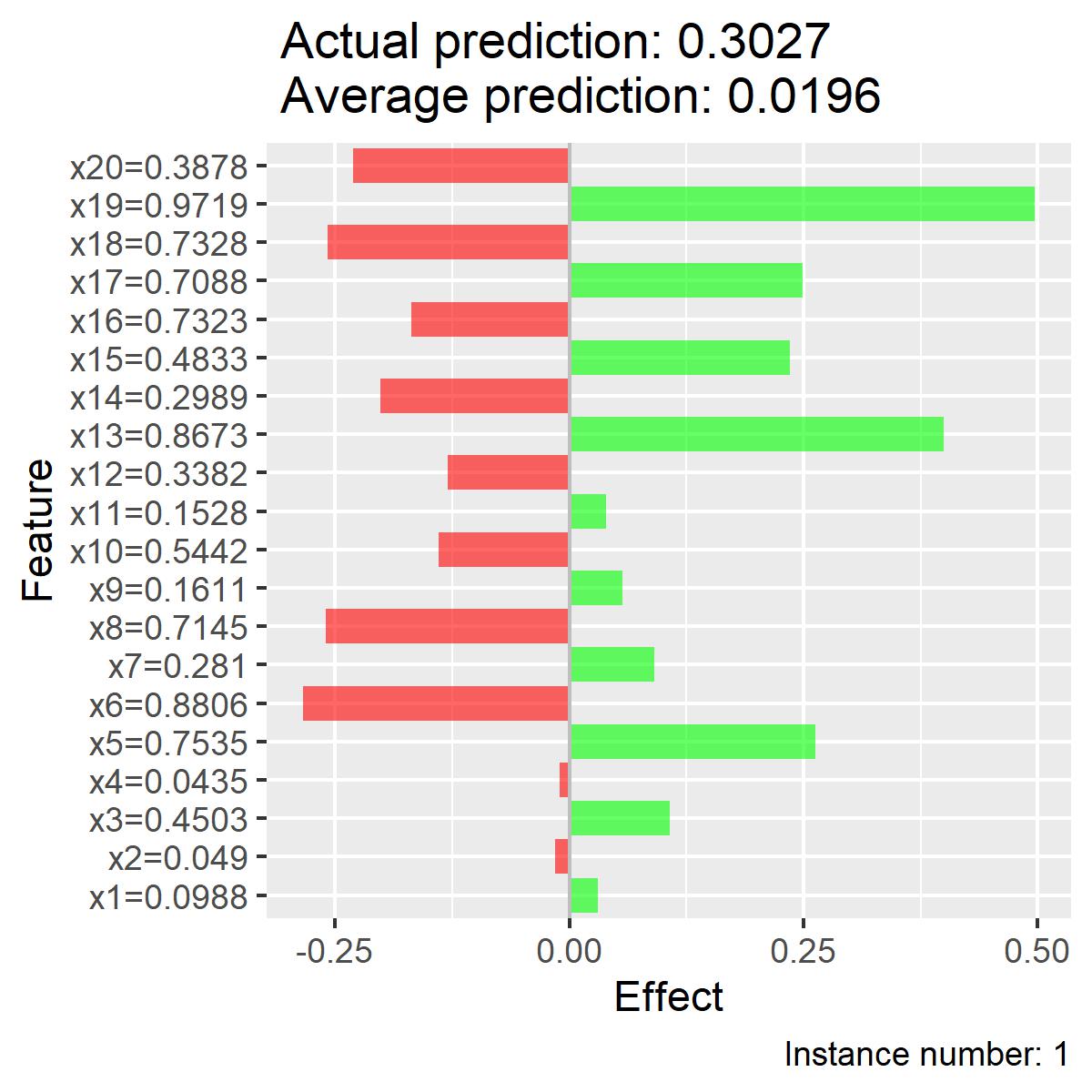}
            \caption[]%
            {{\small SupClus}} 
        \end{subfigure}
        \begin{subfigure}[b]{0.32\textwidth}  
            \centering 
            \includegraphics[width=\textwidth]{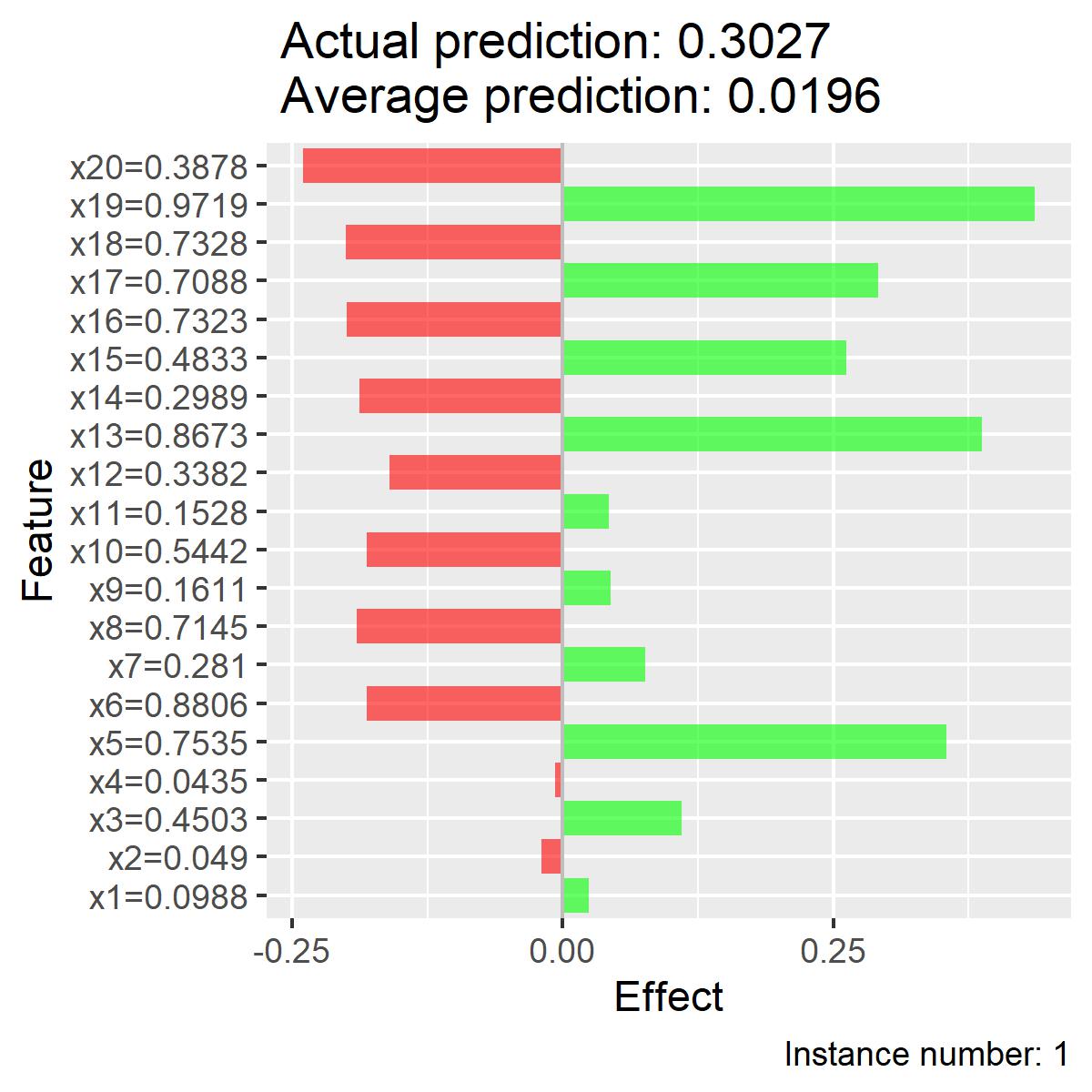}
            \caption[]%
            {{\small LIME}} 
        \end{subfigure}
        \hfill
        \begin{subfigure}[b]{0.32\textwidth}  
            \centering 
            \includegraphics[width=\textwidth]{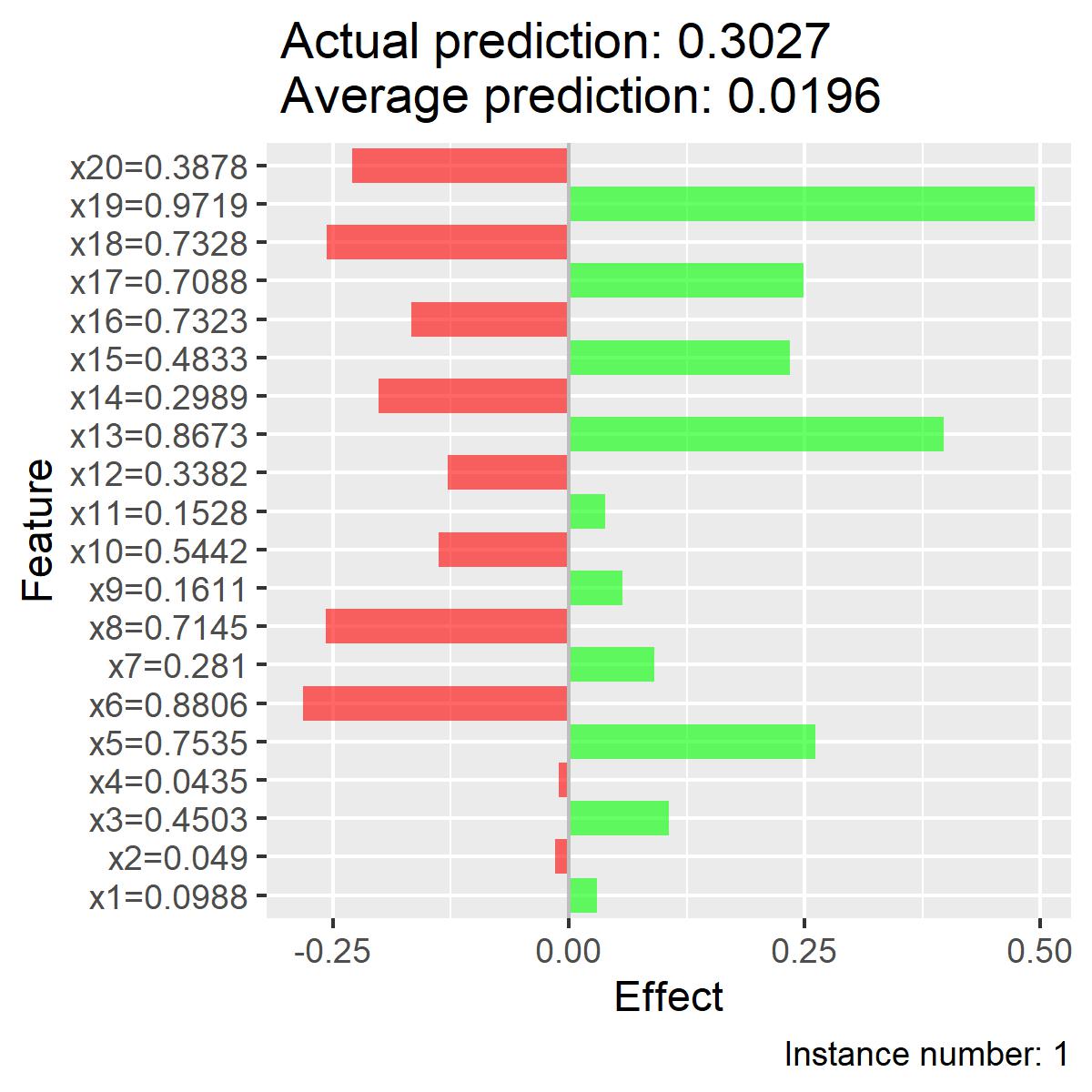}
            \caption[]%
            {{\small IML}} 
        \end{subfigure}
        \begin{subfigure}[b]{0.32\textwidth}  
            \centering 
            \includegraphics[width=\textwidth]{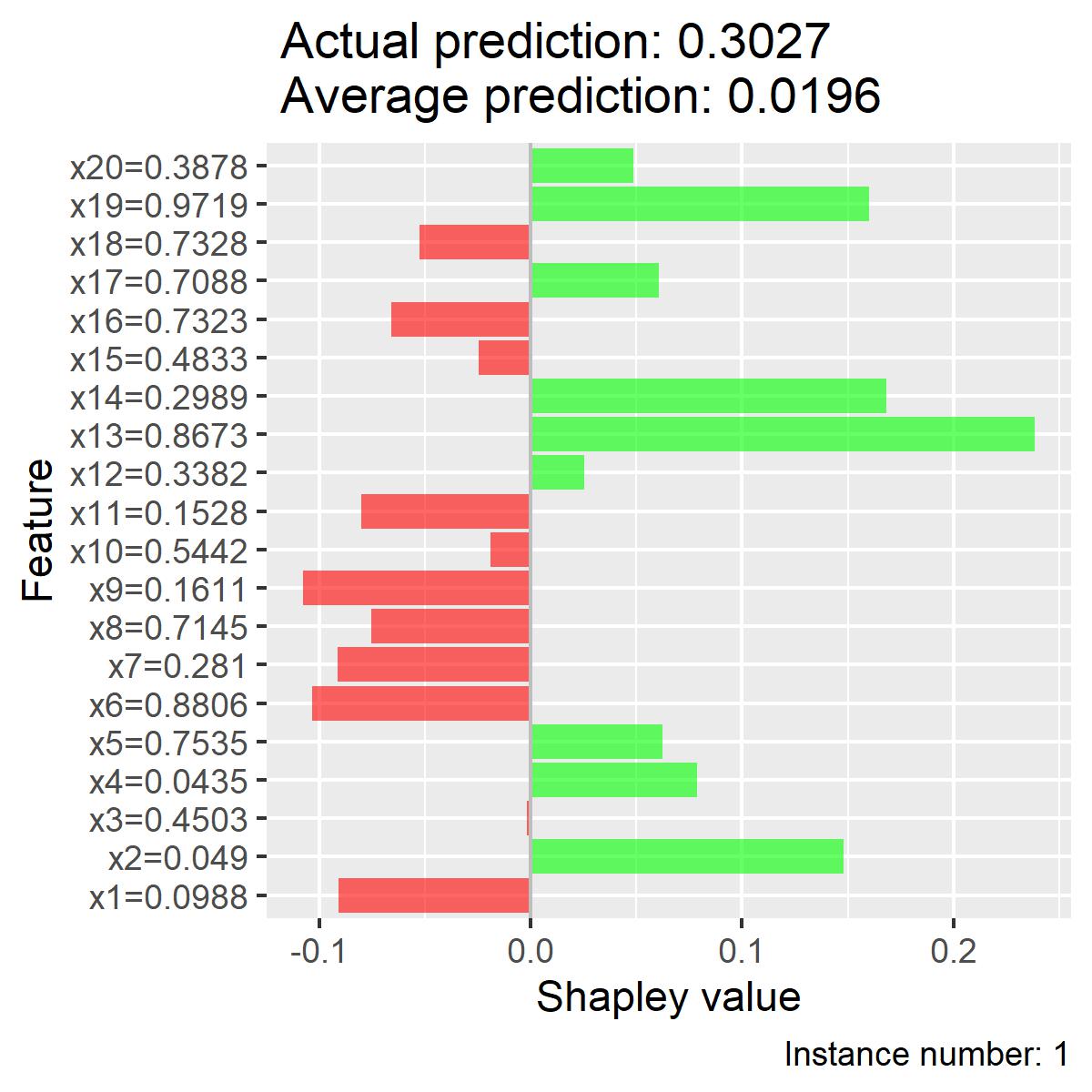}
            \caption[]%
            {{\small Shapley values}} 
        \end{subfigure}
        \caption[]
        {\small Estimated effects of an instance for dataset 3.} 
\end{figure}

\begin{figure}[H]
        \centering
        \begin{subfigure}[b]{0.32\textwidth}
            \centering
            \includegraphics[width=\textwidth]{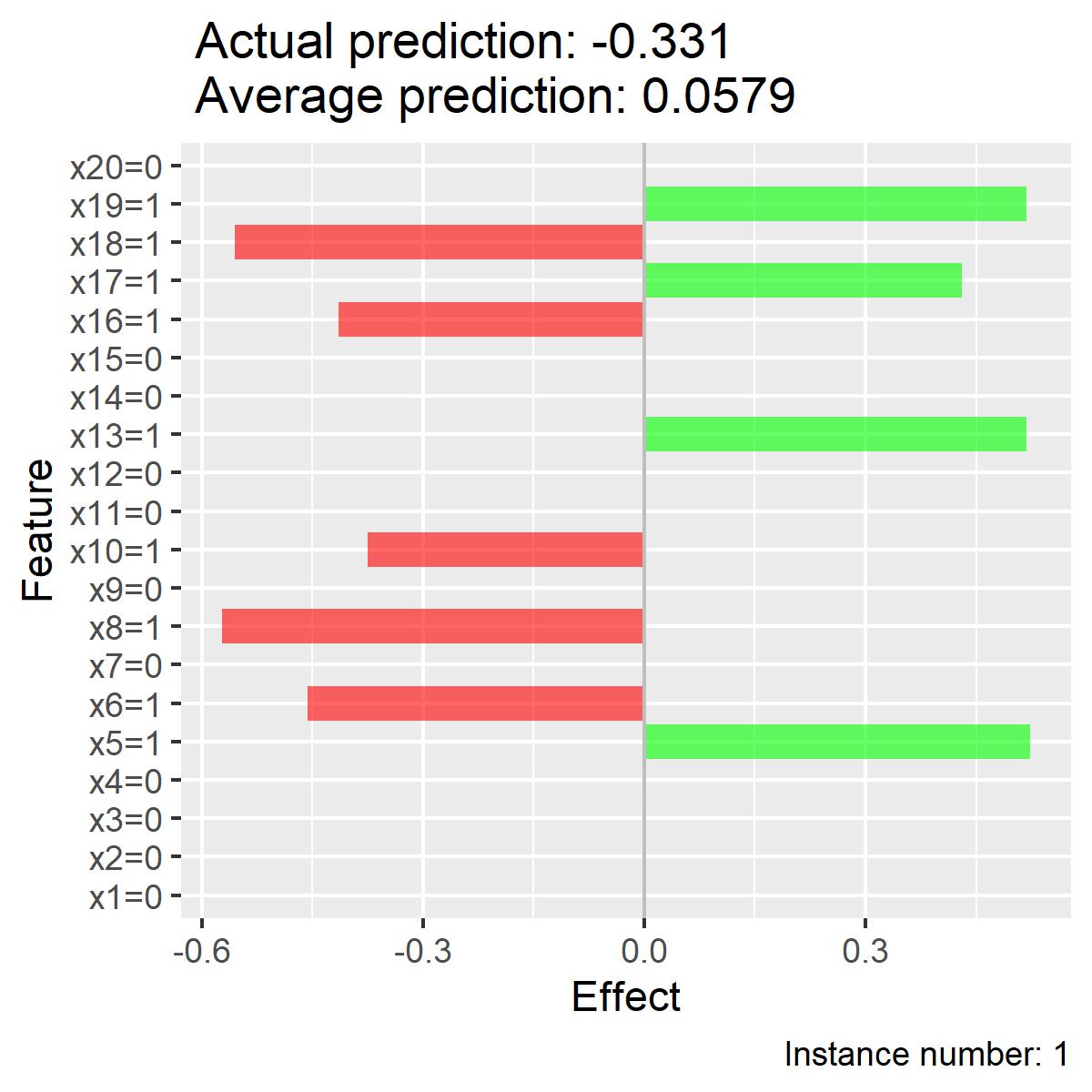}
            \caption[]%
            {{\small VarImp}}   
        \end{subfigure}
        \begin{subfigure}[b]{0.32\textwidth}  
            \centering 
            \includegraphics[width=\textwidth]{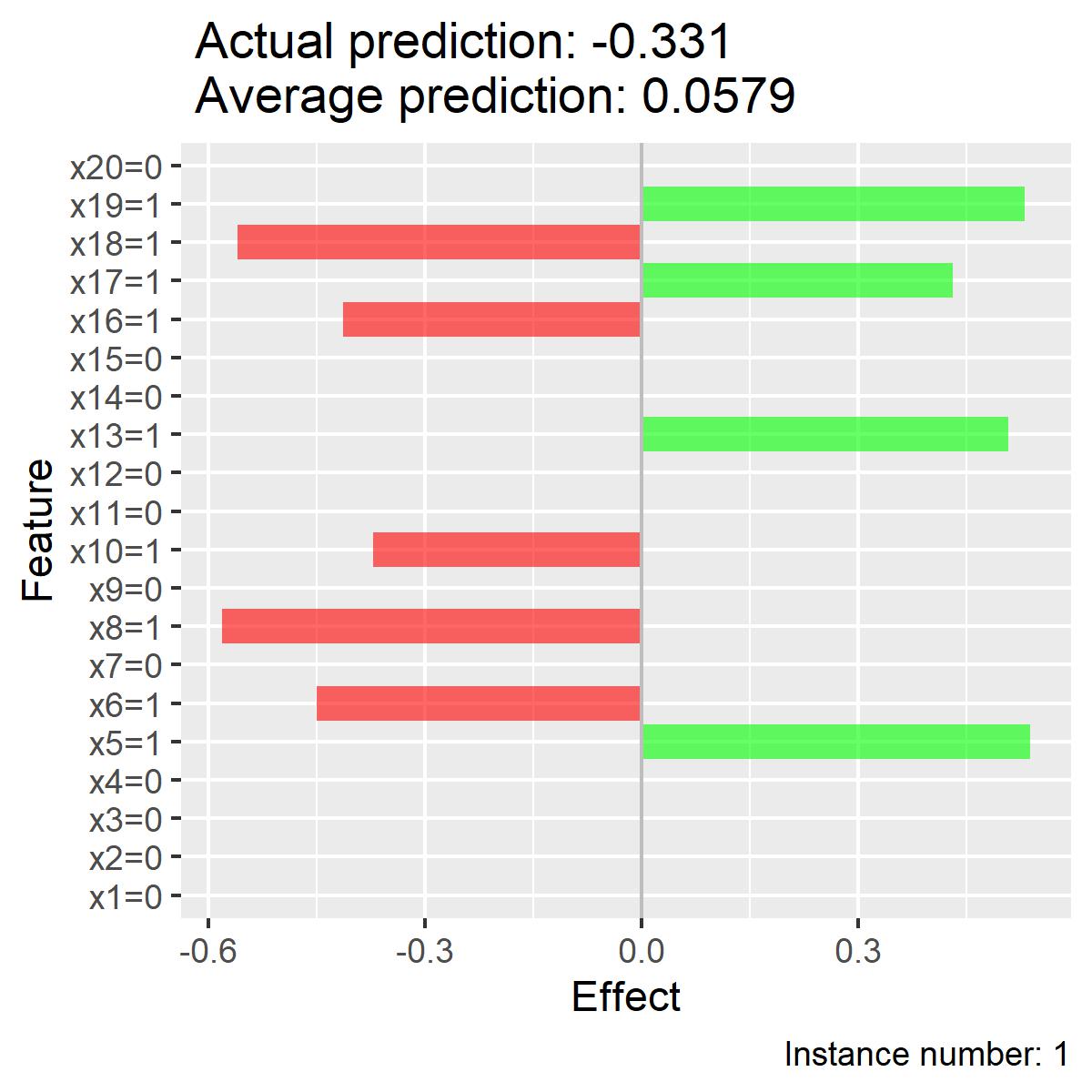}
            \caption[]%
            {{\small SupClus}} 
        \end{subfigure}
        \begin{subfigure}[b]{0.32\textwidth}  
            \centering 
            \includegraphics[width=\textwidth]{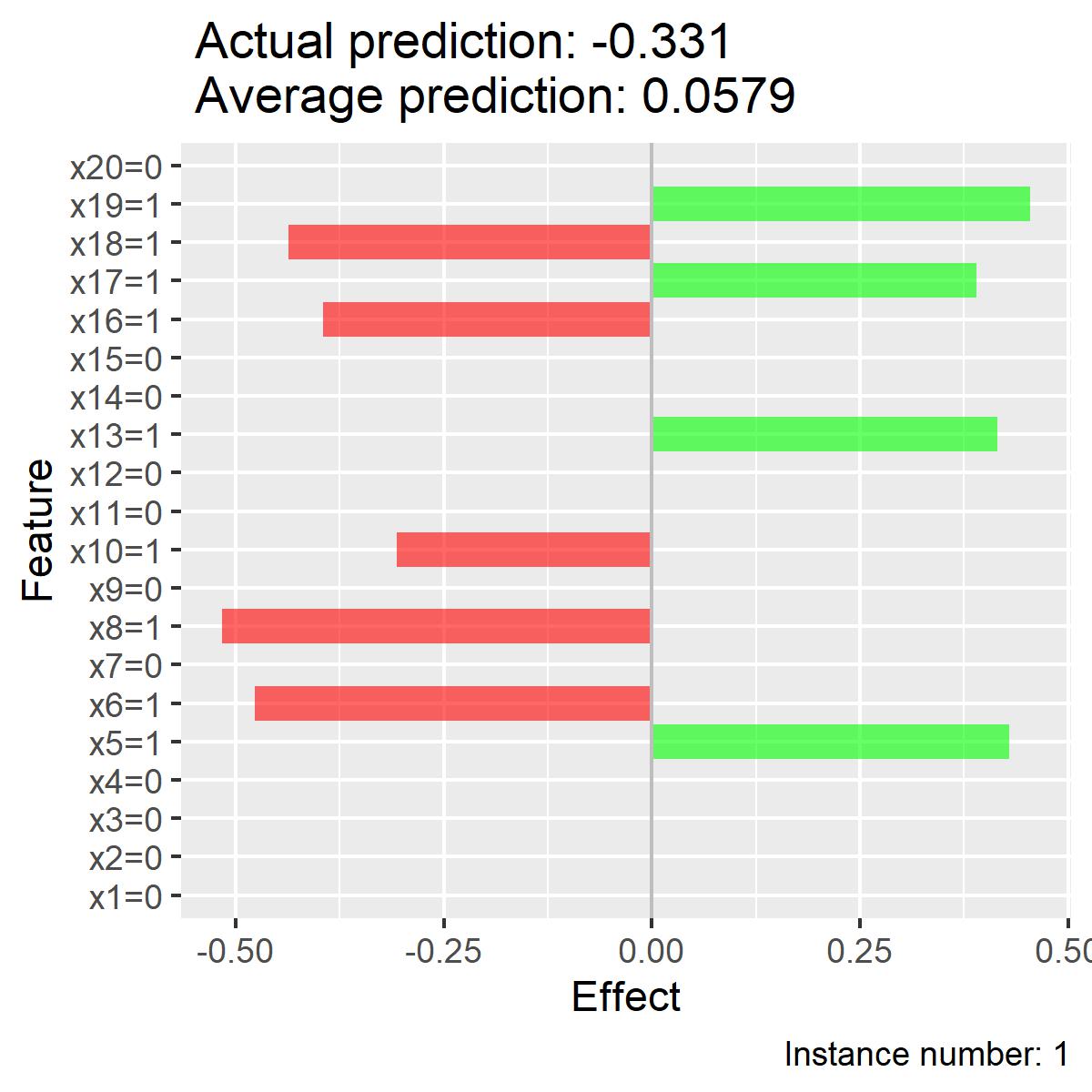}
            \caption[]%
            {{\small LIME}} 
        \end{subfigure}
        \hfill
        \begin{subfigure}[b]{0.32\textwidth}  
            \centering 
            \includegraphics[width=\textwidth]{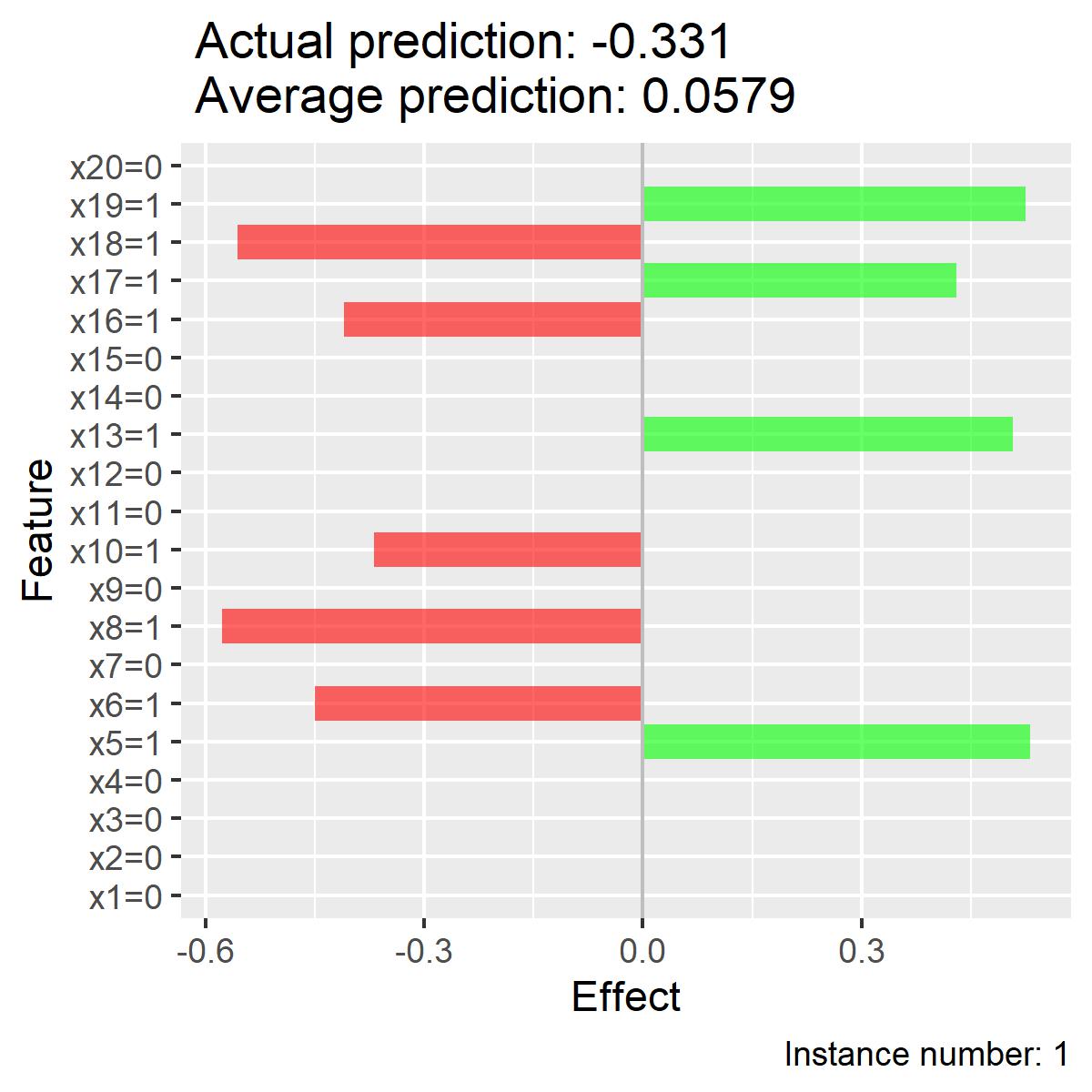}
            \caption[]%
            {{\small IML}} 
        \end{subfigure}
        \begin{subfigure}[b]{0.32\textwidth}  
            \centering 
            \includegraphics[width=\textwidth]{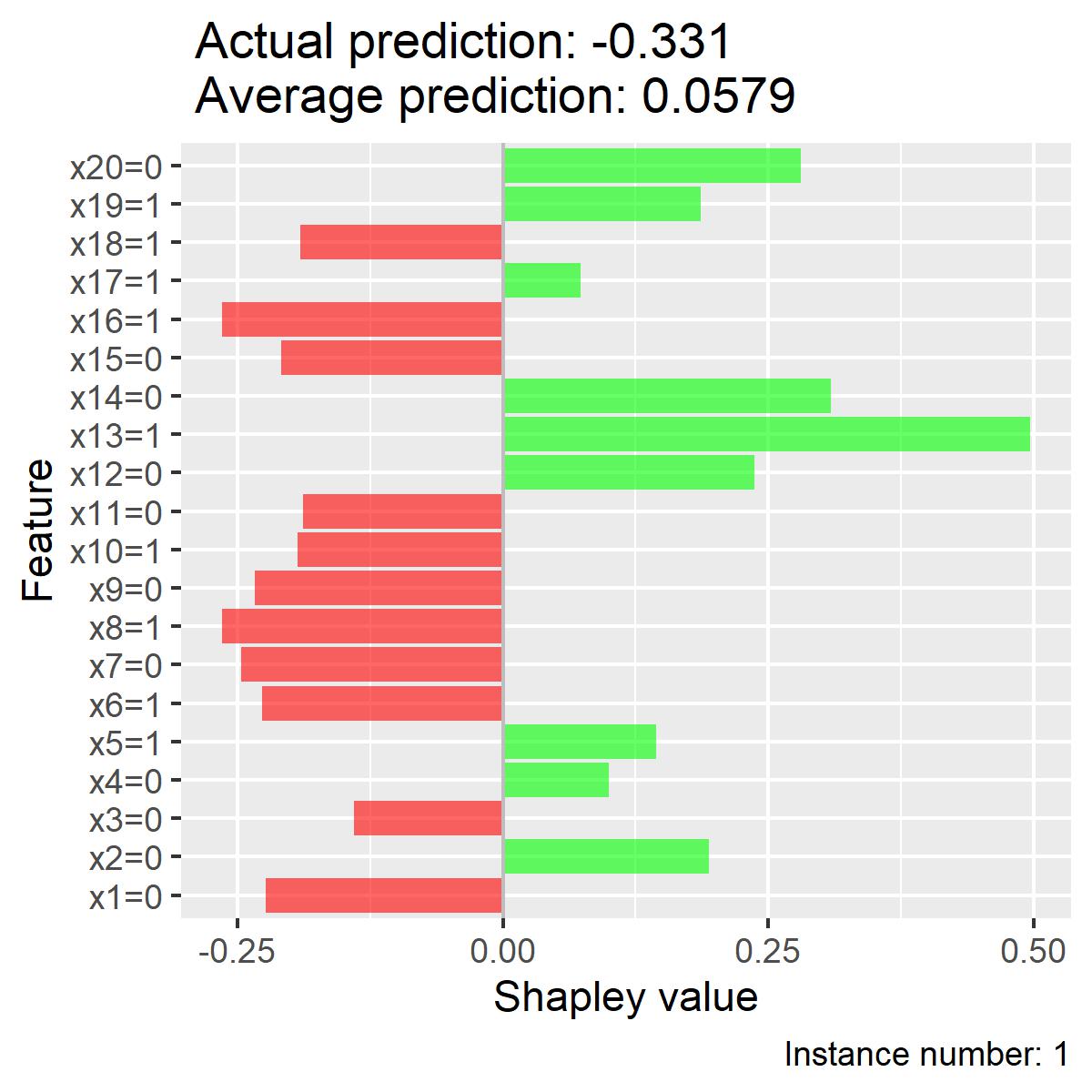}
            \caption[]%
            {{\small Shapley values}} 
        \end{subfigure}
        \caption[]
        {\small Estimated effects of an instance for dataset 4.} 
\end{figure}

\begin{figure}[H]
        \centering
        \begin{subfigure}[b]{0.32\textwidth}
            \centering
            \includegraphics[width=\textwidth]{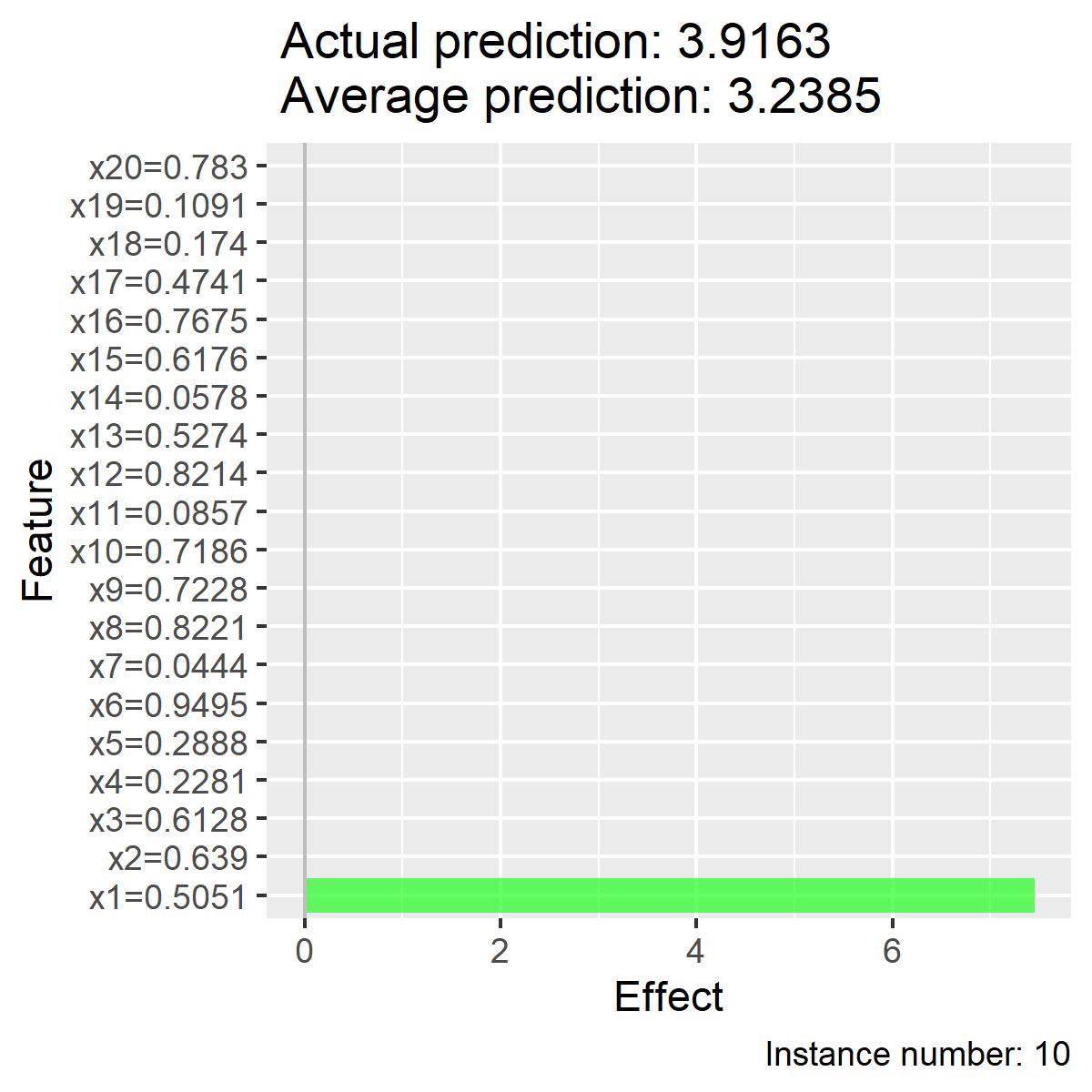}
            \caption[]%
            {{\small VarImp}}   
        \end{subfigure}
        \begin{subfigure}[b]{0.32\textwidth}  
            \centering 
            \includegraphics[width=\textwidth]{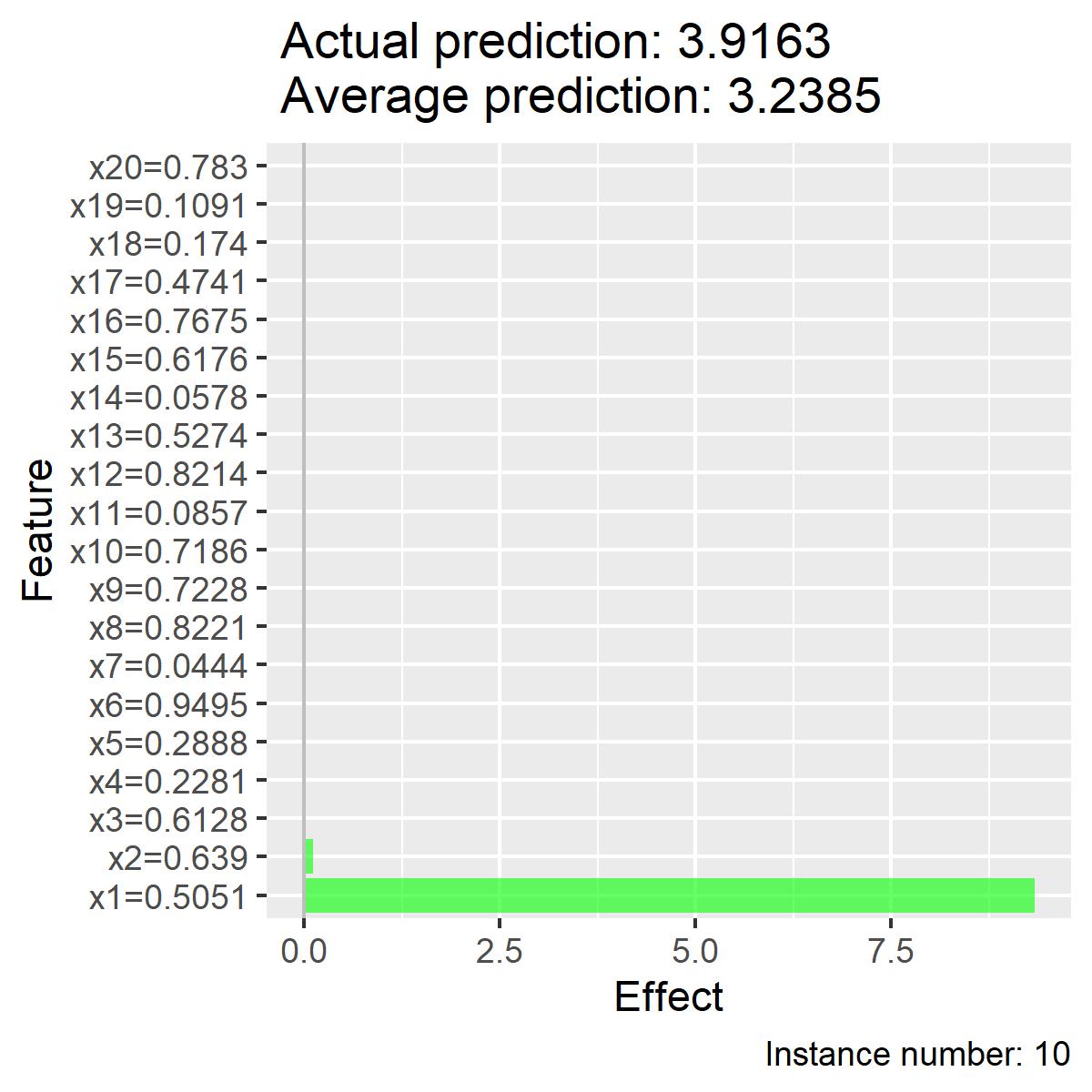}
            \caption[]%
            {{\small SupClus}} 
        \end{subfigure}
        \begin{subfigure}[b]{0.32\textwidth}  
            \centering 
            \includegraphics[width=\textwidth]{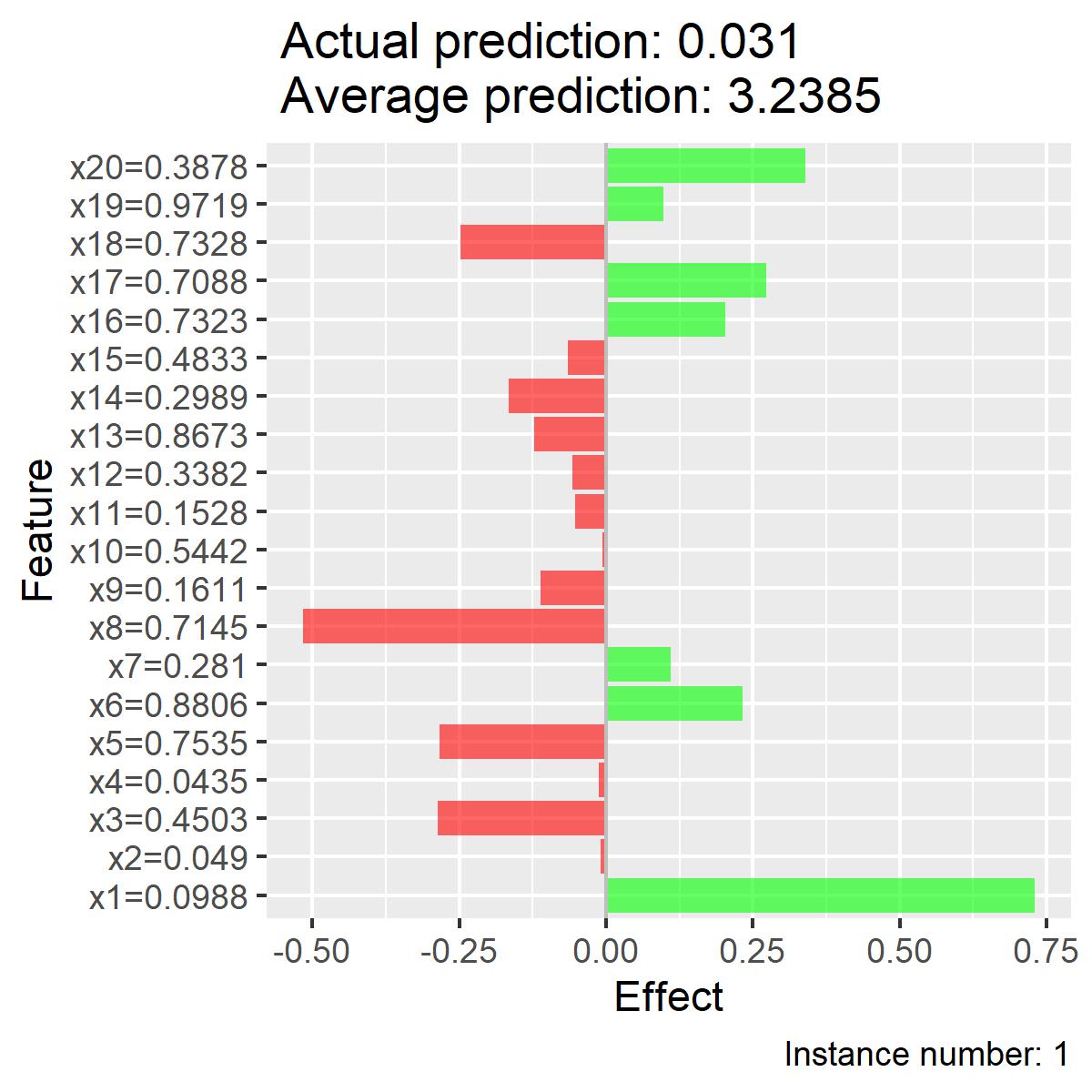}
            \caption[]%
            {{\small LIME}} 
        \end{subfigure}
        \hfill
        \begin{subfigure}[b]{0.32\textwidth}  
            \centering 
            \includegraphics[width=\textwidth]{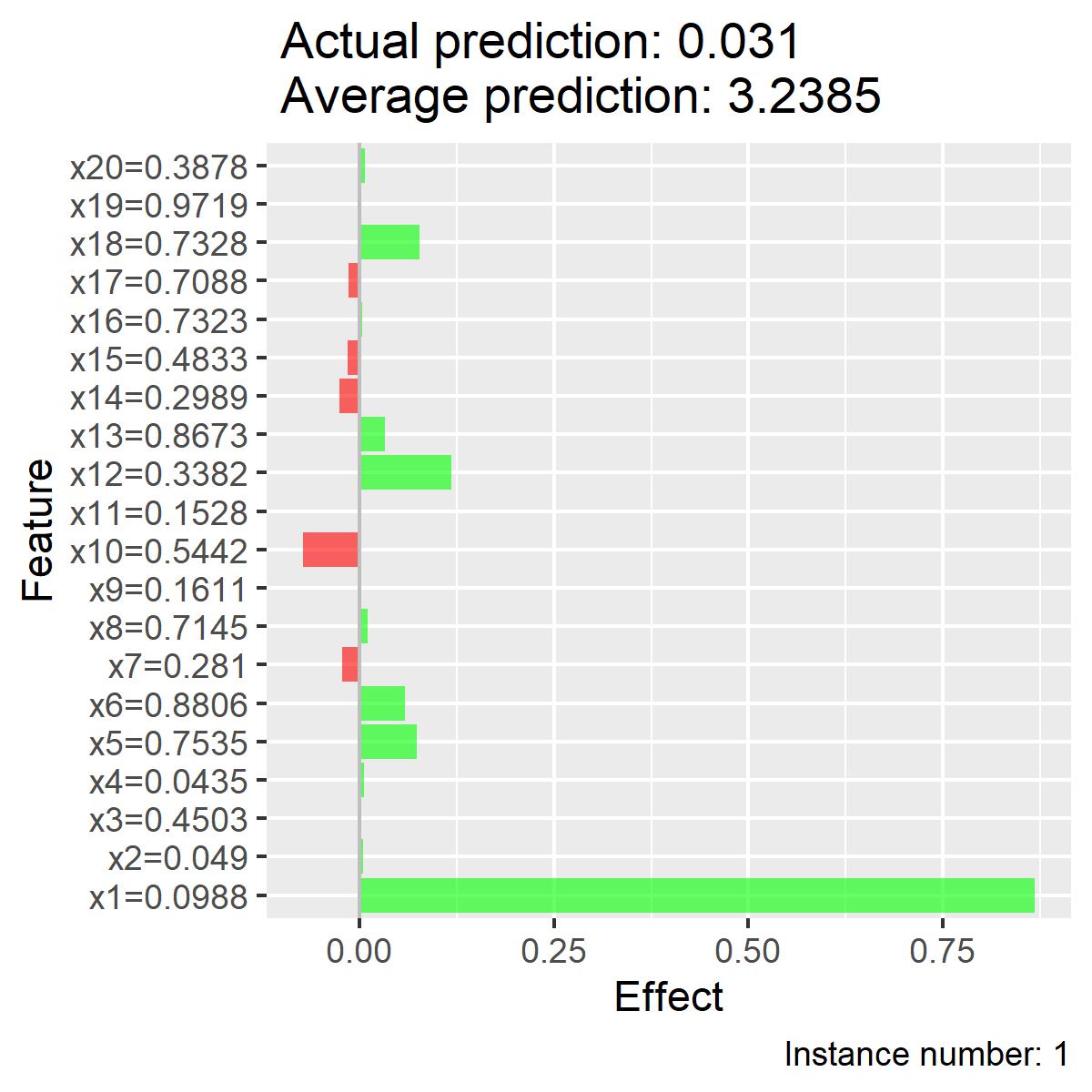}
            \caption[]%
            {{\small IML}} 
        \end{subfigure}
        \begin{subfigure}[b]{0.32\textwidth}  
            \centering 
            \includegraphics[width=\textwidth]{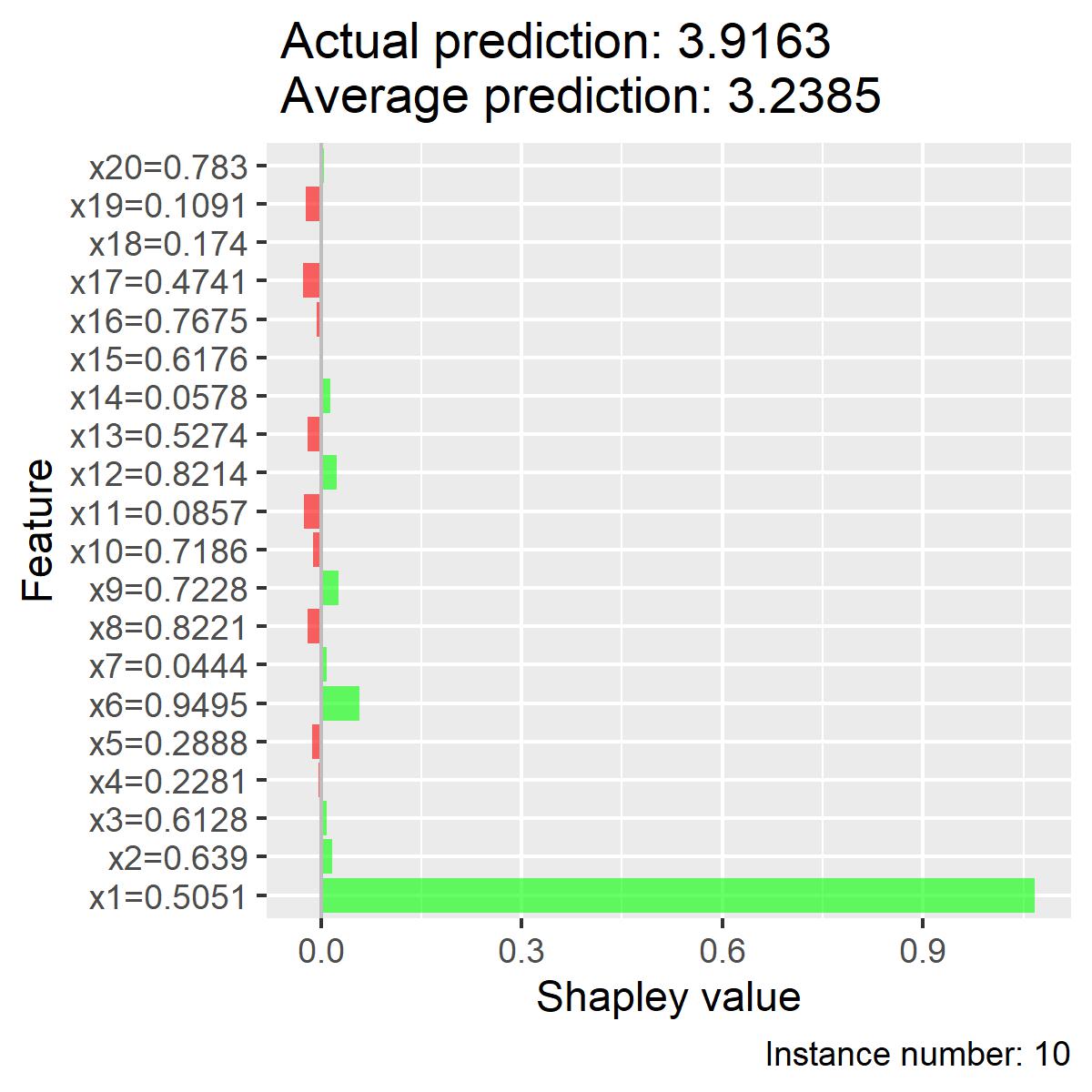}
            \caption[]%
            {{\small Shapley values}} 
        \end{subfigure}
        \caption[]
        {\small Estimated effects of an instance for dataset 5.} 
\end{figure}

\begin{figure}[H]
        \centering
        \begin{subfigure}[b]{0.32\textwidth}
            \centering
            \includegraphics[width=\textwidth]{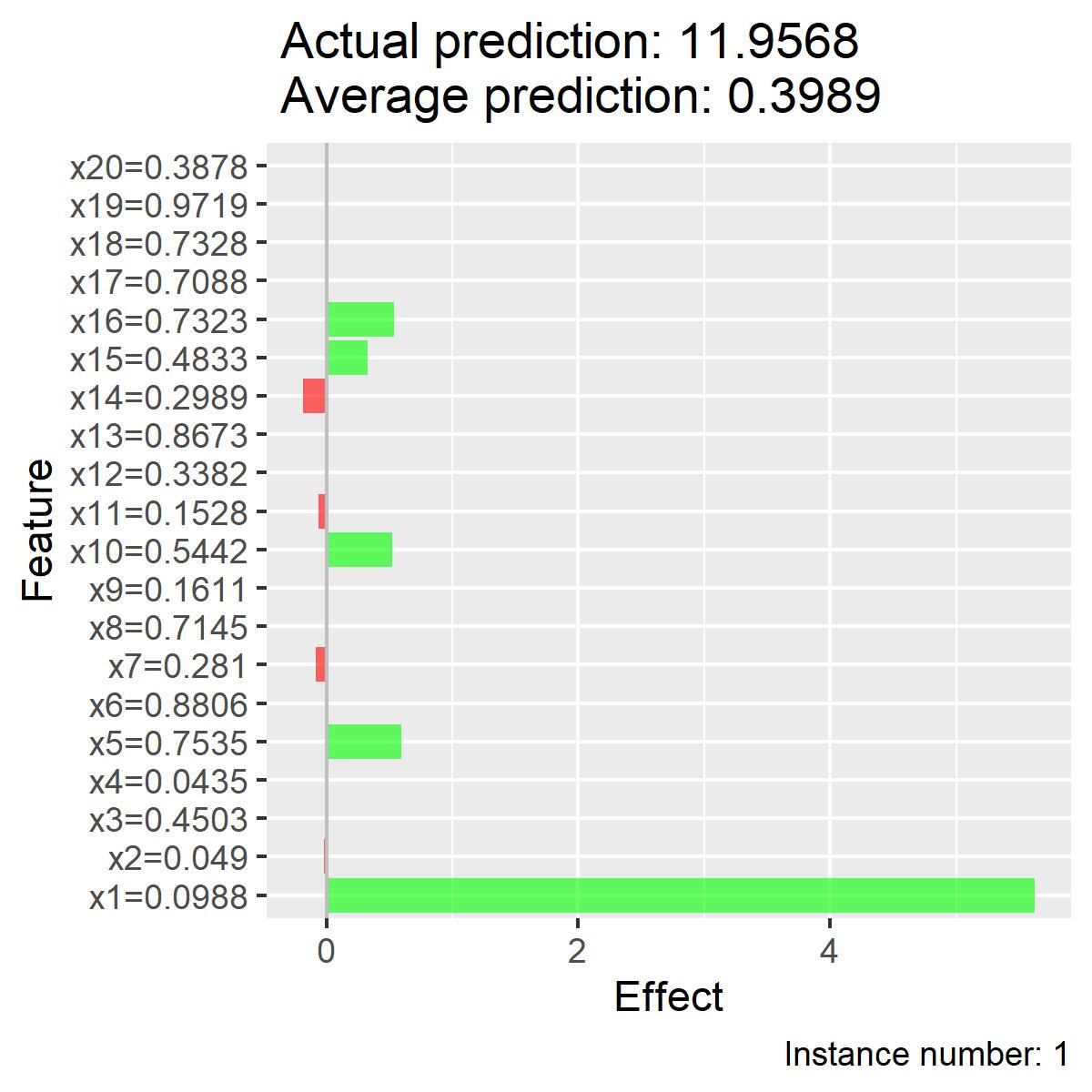}
            \caption[]%
            {{\small VarImp}}   
        \end{subfigure}
        \begin{subfigure}[b]{0.32\textwidth}  
            \centering 
            \includegraphics[width=\textwidth]{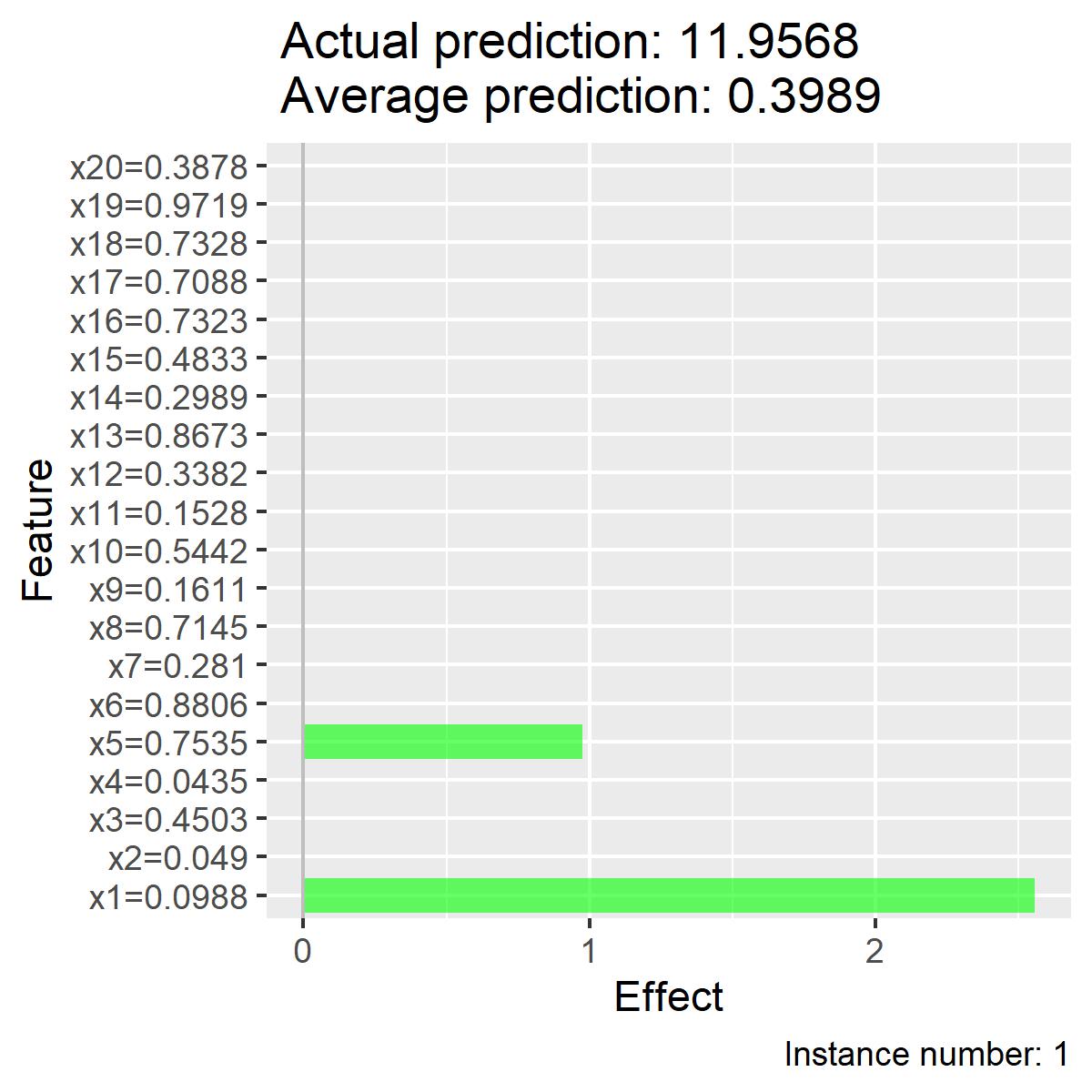}
            \caption[]%
            {{\small SupClus}} 
        \end{subfigure}
        \begin{subfigure}[b]{0.32\textwidth}  
            \centering 
            \includegraphics[width=\textwidth]{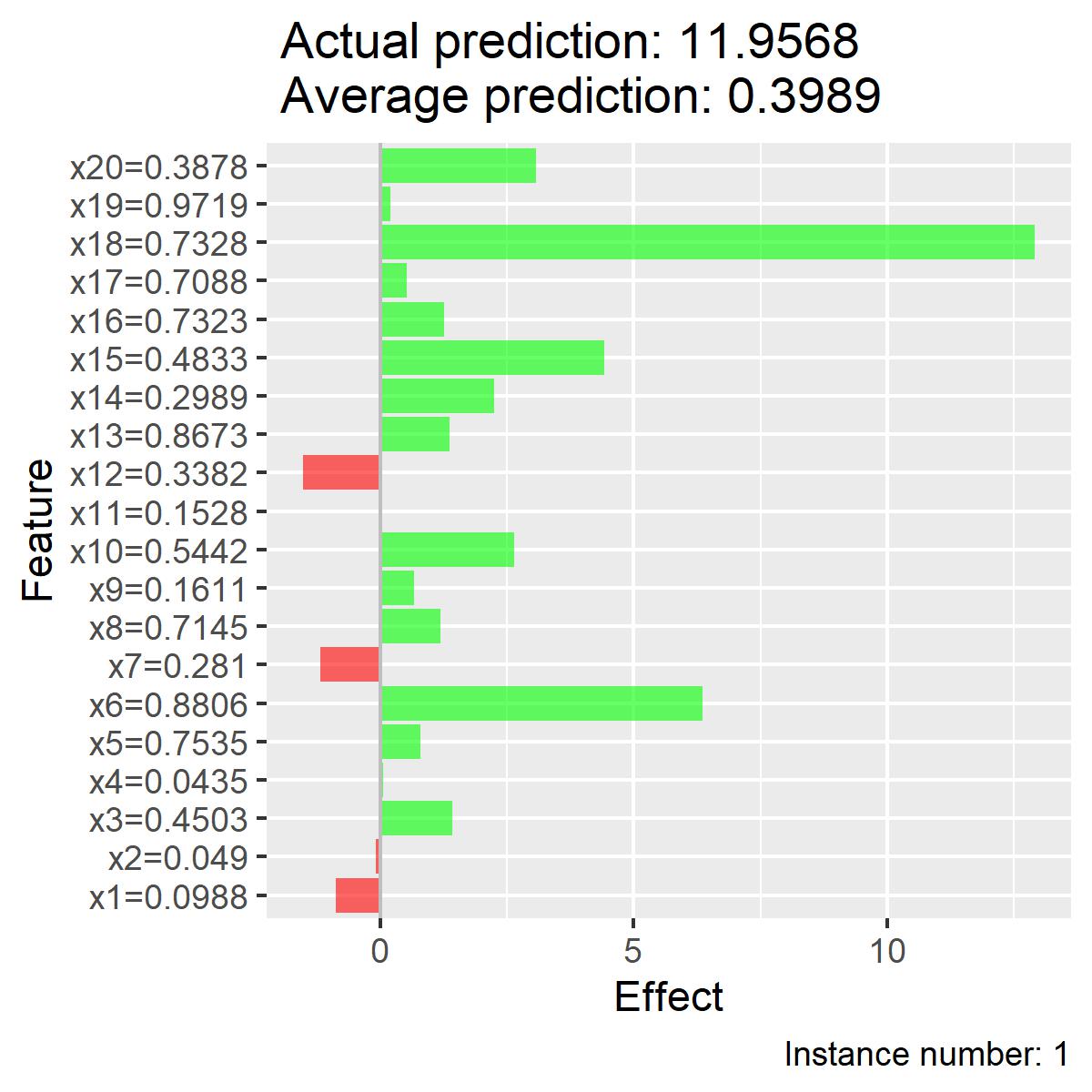}
            \caption[]%
            {{\small LIME}} 
        \end{subfigure}
        \hfill
        \begin{subfigure}[b]{0.32\textwidth}  
            \centering 
            \includegraphics[width=\textwidth]{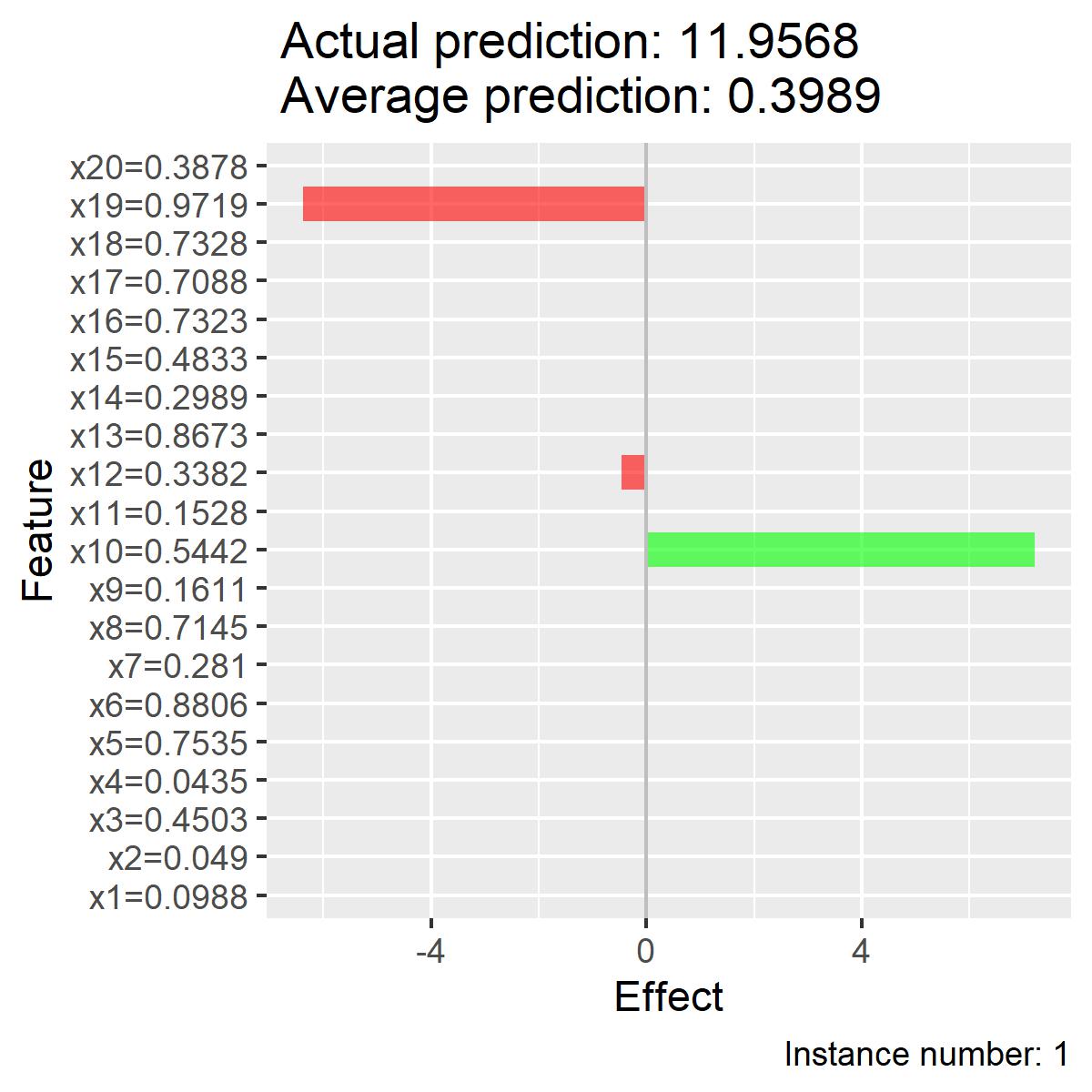}
            \caption[]%
            {{\small IML}} 
        \end{subfigure}
        \begin{subfigure}[b]{0.32\textwidth}  
            \centering 
            \includegraphics[width=\textwidth]{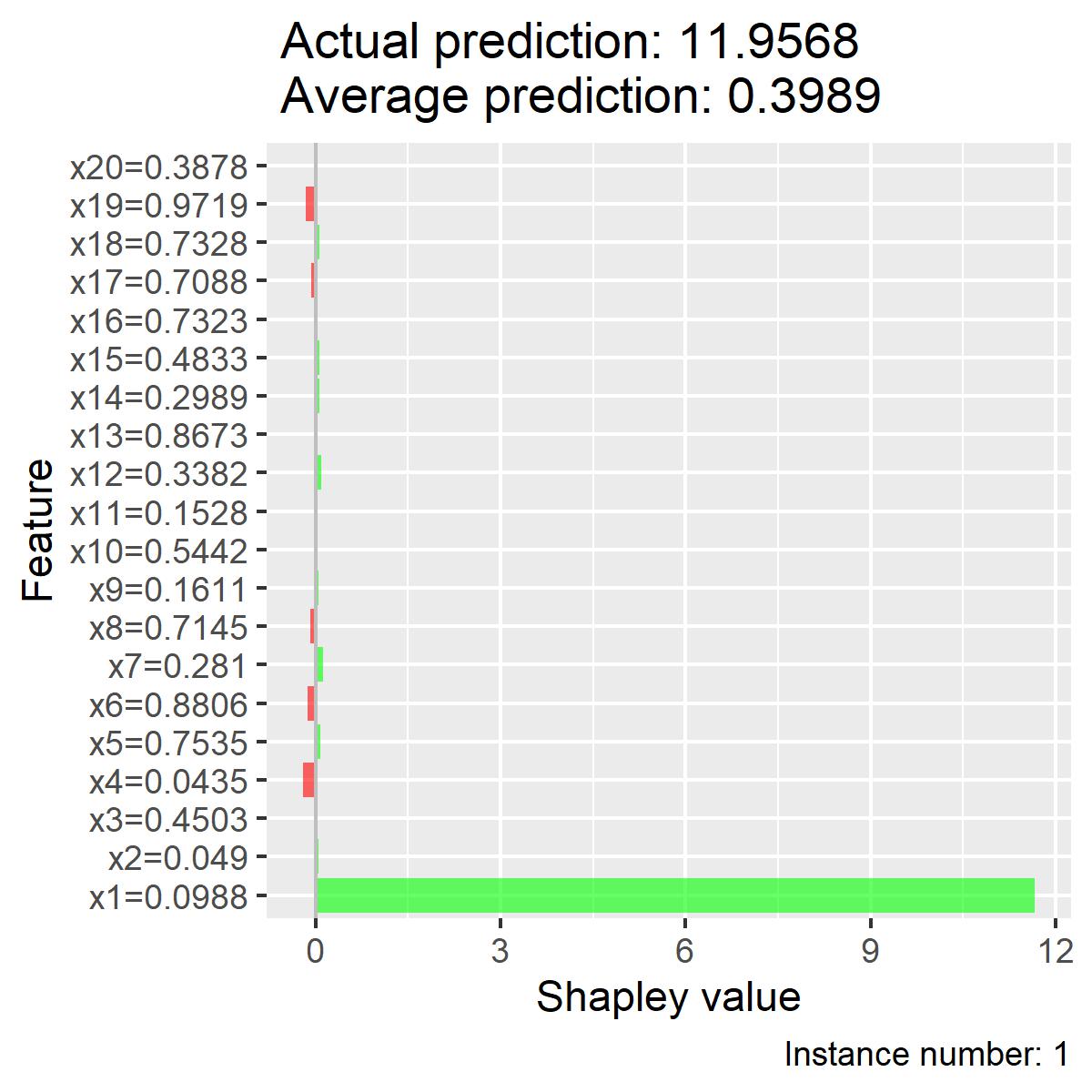}
            \caption[]%
            {{\small Shapley values}} 
        \end{subfigure}
        \caption[]
        {\small Estimated effects of an instance for dataset 6.} 
\end{figure}




\end{appendices}


\bibliography{sn-bibliography}


\end{document}